\documentclass{article} 
\usepackage{iclr2027_conference,times}
\usepackage{wrapfig}
\usepackage{placeins}
\usepackage{float}     
\newcommand{\benchlab}[1]{\makebox[0.106\linewidth][c]{\fontsize{5.5}{6.2}\selectfont #1}}   

\usepackage{amsmath,amsfonts,bm}

\def\eqref#1{equation~\ref{#1}}

\def\1{\bm{1}}

\DeclareMathAlphabet{\mathsfit}{\encodingdefault}{\sfdefault}{m}{sl}
\SetMathAlphabet{\mathsfit}{bold}{\encodingdefault}{\sfdefault}{bx}{n}

\usepackage{hyperref}
\usepackage{url}
\usepackage{graphicx}   
\usepackage{booktabs}
\usepackage{tikz}
\usetikzlibrary{arrows.meta,positioning,fit,backgrounds,calc,shapes.geometric}

\title{The Domain Is a Residue: Adapting \\ Self-Supervised Features, Not Generators}

\author{Thomas Deixelberger \& Markus Steinberger \\
Huawei, Austria  \\
Graz University of Technology, Austria  \\
\url{https://d3ixi.github.io/RepresentationFeatureAdapter/}
}

\iclrfinalcopy 
\newcommand{\snow}[1][blue!70]{\makebox[3.6mm]{\tikz[baseline=-0.55ex,scale=0.115]{
  \foreach \a in {0,60,120}{\draw[#1,line width=0.55pt] (\a:1) -- (\a+180:1);}
  \foreach \a in {0,60,120,180,240,300}{
    \draw[#1,line width=0.45pt] (\a:0.58) -- ++(\a+42:0.36);
    \draw[#1,line width=0.45pt] (\a:0.58) -- ++(\a-42:0.36);}}}}
\newcommand{\flame}[1][orange!88!red]{\makebox[3.2mm]{\tikz[baseline=-0.55ex,scale=0.115]{
  \fill[#1] (0,-1) .. controls (1.1,-0.2) and (0.4,0.55) .. (0.26,1.2)
            .. controls (0.05,0.6) and (-0.22,0.7) .. (-0.36,0.34)
            .. controls (-1.1,-0.05) and (-0.72,-0.6) .. (0,-1);}}}

\begin{document}

\maketitle

\begin{abstract}
Clearing fog, rain or snow from footage, or turning renders into photographs, must remove the source domain and keep the scene. Unpaired translators carry it through because their generator sees the source appearance (pixels, a near-invertible latent or a control map) and keeps it. A DINO feature map fixes what is in the scene and carries weather, lighting and rendering style as a residue of 13 to 14\,\% of the feature norm. We propose the Representation Feature Adapter (RFA), a 2.9\,M-parameter network that moves this residue. We train only the adapter and its discriminators; the encoder and a feature-conditioned decoder, trained once for all conditions, stay frozen. Against CycleGAN-Turbo it is ahead on both metrics on fog and on KID on night, and level within noise on snow, rain and haze. On sim-to-real it leads REGEN and HyPER-GAN on both metrics. Only the RFA removes the rain while keeping the scene. The removal costs scene structure: CycleGAN-Turbo keeps more on every condition but fog. On VAE latents the identical adapter collapses to the identity, and decoders from other groups that never saw it render its output. The RFA has about 160 times fewer trainable parameters than CycleGAN-Turbo and under a fifth of its per-condition training time.
\end{abstract}

\section{Introduction}
\label{sec:intro}

Unpaired translators carry the source domain through, because their generator is shown the source appearance and keeps what it is shown. Clearing fog, rain or snow from a frame, turning a render into a photograph and night into day all require the source domain to disappear while the scene stays. Pixel-space GANs such as CycleGAN \citep{zhu2017cyclegan} and CUT \citep{park2020cut} learn the translation from unpaired data but have to learn to generate as well. One-step diffusion finetuning such as CycleGAN-Turbo \citep{parmar2024cycleganturbo} adapts a pretrained generator until it produces the target domain, and controllable generators such as Cosmos-Transfer \citep{nvidia2025cosmostransfer} train nothing and steer a pretrained video model with control maps. All three show their generator the input, as pixels, as a near-invertible latent of them or as a control map computed from them, and we call such methods pixel-fed. The usual escape is a geometric control such as depth or edges, which captures only a narrow aspect of the scene \citep{dominici2026controldino} and leaves the generator to invent the rest.

A DINO feature map fixes what is in the scene and carries weather, lighting and rendering style only as a small residue. Dense self-supervised features carry the whole scene, its structure and semantics with the appearance entangled in them, and suffice to condition a frozen diffusion model \citep{dominici2026controldino}. We call the part of the feature map that an adapter must change to turn fog into clear weather, or a render into a photograph, the residue, and Section~\ref{sec:why-features} measures it. We therefore train a feature-conditioned decoder once to reconstruct images from their feature maps, freeze it, and translate in front of it by moving the residue and leaving the rest of the map in place.

We propose the \textbf{Representation Feature Adapter (RFA)}, a small feed-forward residual network that maps one DINO feature map to another (Section~\ref{sec:rfa}). It is trained CycleGAN-style, with the adversarial gradient passing back through the frozen decoder, and everything else it relies on is frozen: the encoder, the decoder and the vision model inside its discriminator \citep{kumari2022visionaided}. Because the adapter learns only the residue, it fits in a small network and trains in a few thousand steps. On VAE latents the identical adapter does not translate (Section~\ref{sec:why-features}), which shows that the feature space matters. To our knowledge, the RFA is the first learned mapping between domains inside a frozen self-supervised feature space.

We make four contributions, and the first is a measured diagnosis of why pixel-fed translators fail: their generator can still recover the source domain from what it is shown. The identical adapter translates on DINO features and collapses to the identity on VAE latents, and the domain it moves is 13 to 14\,\% of the feature norm (Section~\ref{sec:why-features}). Second, the RFA is a 2.9\,M-parameter adapter on a frozen encoder and decoder. It is ahead of a finetuned CycleGAN-Turbo on both distribution metrics (KID and FID) on fog and on KID on night, and level with it within noise on snow, rain and haze (Appendix~\ref{app:se}). On rain it removes the weather where the pixel-fed baselines that keep the scene carry the wet sheen through (Section~\ref{sec:weather}, Figure~\ref{fig:f1}, Appendix~\ref{app:gone}). Third, we measure the trade that removal costs, since the lossiness that lets the adapter remove the weather also moves scene structure, and we report both ends (Section~\ref{sec:cost-of-removal}). Fourth, because the adapter maps a feature map to a feature map, decoders from other groups that never saw it render its output (Section~\ref{sec:portability}).
\section{Related work}
\label{sec:related}

\textbf{What the generator is given.} Every unpaired translator family shows its generator the source appearance. Pixel-space GANs (CycleGAN \citep{zhu2017cyclegan}, CUT \citep{park2020cut} and the ViT-generator line \citep{torbunov2023uvcganv2}) see the input image. CycleGAN-Turbo \citep{parmar2024cycleganturbo} finetunes the pretrained one-step generator SD-Turbo \citep{sauer2023add} on the VAE-encoded input, and Cosmos-Transfer \citep{nvidia2025cosmostransfer} sees control maps computed from the input. UNIT-DDPM \citep{sasaki2021unitddpm}, CycleGAN's diffusion counterpart without a cycle term, translates by many-step sampling conditioned on the source image. Our decoder sees a semantic feature map, and Section~\ref{sec:weather} measures what that changes.

\textbf{Translating in feature space.} Earlier feature-space translators differ in whose features they adapt, whether they render, or what they train. CyCADA \citep{hoffman2018cycada} adapted a task network's own activations with a jointly trained decoder and found feature-level adaptation stronger on harder shifts; we freeze encoder and decoder. DosGAN \citep{lin2019dosgan} learns to split its own encoder's code into domain-specific and domain-independent parts; we measure this split in a frozen self-supervised feature space (Section~\ref{sec:why-features}). DA-F2F \citep{oh2026daf2f} modulates a detector's features and never renders. Our nearest neighbour, the Self-Supervised Semantic Bridge \citep{ssb2026bridge}, passes through a frozen, appearance-invariant DINO feature space only as a meeting point for generators trained per domain. It learns nothing inside that feature space, so a new pair needs a new generator, while we learn the map there and train no generator. PRISM \citep{prism2026} adapts inside a frozen text-to-image generator, whereas we adapt in front of one. Cyclone \citep{nguyen2026cyclone}, the closest contemporaneous work on our task, finetunes a Stable Diffusion UNet on the source frame's VAE latent, so its generator is pixel-fed.

\textbf{Self-supervised features as a generative space.} DINOv2 and DINOv3 features \citep{oquab2024dinov2,simeoni2025dinov3} are an established stand-in for a VAE latent. Representation Autoencoders \citep{zheng2026rae,singh2026raev2} and VFM-VAE \citep{bi2026vfmvae} train a decoder for them. Representation alignment \citep{yu2025repa} speeds up diffusion training in this feature space, SVG \citep{shi2026svg} runs the diffusion itself there, and PiD \citep{lu2026pid} decodes it with pixel diffusion. DINO-WM \citep{zhou2025dinowm} and MIRA \citep{hu2026mira} build world models on it because it stays recoverable under prediction error, unlike a reconstruction codec's latent \citep{nilaksh2026latent}. None of them tests whether this feature space is a good place for domain adaptation.

\textbf{Conditioning on features, and adapting them.} Our decoder, the case for swapping it, and our adversarial signal come from work on frozen features. Control-DINO \citep{dominici2026controldino} conditions a video diffusion ControlNet \citep{zhang2023controlnet} on dense DINOv3 features and is our decoder's ancestor. Driving with DINO \citep{drivingwithdino2026} does this on a frozen Cosmos model and reports the trade of realism against scene structure we also measure. Splicing ViT Features \citep{tumanyan2022splicing} edits a frozen DINO feature space per image, without a generative model. Latent-space anchoring \citep{latentanchoring2023} and FAE \citep{fae2025} argue for swappable decoders. FAE treats the tension that representation encoders favour high-dimensional features and generators low-dimensional latents as a mismatch to close, whereas we exploit it. Our zero-initialised head is ControlNet's zero-initialised residual \citep{zhang2023controlnet,adaptertune2026}. The adversarial signal follows vision-aided GAN training \citep{kumari2022visionaided} with a frozen DINOv2 inside the discriminator, so encoder, adapter and discriminator share one feature space.

\textbf{The two applications.} Prior work on both applications differs from ours in failure mode or protocol. Sim-to-real photorealism runs from EPE \citep{richter2022epe} to its real-time distillations REGEN \citep{regen2026} and HyPER-GAN \citep{hypergan2026}, the baselines of Table~\ref{tab:t3}, which keep scene structure exactly and transfer appearance only, the opposite failure mode to ours. Adverse-weather restoration \citep{xrestormerpp2026,weatherbench2025} evaluates on paired synthetic degradations with PSNR and SSIM. Our ACDC \citep{sakaridis2021acdc} and haze settings are unpaired real data scored with KID \citep{binkowski2018kid} and structure distance, so the protocols are not interchangeable. Structure distance \citep{tumanyan2022splicing} compares an output with its own input: it is the mean squared difference of their DINO key self-similarity maps, and it is the DINO column of our tables.
\section{Why features and not latents}
\label{sec:why-features}
\begin{table}[t]
\caption{The identical adapter and recipe on three conditionings. \emph{Raw change} is how much the adapter changes its input; \emph{translation} is that change with a per-channel affine colour map factored out, so a global re-grade does not count; \emph{colour share} is the part of the raw change that the affine map explains. A negative share means the change is not a global colour shift at all.}
\label{tab:mechanism}
\begin{center}
\footnotesize\setlength{\tabcolsep}{4pt}
\begin{tabular}{lcccc}
\toprule
\textbf{conditioning} & \textbf{reconstruction MAE} $\downarrow$ & \textbf{raw change} &
\textbf{colour share} & \textbf{translation} $\uparrow$ \\
\midrule
Sana DC-AE latent (32$\times$32$\times$32) & \textbf{0.042} & 0.583 & 43\,\% & 0.331 \\
FLUX VAE latent (16$\times$128$\times$128)  & 0.061 & 0.404 & 41\,\% & 0.239 \\
DINOv2-reg (768$\times$32$\times$32)      & 0.113 & 0.516 & $-11$\,\% & \textbf{0.572} \\
\bottomrule
\end{tabular}
\end{center}
\end{table}

On a VAE latent the RFA collapses to the identity or re-grades the image, although the latent reconstructs better (Table~\ref{tab:mechanism}). We trained the identical adapter on two latents, the DC-AE of our decoder's backbone (Sana-Sprint, Section~\ref{sec:pipeline}) and FLUX's VAE, which reconstruct at a third to a half of DINOv2's error. At the loss weights of Section~\ref{sec:objective} both stay at the identity, and when pushed out of collapse by a higher adversarial weight they re-grade the image, with a global tone shift accounting for over 40\,\% of what they change. With that shift factored out, the adapter on DINOv2 does 1.7 to 2.4 times the translation of either latent. The two latents thus reconstruct better, re-grade and translate less, while DINOv2 reconstructs worst, shows no colour share and translates most.

One mechanism predicts this split: the adapter's leverage is whatever the conditioning leaves undetermined. DINO features are lossy and semantic, and they were never trained to be decoded. Many images share one feature map, so the map fixes \emph{what is in the scene} and leaves \emph{how it looks} to the backbone's appearance prior. A VAE latent is near-invertible and leaves the backbone nothing to decide. Every change of appearance must then be written by the adapter inside the manifold of valid latents, where the cheapest direction is a global re-grade.

The residue, what the adapter changes on a DINO feature map, is 13 to 14\,\% of the feature norm in every condition, fog, snow, rain, haze and night alike. We measure it everywhere in the paper as the adapter's relative move on a test frame's feature map, $\|G(f)-f\|/\|f\|$, averaged over each condition's test crops. Here $G$ is the source-to-target adapter and $f$ the DINOv2-B/reg map in the decoder's per-channel normalisation (Appendix~\ref{app:cn}). On an already clear input the trained adapters of Section~\ref{sec:results} move the map by 3 to 6\,\% (Table~\ref{tab:residue}), and any other photograph, of any weather, sits about a full norm away. We also report a DINOv3-S configuration for replication, which reads DINOv3-S features through a 0.6\,B ControlNet (Section~\ref{sec:pipeline}). There every distance is about half as large, and this measure reads 5 to 9\,\%. In both feature spaces the move is 10 to 16\,\% of the distance between two frames of one weather (Appendix~\ref{app:residue}).
\section{Method}
\label{sec:method}

\subsection{Pipeline}
\label{sec:pipeline}

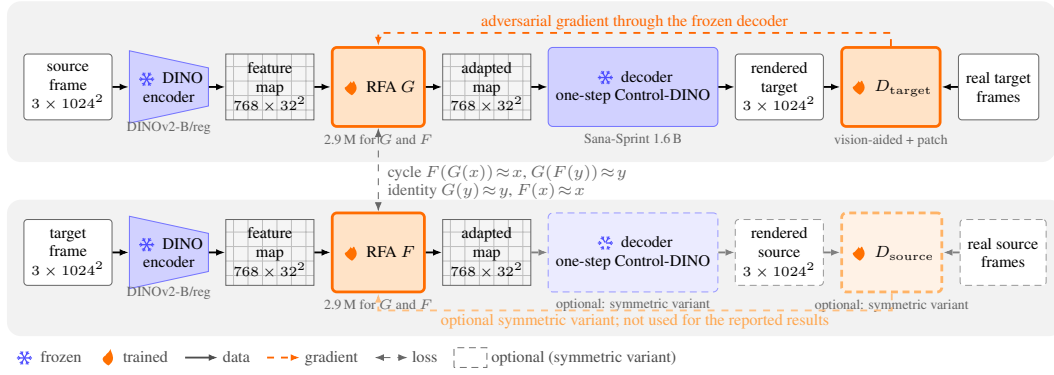
\begin{figure}[t]
\centering
\resizebox{\linewidth}{!}{%
\begin{tikzpicture}[node distance=3mm, every node/.style={font=\scriptsize}]
\tikzset{
  frame/.style   = {draw=black!60, fill=white, rounded corners=1.5pt, minimum width=10mm, minimum height=10mm, align=center, font=\scriptsize},
  fmap/.style    = {draw=black!60, fill=black!4, minimum width=10mm, minimum height=10mm, align=center, font=\scriptsize,
                    path picture={\draw[black!25, line width=0.3pt] (path picture bounding box.south west) grid[step=2.2mm] (path picture bounding box.north east);}},
  encoder/.style = {trapezium, shape border rotate=270, trapezium angle=78, trapezium stretches=true, draw=blue!55, fill=blue!18,
                    minimum height=12mm, minimum width=11mm, inner sep=1pt, align=center, font=\scriptsize},
  adapter/.style = {draw=orange!85!red, fill=orange!22, very thick, rounded corners=2pt, minimum width=14mm, minimum height=12mm, align=center, font=\scriptsize},
  decoder/.style = {draw=blue!55, fill=blue!18, rounded corners=2pt, minimum width=19mm, minimum height=12mm, align=center, font=\scriptsize},
  critic/.style  = {draw=orange!85!red, fill=orange!22, very thick, rounded corners=2pt, minimum width=12mm, minimum height=12mm, align=center, font=\scriptsize},
  realbox/.style = {draw=black!60, fill=white, rounded corners=1.5pt, minimum width=10mm, minimum height=10mm, align=center, font=\scriptsize},
  flow/.style    = {-{Latex[length=4pt]}, thick},
  grad/.style    = {-{Latex[length=4pt]}, dashed, orange!80!red, thick},
  loss/.style    = {{Latex[length=4pt]}-{Latex[length=4pt]}, dashed, black!60},
  lab/.style     = {font=\scriptsize, black!60, inner sep=1pt},
  row/.style     = {draw=none, fill=black!5, rounded corners=6pt, inner xsep=2.5mm, inner ysep=6mm},
}
\node (x)    [frame] {source\\frame\\{\tiny $3\times1024^2$}};
\node (encA) [encoder, right=2.5mm of x] {\snow\ DINO\\encoder};
\node (fx)   [fmap, right=2.5mm of encA] {feature\\map\\{\tiny $768\times32^2$}};
\node (G)    [adapter, right=2.5mm of fx] {\flame\ RFA $G$};
\node (fxa)  [fmap, right=2.5mm of G] {adapted\\map\\{\tiny $768\times32^2$}};
\node (decA) [decoder, right=2.5mm of fxa] {\snow\ decoder\\one-step Control-DINO};
\node (yhat) [frame, right=2.5mm of decA] {rendered\\target\\{\tiny $3\times1024^2$}};
\node (DY)   [critic, right=2.5mm of yhat] {\flame\ $D_{\mathrm{target}}$};
\foreach \a/\b in {x/encA, encA/fx, fx/G, G/fxa, fxa/decA, decA/yhat, yhat/DY} {\draw [flow] (\a) -- (\b);}
\node (y)    [frame, below=15mm of x] {target\\frame\\{\tiny $3\times1024^2$}};
\node (encB) [encoder, right=2.5mm of y] {\snow\ DINO\\encoder};
\node (fy)   [fmap, right=2.5mm of encB] {feature\\map\\{\tiny $768\times32^2$}};
\node (F)    [adapter, right=2.5mm of fy] {\flame\ RFA $F$};
\node (fya)  [fmap, right=2.5mm of F] {adapted\\map\\{\tiny $768\times32^2$}};
\node (decB) [decoder, densely dashed, fill=blue!8, draw=blue!40, right=2.5mm of fya] {\snow\ decoder\\one-step Control-DINO};
\node (xhat) [frame, densely dashed, draw=black!40, right=2.5mm of decB] {rendered\\source\\{\tiny $3\times1024^2$}};
\node (DX)   [critic, densely dashed, fill=orange!10, draw=orange!60, right=2.5mm of xhat] {\flame\ $D_{\mathrm{source}}$};
\foreach \a/\b in {y/encB, encB/fy, fy/F, F/fya} {\draw [flow] (\a) -- (\b);}
\foreach \a/\b in {fya/decB, decB/xhat, xhat/DX} {\draw [flow, densely dashed, black!50] (\a) -- (\b);}
\tikzset{sub/.style={font=\tiny, black!60, inner sep=0.5pt, anchor=north}}
\foreach \n/\t in {encA/{DINOv2-B/reg}, encB/{DINOv2-B/reg}, G/{2.9\,M for $G$ and $F$}, F/{2.9\,M for $G$ and $F$},
                   decA/{Sana-Sprint 1.6\,B}, DY/{vision-aided + patch},
                   decB/{optional: symmetric variant}, DX/{optional: symmetric variant}} {\node [sub] at ($(\n.south)+(0,-0.7mm)$) {\t};}
\node (ry) [realbox, right=2.5mm of DY] {real target\\frames};
\node (rx) [realbox, densely dashed, draw=black!40, right=2.5mm of DX] {real source\\frames};
\draw [flow] (ry) -- (DY);
\draw [flow, densely dashed, black!50] (rx) -- (DX);
\begin{scope}[on background layer]
  \coordinate (row1top) at ($(DY.north)+(0,5.0mm)$);
  \coordinate (row1bot) at ($(G.south)+(0,-3.4mm)$);
  \node (row1) [row, inner ysep=1.8mm, fit=(x)(DY)(ry)(row1top)(row1bot)] {};
  \coordinate (row2bot) at ($(F.south)+(0,-4.7mm)$);
  \node (row2) [row, inner ysep=1.8mm, fit=(y)(DX)(rx)(row2bot)] {};
\end{scope}
\draw [loss] (G.south) -- node[lab, right=1mm, align=left] {\\\\cycle $F(G(x))\!\approx\!x$, $G(F(y))\!\approx\!y$\\identity $G(y)\!\approx\!y$, $F(x)\!\approx\!x$} (F.north);
\draw [grad] (DY.north) |- ($(DY.north)+(0,2mm)$) -| node[lab, pos=0.25, above, orange!80!red] {adversarial gradient through the frozen decoder} (G.north);
\draw [grad, orange!50] (DX.south) |- ($(DX.south)+(0,-2.6mm)$) -| node[lab, pos=0.25, below, orange!70] {optional symmetric variant; not used for the reported results} (F.south);
\node [lab, anchor=north west, black!70, align=left] at ($(row2.south west)+(0.5mm,-1.2mm)$) {%
  \snow\ frozen\quad
  \flame\ trained\quad
  \tikz[baseline=-0.6ex]{\draw[flow] (0,0) -- (5mm,0);}\ data\quad
  \tikz[baseline=-0.6ex]{\draw[grad] (0,0) -- (5mm,0);}\ gradient\quad
  \tikz[baseline=-0.6ex]{\draw[loss] (0,0) -- (5mm,0);}\ loss\quad
  \tikz[baseline=-0.6ex]{\draw[densely dashed, black!55] (0,-1.5mm) rectangle (5mm,1.5mm);}\ optional (symmetric variant)};
\end{tikzpicture}}
\caption{\textbf{Training is one-sided by default.} Top row: the frozen encoder maps the frame to a feature map, $G$ adapts it, the frozen decoder renders it, and the target discriminators judge the render against real frames, their gradient reaching $G$ through the decoder. Bottom row: $F$ maps target features back and learns only from the cycle and identity terms. Dashed: the symmetric variant (Section~\ref{sec:objective}). At inference only the top row runs, with the decoder at two steps.}
\label{fig:pipeline}
\end{figure}

We train only the two RFA adapters ($G$ and $F$, one per direction) and the discriminators, and keep the encoder, the ControlNet and the Sana-Sprint backbone frozen throughout. The decoder, meaning the ControlNet and the backbone it drives, only maps feature maps to pixels, so every domain pair shares one decoder and a new pair needs only a new adapter. CycleGAN-Turbo instead builds one generator per domain pair, with the translation baked into LoRA weights inside the UNet.

\textbf{The decoder has to be in the loop, and differentiable in one pass.} With a feature-space discriminator alone, the adapter either stays at the identity or shifts the feature statistics in directions the frozen decoder discards, so the rendered image barely changes. The residue becomes identifiable only when the discriminator judges the rendered image and the gradient returns through the frozen decoder (Appendix~\ref{app:mechanism}). We therefore compute the adversarial and pixel-space terms on the decoded image and back-propagate them through the frozen decoder to the adapter. This needs a decoder that maps features to pixels in one differentiable pass, because unrolling several sampling steps would hamper gradient propagation and multiply the memory and cost of every step. Ours is a DINO ControlNet on Sana-Sprint 1.6\,B \citep{chen2025sanasprint}, a distilled few-step backbone, trained once against the backbone's teacher and run on the student at 1024\,$\times$\,1024 (Appendix~\ref{app:cn}). The decoder runs one step during adapter training and two at inference, its shipped default (Appendix~\ref{app:onestep}). Any single differentiable pass from a feature map to pixels qualifies, and Appendix~\ref{app:s6} trains the adapter through three such decoders, among them RAE's \citep{zheng2026rae,singh2026raev2}.

Every reported RFA result uses DINOv2-B/reg feature maps of size $768 \times 32 \times 32$, from a 448-px input, in the per-channel normalisation PiD consumes (Appendix~\ref{app:cn}). The 1.6\,B ControlNet was trained on this map, and PiD and RAE read it too (Section~\ref{sec:portability}). The DINOv3-S configuration swaps the encoder to test replication: it reads DINOv3-S at 1024\,px, a learned downscale brings its $384 \times 64 \times 64$ map to Sana's $32 \times 32$ control map, and a 0.6\,B ControlNet reads it.

\subsection{The RFA}
\label{sec:rfa}
The RFA reads a feature map with C channels and writes a map of equal shape. A 1 $\times$ 1 stem compresses C $\rightarrow$ 256 and four depthwise-separable dilated residual blocks follow (dilations 1, 2, 4, 1; GroupNorm-8; SiLU; MLP ratio 2). A zero-initialised 1 $\times$ 1 head returns 256 $\rightarrow$ C through a global residual and makes \texttt{forward(x) == x} at initialisation. Training thus starts from passing features through unchanged, so the frozen decoder is in distribution from step 0 and the adversarial loss never judges a meaningless render. The adapter is a feed-forward map and nothing more. It does no sampling or iteration and never sees the decoder, so anything that consumes a feature map can consume the adapted one (Section~\ref{sec:portability}). It uses convolutions, because full self-attention over a dense feature map is quadratic in the number of positions and is not needed to move the residue.

\subsection{Objective and discriminators}
\label{sec:objective}

The objective computes CycleGAN's cycle and identity terms on feature maps instead of pixels. $G$ maps source to target and $F$ maps target to source, and each should invert the other's output and leave its own domain unchanged. The adversarial part is one-sided by default (Figure~\ref{fig:pipeline}). Only $G$'s output is rendered through the frozen decoder and judged against real frames by the target domain's two discriminators. $G$ thus receives a gradient from an image, while $F$, the cycle's return path, learns from the feature-space terms alone. The symmetric variant, which also renders and judges $F$'s output, reached similar distribution scores with less structure drift in one matched sim-to-real comparison, at 1.5 times the time per step and more memory. Every reported result is one-sided.

Two discriminators judge the decoded image. The main signal comes from a vision-aided discriminator \citep{kumari2022visionaided}, a frozen DINOv2-with-registers model with a small trained head. With the encoder and decoder frozen as well, the game has one trainable network on each side. A small PatchGAN judges local patches of the render and keeps a useful gradient on materials and road surface after the vision-aided term has settled. Table~\ref{tab:t1} lists the loss weights (Appendix~\ref{app:weights}).

\subsection{Training configuration}
\label{sec:config}
The RFA is trained for 5\,000 steps at 1024\,px and batch 4 on the frozen 1.6\,B ControlNet of Section~\ref{sec:pipeline}, with learning rate 2e-4 for adapters and discriminators and 500 warmup steps. Augmentation is flips and $\pm$15$^\circ$ rotation with a corner-free crop, and Table~\ref{tab:config} gives the full configuration.
\section{Experimental setup}
\label{sec:setup}

ACDC \citep{sakaridis2021acdc} supplies fog, snow and rain as real frames, which we use unpaired for training. Its clear frame of each place, shot in another season, is the paired reference, used for instance in the MAE column and the last column of Figure~\ref{fig:f1}. River-camera haze (HIVIS) consists of USGS frames in the selection of \citet{henein2026}. BDD100K \citep{yu2020bdd100k} supplies night to day, and PreSIL \citep{hurl2019presil} with Mapillary Vistas \citep{neuhold2017mapillary} the sim-to-real pair (Appendix~\ref{app:data}).

KID \citep{binkowski2018kid} is the primary distribution metric, because FID is rank-deficient with 50 generated images against 2048-dimensional features. Both come from clean-fid (KID as the mean over its subsets), and re-scoring unchanged crops moves KID by about 0.001 on 50 crops and about 0.002 on the 25 haze crops. Sampling error over the crops is larger, and Appendix~\ref{app:se} gives jackknife standard errors. Every method is scored on one shared centre-square crop and reported at the end of its own published recipe, and every retrained baseline runs its repository's full recipe.

\section{Results}
\label{sec:results}

\subsection{Adverse-weather restoration}
\label{sec:weather}

On fog the RFA is ahead of CycleGAN-Turbo on KID and FID by more than two standard errors, and on night it is ahead on KID (Table~\ref{tab:t2}, Appendix~\ref{app:se}). On snow, rain and haze the two are level: the KID differences are within one standard error, and the RFA's lower FID on rain and haze is one to one and a half. CUT's fog KID, 0.0255 against the RFA's 0.0262, is the one distribution-metric cell where a pixel-space GAN leads. CycleGAN-Turbo's end-of-recipe fog model sits on a spike of an oscillating curve, and like every baseline it is reported at the end of its recipe (Section~\ref{sec:setup}). Cosmos-Transfer shows both failure modes of Section~\ref{sec:intro} at once, since its blur control carries the weather through and its edge control clears it by inventing the scene (Appendix~\ref{app:cosmos}). CycleGAN-Sprint, our rebuild of CycleGAN-Turbo on the Sana-Sprint backbone, is discussed in Section~\ref{sec:backbone}.

The distribution metrics do not say whether the weather is gone, and ACDC's references, shot in another season, cannot say it either. In the fog row of Figure~\ref{fig:f1} the foggy frames have bare trees and the clear ones full summer foliage, so matching the clear distribution means growing leaves, and the RFA and CycleGAN-Turbo both do. MAE against these references therefore cannot rank how well a method keeps scene structure (Section~\ref{sec:cost-of-removal}). We measure removal with a logistic classifier on pooled DINOv2-B/reg features, trained on 400 adverse and 400 clear ACDC training crops per condition and never on a test or rendered frame. It reaches held-out accuracy 1.00, and trained to split one class into random halves it stays at or below chance (0.30 to 0.46).
On rain it calls every RFA output clear, against 0.36 to 0.54 for CycleGAN-Turbo, CycleGAN-Sprint, CycleGAN and CUT and none for Cosmos-Transfer's blur control. The only other method that clears rain, Cosmos-Transfer's edge control, does so by redrawing the scene. On fog the classifier calls fewer RFA outputs clear (0.68) than CycleGAN or CUT (1.00) or CycleGAN-Turbo (0.84). This is confounded, because its clear training crops are summer frames and a method that matches them grows leaves (Appendix~\ref{app:gone}). Snow does not discriminate, since nearly every method is called clear there, and haze and night have no training crops and are not scored.

The change from input to RFA output comes from the adapter, although the decoder was trained on clear daytime frames. On fog and night the decoder-only control, the frozen decoder on the unadapted feature map (Appendix~\ref{app:bias}), scores no better than doing nothing (KID 0.1096 against 0.1066 on fog, 0.1277 against 0.1258 on night). On snow and rain it removes a fraction of what the adapter removes (snow KID 0.068 against 0.073 for doing nothing and 0.021 for the RFA, rain 0.032 against 0.042 and 0.016). The classifier calls almost none of its outputs clear (0.00, 0.00 and 0.02 on fog, snow and rain; Appendix~\ref{app:gone}).

\begin{table}[t]
\caption{Every method at the end of its own recipe, or as released (last column). CycleGAN-Sprint is CycleGAN-Turbo rebuilt on our Sana-Sprint backbone (Section~\ref{sec:backbone}). Fifty crops per condition, 25 for haze; haze and night are unpaired, so no MAE. DINO is structure distance. Bold best, underline second best.}
\label{tab:t2}
\begin{center}
\scriptsize\renewcommand{\arraystretch}{0.92}\setlength{\tabcolsep}{4pt}
\begin{tabular}{lccccc}
\toprule
\textbf{condition / method} & \textbf{KID $\downarrow$} & \textbf{FID $\downarrow$} & \textbf{DINO $\downarrow$} & \textbf{MAE $\downarrow$} & \textbf{step} \\
\midrule
\addlinespace
\multicolumn{6}{l}{\emph{ACDC fog $\rightarrow$ clear}} \\
do nothing & 0.1066 & 166.73 & 0.0000 & 0.1761 & --- \\
decoder only (identity) & 0.1096 & 169.89 & 0.0140 & 0.1834 & --- \\
RFA & \underline{0.0262} & \textbf{116.68} & 0.0213 & 0.1766 & 5000 \\
CycleGAN-Turbo & 0.0478 & 127.19 & 0.0639 & 0.1605 & 25000 \\
CycleGAN & 0.0333 & 130.73 & 0.0462 & \textbf{0.1451} & 400 ep \\
CUT & \textbf{0.0255} & \underline{120.51} & 0.0563 & \underline{0.1527} & 400 ep \\
CycleGAN-Sprint & 0.0337 & 126.68 & \underline{0.0191} & 0.1643 & 25000 \\
Cosmos-Transfer 2.5, blur & 0.1105 & 174.88 & \textbf{0.0186} & 0.1775 & released \\
Cosmos-Transfer 2.5, edge & 0.0644 & 163.51 & 0.0502 & 0.2460 & released \\
\addlinespace
\multicolumn{6}{l}{\emph{ACDC snow $\rightarrow$ clear}} \\
do nothing & 0.0729 & 147.33 & 0.0000 & 0.2051 & --- \\
decoder only (identity) & 0.0675 & 148.71 & 0.0139 & 0.2178 & --- \\
RFA & \textbf{0.0214} & \underline{119.66} & 0.0254 & 0.2060 & 5000 \\
CycleGAN-Turbo & \underline{0.0241} & \textbf{117.75} & \textbf{0.0166} & 0.2127 & 25000 \\
CycleGAN & 0.0259 & 124.09 & 0.0336 & \underline{0.1996} & 400 ep \\
CUT & 0.0637 & 161.86 & 0.0465 & 0.2486 & 400 ep \\
CycleGAN-Sprint & 0.0320 & 129.90 & 0.0256 & \textbf{0.1975} & 25000 \\
Cosmos-Transfer 2.5, blur & 0.0689 & 148.24 & \underline{0.0196} & 0.2060 & released \\
Cosmos-Transfer 2.5, edge & 0.0507 & 141.45 & 0.0264 & 0.2134 & released \\
\addlinespace
\multicolumn{6}{l}{\emph{ACDC rain $\rightarrow$ clear}} \\
do nothing & 0.0416 & 148.46 & 0.0000 & 0.1443 & --- \\
decoder only (identity) & 0.0323 & 141.57 & 0.0173 & 0.1608 & --- \\
RFA & \textbf{0.0156} & \textbf{120.80} & 0.0246 & 0.1624 & 5000 \\
CycleGAN-Turbo & \underline{0.0162} & 128.71 & \textbf{0.0124} & 0.1640 & 25000 \\
CycleGAN & 0.0231 & 137.77 & 0.0209 & 0.1646 & 400 ep \\
CUT & 0.0182 & 131.23 & 0.0253 & 0.1773 & 400 ep \\
CycleGAN-Sprint & 0.0190 & \underline{126.42} & \underline{0.0158} & \underline{0.1521} & 25000 \\
Cosmos-Transfer 2.5, blur & 0.0390 & 141.74 & 0.0221 & \textbf{0.1447} & released \\
Cosmos-Transfer 2.5, edge & 0.0243 & 133.47 & 0.0349 & 0.1885 & released \\
\addlinespace
\multicolumn{6}{l}{\emph{HIVIS haze $\rightarrow$ clear}} \\
do nothing & 0.0496 & 179.69 & 0.0000 & --- & --- \\
decoder only (identity) & 0.0798 & 212.18 & 0.0325 & --- & --- \\
RFA & 0.0379 & \textbf{165.24} & 0.0433 & --- & 5000 \\
CycleGAN-Turbo & \underline{0.0366} & 174.72 & \textbf{0.0176} & --- & released \\
CycleGAN & 0.0497 & 188.30 & 0.0444 & --- & 136 ep \\
CUT & 0.0681 & 214.19 & 0.0580 & --- & 136 ep \\
CycleGAN-Sprint & \textbf{0.0288} & \underline{169.14} & 0.0584 & --- & 25000 \\
Cosmos-Transfer 2.5, blur & 0.0673 & 228.36 & \underline{0.0333} & --- & released \\
Cosmos-Transfer 2.5, edge & 0.1186 & 299.72 & 0.1150 & --- & released \\
\addlinespace
\multicolumn{6}{l}{\emph{BDD100K night $\rightarrow$ day}} \\
do nothing & 0.1258 & 189.70 & 0.0000 & --- & --- \\
decoder only (identity) & 0.1277 & 193.55 & 0.0193 & --- & --- \\
RFA & \textbf{0.0142} & \textbf{128.12} & 0.0357 & --- & 5000 \\
CycleGAN-Turbo & 0.0229 & \underline{130.41} & \underline{0.0293} & --- & released \\
CycleGAN & 0.0270 & 141.43 & 0.0369 & --- & 114 ep \\
CUT & 0.0379 & 160.57 & 0.0382 & --- & 114 ep \\
CycleGAN-Sprint & \underline{0.0208} & 134.06 & 0.0333 & --- & 25000 \\
Cosmos-Transfer 2.5, blur & 0.1382 & 200.36 & \textbf{0.0169} & --- & released \\
Cosmos-Transfer 2.5, edge & 0.0826 & 197.84 & 0.0647 & --- & released \\
\bottomrule
\end{tabular}
\end{center}
\end{table}

\begin{figure}[t]
\centering
\setlength{\tabcolsep}{0.35pt}\renewcommand{\arraystretch}{0.45}
\begin{tabular}{@{}c@{\hspace{1pt}}ccccccccc@{}}
\rotatebox{90}{\fontsize{5.5}{6.2}\selectfont ACDC fog} & \includegraphics[width=0.106\linewidth]{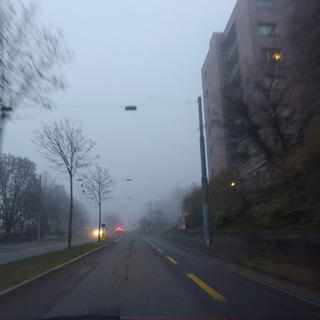} & \includegraphics[width=0.106\linewidth]{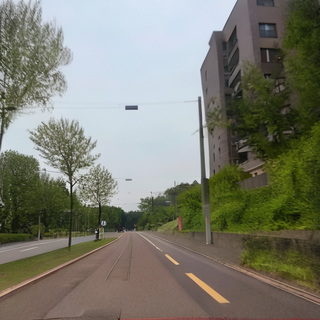} & \includegraphics[width=0.106\linewidth]{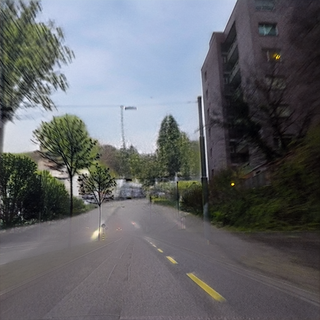} & \includegraphics[width=0.106\linewidth]{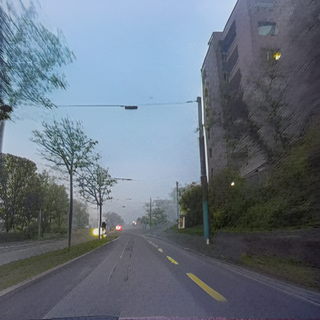} & \includegraphics[width=0.106\linewidth]{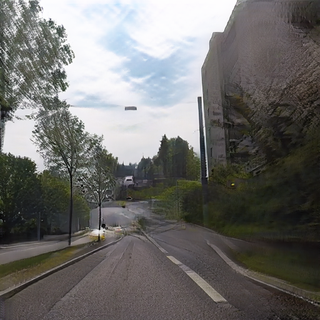} & \includegraphics[width=0.106\linewidth]{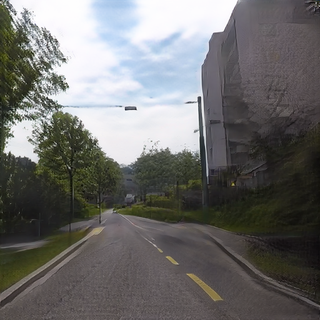} & \includegraphics[width=0.106\linewidth]{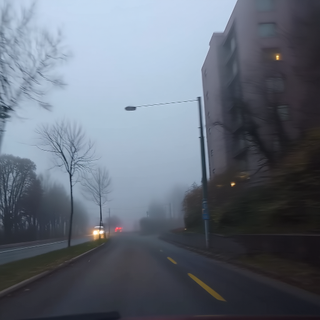} & \includegraphics[width=0.106\linewidth]{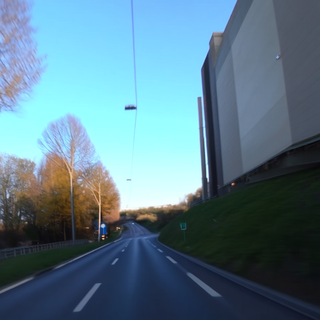} & \includegraphics[width=0.106\linewidth]{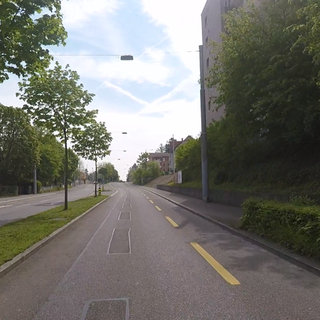} \\
\rotatebox{90}{\fontsize{5.5}{6.2}\selectfont ACDC snow} & \includegraphics[width=0.106\linewidth]{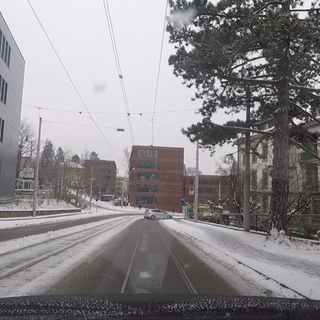} & \includegraphics[width=0.106\linewidth]{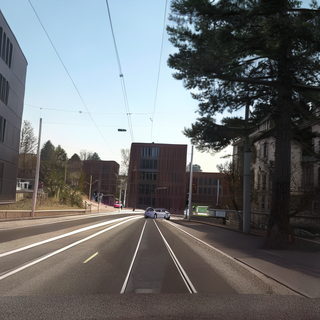} & \includegraphics[width=0.106\linewidth]{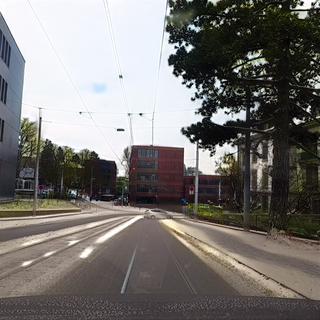} & \includegraphics[width=0.106\linewidth]{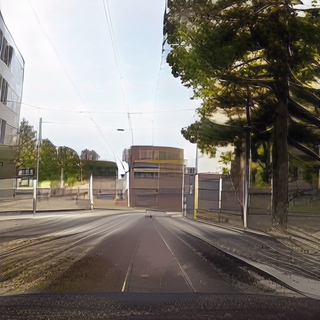} & \includegraphics[width=0.106\linewidth]{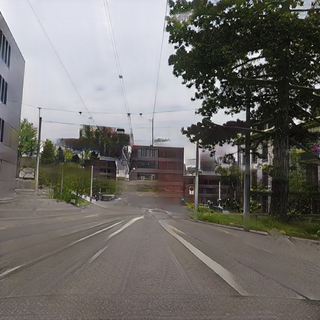} & \includegraphics[width=0.106\linewidth]{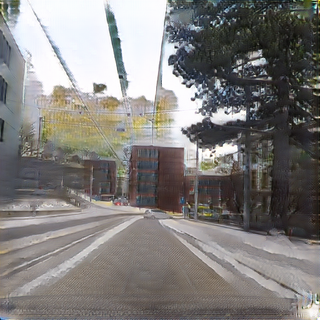} & \includegraphics[width=0.106\linewidth]{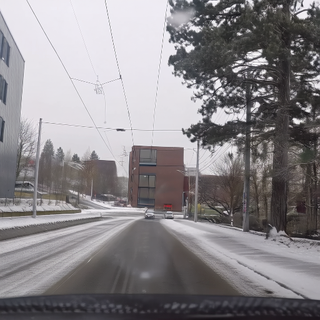} & \includegraphics[width=0.106\linewidth]{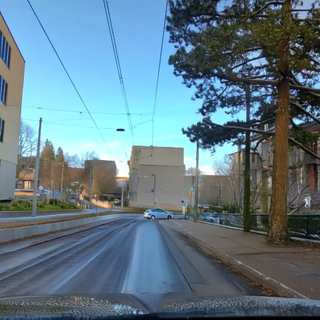} & \includegraphics[width=0.106\linewidth]{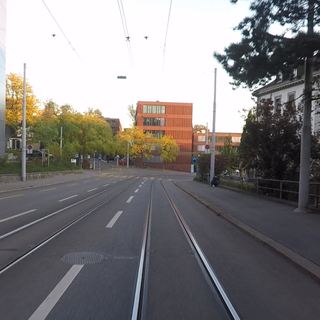} \\
\rotatebox{90}{\fontsize{5.5}{6.2}\selectfont ACDC rain} & \includegraphics[width=0.106\linewidth]{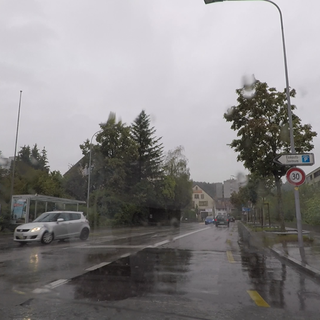} & \includegraphics[width=0.106\linewidth]{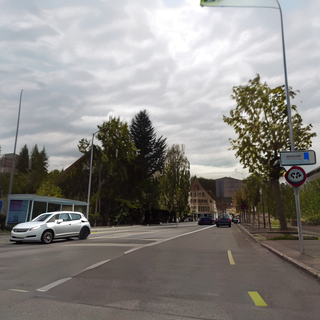} & \includegraphics[width=0.106\linewidth]{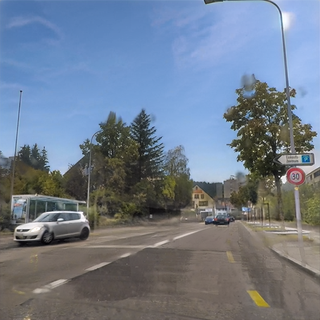} & \includegraphics[width=0.106\linewidth]{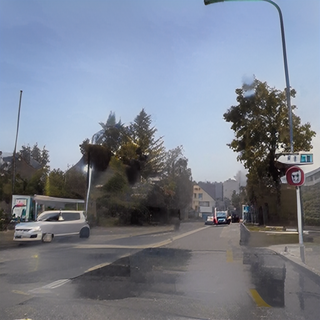} & \includegraphics[width=0.106\linewidth]{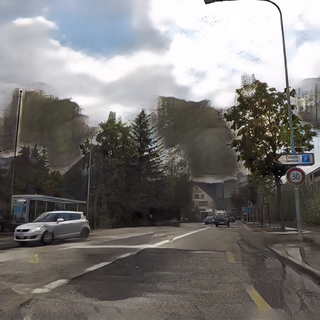} & \includegraphics[width=0.106\linewidth]{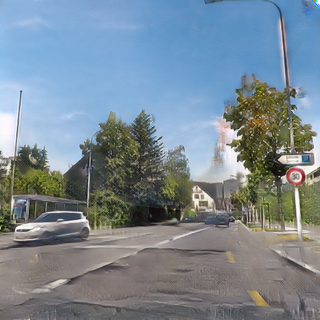} & \includegraphics[width=0.106\linewidth]{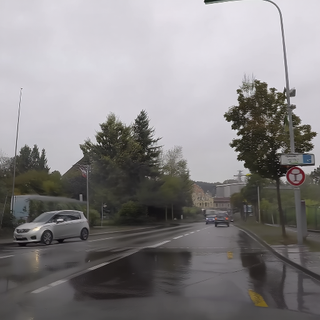} & \includegraphics[width=0.106\linewidth]{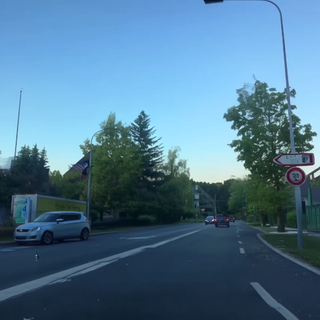} & \includegraphics[width=0.106\linewidth]{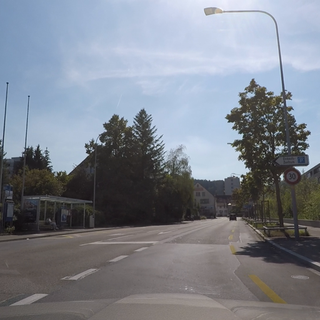} \\
\rotatebox{90}{\fontsize{5.5}{6.2}\selectfont HIVIS haze} & \includegraphics[width=0.106\linewidth]{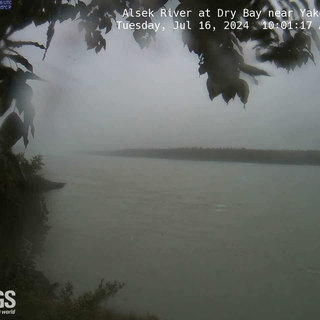} & \includegraphics[width=0.106\linewidth]{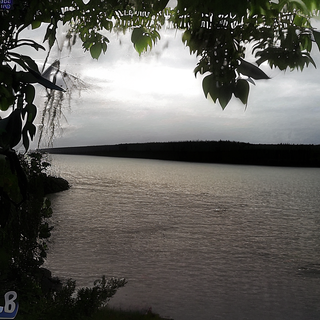} & \includegraphics[width=0.106\linewidth]{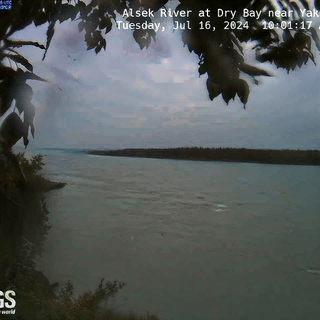} & \includegraphics[width=0.106\linewidth]{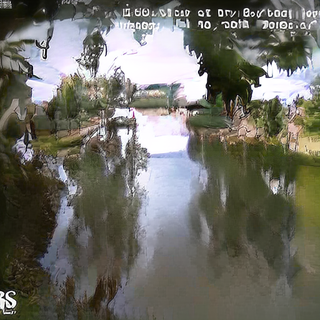} & \includegraphics[width=0.106\linewidth]{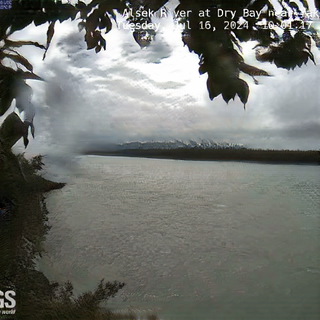} & \includegraphics[width=0.106\linewidth]{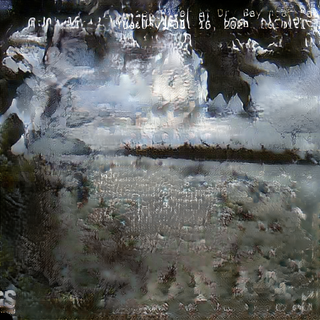} & \includegraphics[width=0.106\linewidth]{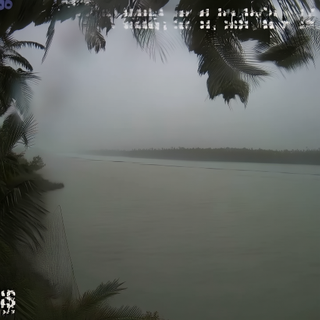} & \includegraphics[width=0.106\linewidth]{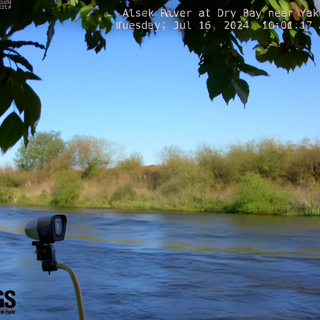} & \includegraphics[width=0.106\linewidth]{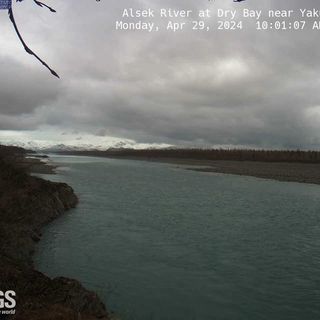} \\
\rotatebox{90}{\fontsize{5.5}{6.2}\selectfont BDD night} & \includegraphics[width=0.106\linewidth]{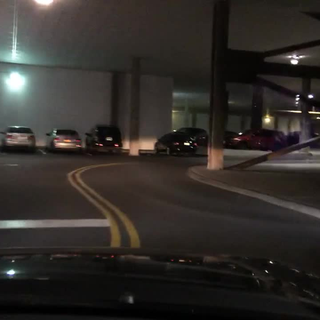} & \includegraphics[width=0.106\linewidth]{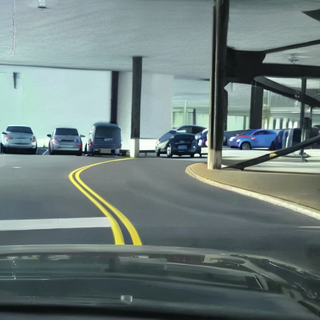} & \includegraphics[width=0.106\linewidth]{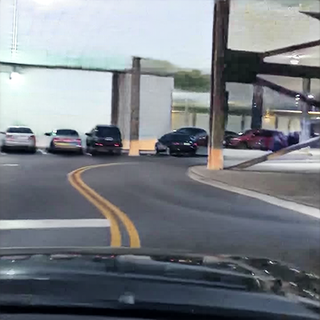} & \includegraphics[width=0.106\linewidth]{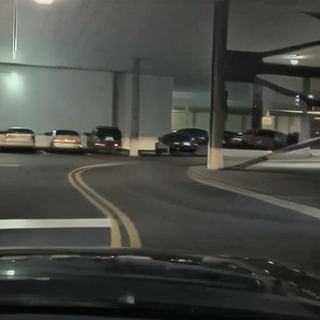} & \includegraphics[width=0.106\linewidth]{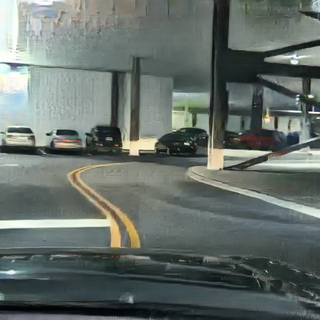} & \includegraphics[width=0.106\linewidth]{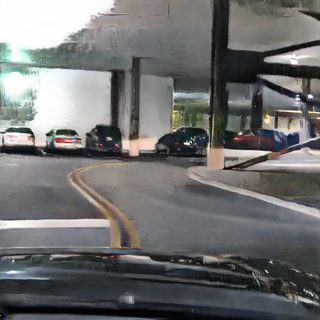} & \includegraphics[width=0.106\linewidth]{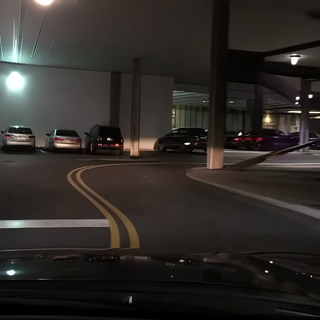} & \includegraphics[width=0.106\linewidth]{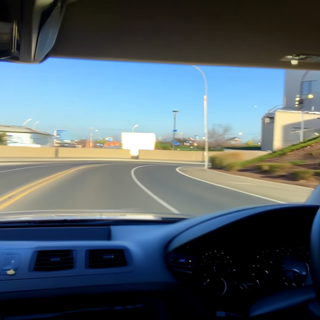} & \includegraphics[width=0.106\linewidth]{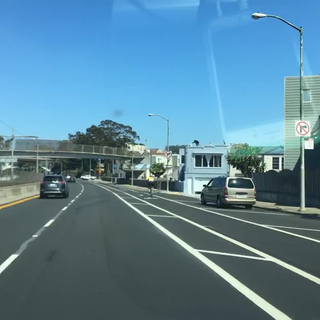} \\
 & \benchlab{input} & \benchlab{RFA (ours)} & \benchlab{CycleGAN-Turbo} & \benchlab{CycleGAN-Sprint} & \benchlab{CycleGAN} & \benchlab{CUT} & \benchlab{Cosmos, blur} & \benchlab{Cosmos, edge} & \benchlab{reference} \\
\end{tabular}
\caption{The weather benchmark (Table~\ref{tab:t2}). Last column: the paired clear frame where ACDC has one, else a target frame. On rain the wet sheen survives every pixel-fed method but Cosmos-Transfer's edge control, which redraws the scene, and only the RFA removes the rain while keeping the scene. CycleGAN-Turbo is our training with its released recipe per condition, except haze (the released module of \citet{henein2026}) and night (the released night-to-day model).}
\label{fig:f1}
\end{figure}

\subsection{Sim-to-real}
\label{sec:sim2real}

\begin{figure}[t]
\centering
\setlength{\tabcolsep}{0.8pt}\renewcommand{\arraystretch}{0.6}
\begin{tabular}{@{}cccccc@{}}
\includegraphics[width=0.196\linewidth]{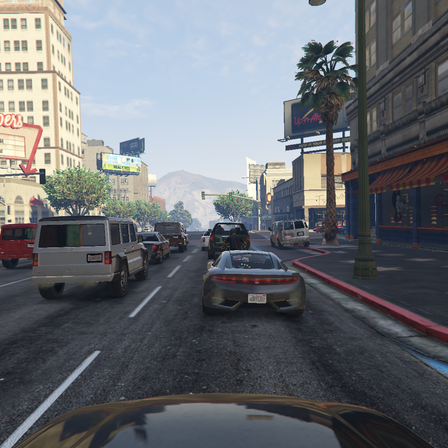} & \includegraphics[width=0.196\linewidth]{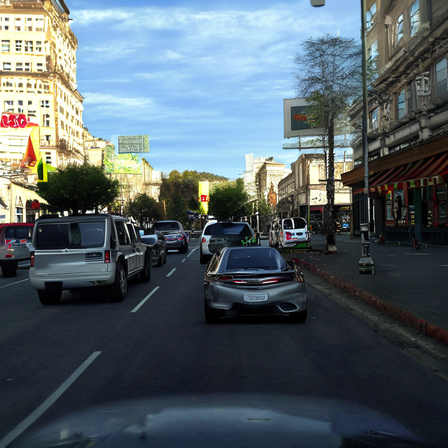} & \includegraphics[width=0.196\linewidth]{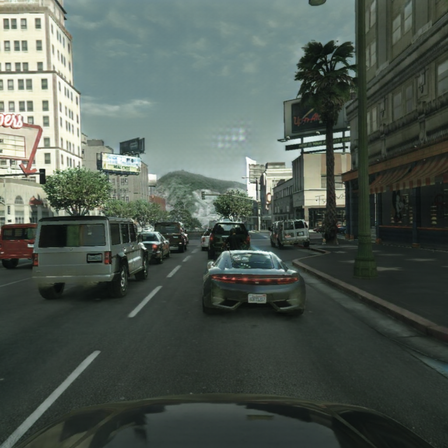} & \includegraphics[width=0.196\linewidth]{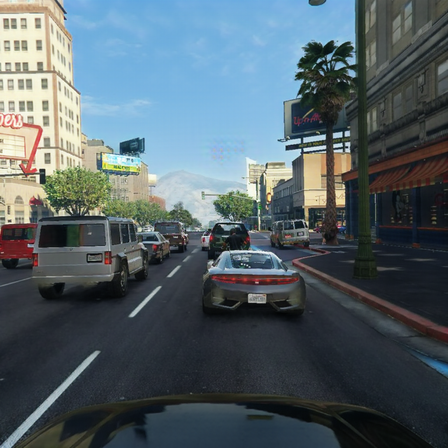} & \includegraphics[width=0.196\linewidth]{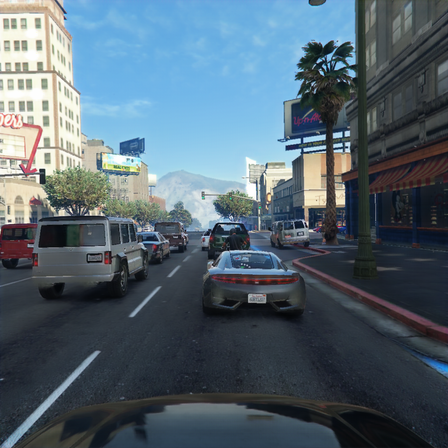} \\
\scriptsize PreSIL input & \scriptsize RFA (ours) & \scriptsize REGEN $\rightarrow$ CS & \scriptsize REGEN $\rightarrow$ Vistas & \scriptsize HyPER-GAN \\
\end{tabular}
\caption{Sim-to-real on one matched crop (Section~\ref{sec:sim2real}); more in Figure~\ref{fig:f7ext}.}
\label{fig:f7}
\end{figure}

The RFA leads every sim-to-real baseline on KID and FID (Table~\ref{tab:t3}), with a KID of 0.012 against at best 0.024.
REGEN \citep{regen2026} and HyPER-GAN \citep{hypergan2026} are the EPE line's released GTA $\rightarrow$ Vistas generators, and REGEN's GTA $\rightarrow$ Cityscapes model has another target. The RFA's structure distance is about four times that of the two Vistas generators and about three times that of REGEN's Cityscapes model. In Figure~\ref{fig:f7} the GANs keep every object in place and shift colour and texture, while the RFA keeps the cars in place, rebuilds facades and trees and replaces sign text with its own. The features fix each car's position and pose, and what they leave free is sampled by the decoder (Section~\ref{sec:portability}).

\begin{table}[t]
\caption{Sim-to-real on 100 matched PreSIL centre crops at 1024\textsuperscript{2}. KID and FID against 2\,000 Mapillary crops; DINO is structure distance to the source crop (protocol details in Appendix~\ref{app:data}). DINOv3-S row: the DINOv3-S configuration (Appendix~\ref{app:cn}). Bold best, underline second best.}
\label{tab:t3}
\begin{center}
\footnotesize
\begin{tabular}{lcccc}
\toprule
\textbf{PreSIL $\rightarrow$ Mapillary} & \textbf{KID $\downarrow$} & \textbf{FID $\downarrow$} & \textbf{DINO $\downarrow$} & \textbf{checkpoint} \\
\midrule
do nothing & 0.0470 & 135.64 & 0.0000 & --- \\
REGEN, GTA $\rightarrow$ Vistas & 0.0346 & 123.94 & \underline{0.0081} & released \\
HyPER-GAN, GTA $\rightarrow$ Vistas & 0.0331 & 125.35 & \textbf{0.0080} & released \\
REGEN, GTA $\rightarrow$ Cityscapes (other target) & \underline{0.0237} & 116.73 & 0.0110 & released \\
RFA (ours) & \textbf{0.0123} & \textbf{107.55} & 0.0318 & 5000 \\
RFA (ours, DINOv3-S, 0.6\,B decoder) & 0.0248 & 116.85 & 0.0261 & 5000 \\
\bottomrule
\end{tabular}
\end{center}
\end{table}

\subsection{What the removal costs}
\label{sec:cost-of-removal}

The lossiness that lets the adapter remove the weather also moves scene structure. What each method keeps and loses follows from what its generator is given (Figure~\ref{fig:f1}, Appendix~\ref{app:asymmetry}). CycleGAN-Turbo's generator sees a near-invertible latent of the input, so it keeps the scene and carries part of the weather through. It keeps structure better on snow, rain and haze, by 1.5 to 2.5 times on structure distance, and slightly on night (0.029 against 0.036), while on fog the RFA keeps more (0.021 against 0.064). CycleGAN-Turbo is also the only method that keeps burned-in overlay text legible, which matters for the HIVIS application.

Structure distance compares each output with its own input frame, so it sees no seasonal change, and it measures how much a method changed, including wanted change. The real clear photograph of the scene, scored against the adverse input, reaches 0.08 on snow and rain, about three times the RFA's value. A low value is therefore not the target, and the column is read with KID. A fixed COCO detector, which does not read DINO features, measures the same cost on a task (Appendix~\ref{app:det}). Its mAP on RFA outputs is about half of that on the untouched BDD100K snow and rain frames. On sim-to-real the RFA keeps about a sixth of the objects in the untranslated render, against two thirds to three quarters for REGEN and HyPER-GAN.

\subsection{Cost}
\label{sec:cost}

The adapter trains about 160 times fewer weights than CycleGAN-Turbo, in under a fifth of the per-condition training time. On one server-grade GPU it trains 2.9\,M weights, against 470\,M for CycleGAN-Turbo and 148\,M for CycleGAN-Sprint. It converges in about 1\,800 of its 5\,000 steps, where both CycleGAN variants run 25\,000, so at a nearly equal cost per step (1.79\,s against 1.86\,s) one condition takes 2.4\,h against 13.6\,h for CycleGAN-Turbo. The decoder adds a one-time 8 to 19\,h, shared by every condition and both directions, and over four conditions the total is 17.6 to 28.6\,h against 54.4\,h (Table~\ref{tab:t4}, Appendix~\ref{app:config}).

\subsection{Portability to PiD and RAE}
\label{sec:portability}

If the adapter's output is a valid feature map, decoders trained only on features of real images should render it. Our conditioning reproduces the RAE encoding that PiD is built on (Section~\ref{sec:pipeline}), so RAE \citep{zheng2026rae} and PiD \citep{lu2026pid}, two decoders from other groups, can test this.
\begin{figure}[t]
\centering
\setlength{\tabcolsep}{0.8pt}\renewcommand{\arraystretch}{0.6}
\begin{tabular}{@{}ccccc@{}}
\includegraphics[width=0.196\linewidth]{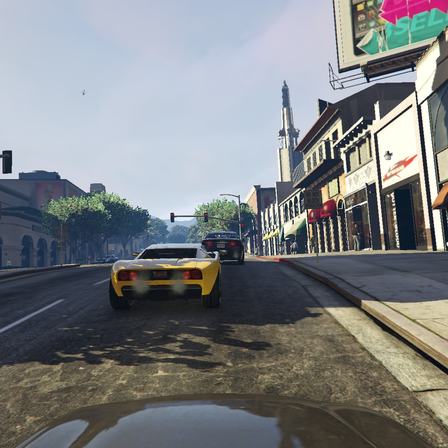} & \includegraphics[width=0.196\linewidth]{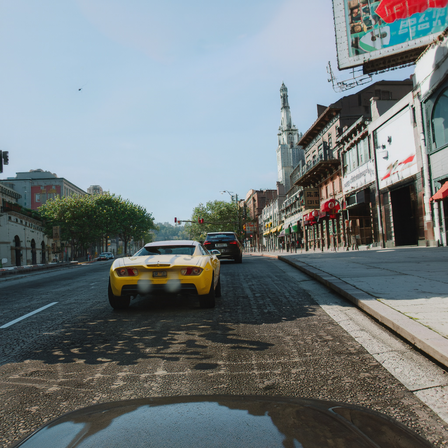} & \includegraphics[width=0.196\linewidth]{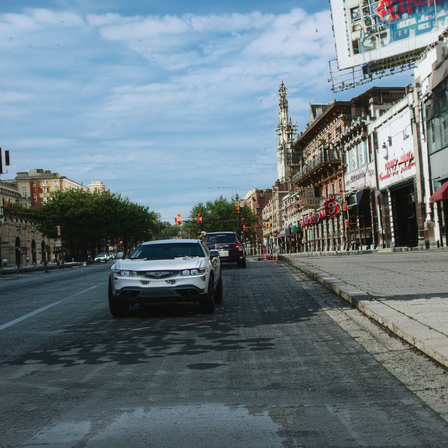} & \includegraphics[width=0.196\linewidth]{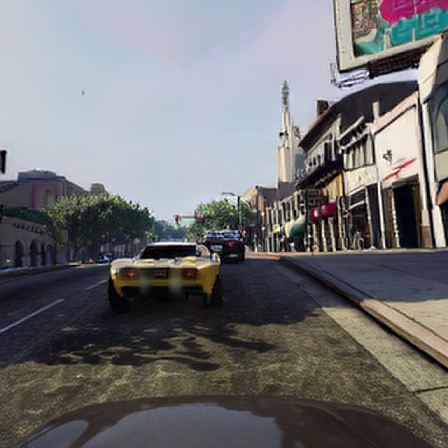} & \includegraphics[width=0.196\linewidth]{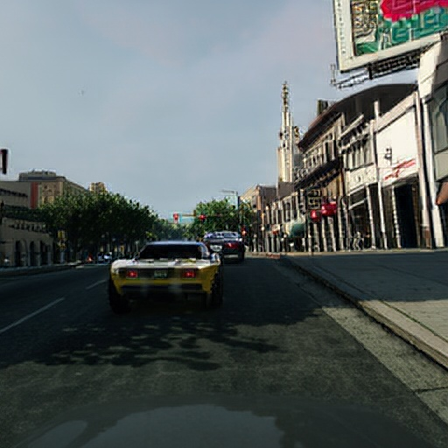} \\
\scriptsize PreSIL input & \scriptsize PiD (raw) & \scriptsize PiD (RFA) & \scriptsize RAE (raw) & \scriptsize RAE (RFA) \\
\end{tabular}
\caption{Portability. PiD (pixel diffusion, four steps) and RAE (one deterministic pass), two decoders of DINOv2 feature maps from other groups, decode the unadapted PreSIL features (``raw'') and the same features after the sim-to-real RFA. Neither decoder saw the RFA, and the RFA saw neither. Three more frames in Figure~\ref{fig:f8ext}.}
\label{fig:f8}
\end{figure}
Both decoders render the adapted maps (Figure~\ref{fig:f8}). From the unadapted features they render the game frame back, and from the adapted features they keep the scene structure and change its appearance. The lane markings, the billboard, the tower and the car's position carry through in every column. The decoders differ only where the features are silent, such as the colour of a car on some frames (Section~\ref{sec:sim2real}), which the decoder samples and the adapter does not set (Appendix~\ref{app:portability}). Two decoders that were never in the loop read these maps alike, the strongest evidence available that the adapted map is a valid point in the feature space and not an artefact tuned to one decoder.

\section{The same-backbone comparison}
\label{sec:backbone}

With the backbone held fixed, the RFA has the lower KID on fog, snow, rain and night and loses it on haze. Table~\ref{tab:t2} is a task-level comparison, and CycleGAN-Sprint, which rebuilds CycleGAN-Turbo on our Sana-Sprint backbone, is the nearest controlled one. It keeps every setting CycleGAN-Turbo's paper or code fixes, except the residual LoRA and adversarial weight the backbone forces (Appendix~\ref{app:backbone}). The port runs in this direction because our decoder recipe needs the backbone's teacher, which SD-Turbo does not publish. On fog, snow, rain and night the RFA's KID is 0.026, 0.021, 0.016 and 0.014 against 0.034, 0.032, 0.019 and 0.021. On haze CycleGAN-Sprint's 0.029 against the RFA's 0.038 is the best KID of any method, at a structure cost (0.058 against 0.043). CycleGAN-Sprint keeps more structure on fog, rain and night by 0.002 to 0.009 and is level on snow. Its fog KID of 0.034 sits between the RFA's 0.026 and CycleGAN-Turbo's 0.048, our training with its released recipe.

\section{Limitations}
\label{sec:limitations}

\textbf{Realism and scene structure trade against each other.} DINO conditioning changes materials convincingly but lets scene structure drift, and latent conditioning keeps structure but only re-grades. The trade follows from the adapter's leverage (Section~\ref{sec:why-features}) and appeared with every decoder we trained; the RGB structure loss in the recipe (Table~\ref{tab:t1}) moves the adapter along it without escaping it. We have not tried structure-consistency constraints or more adapter capacity.

\textbf{No temporal modelling.} The method is per-frame, and we have not measured temporal consistency. The adapter is deterministic, so it adds no frame-to-frame variation of its own; flicker would come from sampling the image backbone independently per frame. In preliminary runs we placed the adapter in front of a video diffusion decoder conditioned on the same DINOv2 features, built as in Control-DINO \citep{dominici2026controldino}. We saw no visible flicker, which we attribute to the video model's temporal prior, although some scene content changes over time. These runs are not part of the evaluation here.

\textbf{Fine detail is bounded by the decoder.} Small objects such as pedestrians lose their shape, and burned-in overlay text is not preserved. Part of this is our decoder's, since a decoder built for these features reconstructs better (Appendix~\ref{app:detail}), and the rest may come from the features.

\textbf{The training data has to be varied.} The adapter learns the residue from unpaired frames through the discriminators alone. When one side is small or narrow, a few dozen frames or a single scene, the discriminators can be matched by changing content instead of appearance, and the adapter then replaces scenes; we have seen this on a small out-of-domain set.

\section{Conclusion}
\label{sec:conclusion}

Unpaired translators carry the source domain through when their generator can still see it. We showed that a frozen DINO feature map holds the domain as a small residue, 13 to 14\,\% of the feature norm, and that an adapter moves it there while the identical adapter collapses on a VAE latent. A 2.9\,M-parameter RFA in front of a decoder trained once is ahead of CycleGAN-Turbo on fog and on night KID and level with it on snow, rain and haze. It leads REGEN and HyPER-GAN on sim-to-real and needs under a fifth of CycleGAN-Turbo's per-condition training time. Decoders that never saw the adapter render its output. The cost is scene structure: the lossiness that removes the weather also moves objects, and both structure distance and a fixed detector measure it. The next step is to keep structure while the residue moves, through structure-consistency constraints or decoders that reconstruct more of the scene.

\FloatBarrier   
\subsection*{AI use statement}

Large language models were used in this work as coding and writing assistants, in the sense of the ICLR policy: to write and refactor training, inference and evaluation scripts; to run and monitor cluster jobs; to give feedback on experimental design and to help interpret results as they came in; to analyse related literature; and to draft and edit the text and figures of this paper. The research question, the method, the final experimental design and every claim made here are the authors' own. All AI-assisted code was read, executed and its outputs inspected by the authors, and every number reported in this paper comes from a logged run on our own hardware, not from a model's description of one. The authors have reviewed all AI-assisted work and take responsibility for the final content, including text, claims and artifacts.

\subsection*{Ethics statement}

All experiments use publicly released datasets under their own licences: ACDC and BDD100K driving footage, PreSIL (GTA V renders), Mapillary Vistas, and river-monitoring camera frames published by the USGS, in the selection released by \citet{henein2026}. The driving datasets are used as their authors released them, with the anonymisation of faces and licence plates that they applied. We add no personal data and train no model that identifies people. The method renders images from semantic features, and, like every image-translation method, could in principle be misused to alter footage. The outputs studied here are weather changes and sim-to-real renders on public research data, and the paper reports the fidelity cost of the method as a first-class result. No human subjects were involved.

\subsection*{Reproducibility statement}

The method is specified in Section~\ref{sec:method}: the adapter architecture in
Section~\ref{sec:rfa}, the objective and its weights in Section~\ref{sec:objective}, and the
complete training configuration, including the checkpoint protocol under which
every result is reported, in Section~\ref{sec:config}. The evaluation protocol and
the datasets are in Section~\ref{sec:setup} and Appendix~\ref{app:data}. The decoder the adapter is trained
through is described in Appendix~\ref{app:cn}. All
baselines are public released models run on our own crops with the scoring code
shared across methods, and the three that we retrained, CycleGAN-Turbo on fog, snow and rain, CycleGAN and CUT, follow their own
repositories' default recipes. Source code and the trained adapters will be
released.

\bibliography{iclr2027_conference}
\bibliographystyle{iclr2027_conference}
\FloatBarrier   

\appendix

\section{Method details}
\label{app:method}

\subsection{Discriminators, loss weights and configuration}
\label{app:weights}

\paragraph{The two discriminators} The vision-aided discriminator reads its frozen DINOv2 backbone at three depths
and trains only the multi-level head on top. The second discriminator is a $70 \times 70$ PatchGAN on RGB,
trained from scratch, which sees each frame as many samples of local texture and has little to memorise. Left alone it overpowers the generator, and neither a
lower discriminator learning rate nor spectral normalisation on every convolution prevented that. An $R_1$ penalty on real images regularises its
gradient field directly, holds it in place, and is the reason the recipe carries an $R_1$ term at all. 

\textbf{Architecture.} The PatchGAN is the standard $70 \times 70$ design: four $4 \times 4$ convolutions with 64, 128, 256 and 512 channels at strides 2, 2, 2 and 1, instance normalisation with affine parameters after all but the first, LeakyReLU with slope 0.2, and a final $4 \times 4$ convolution to one logit per patch. It reads the rendered image at its full 1024\,px, is trained with the least-squares GAN loss against a pool of the last 50 renders, and receives the $R_1$ penalty on real images. The vision-aided discriminator is DINOv2-B with registers, frozen, fed the render at 224\,px, tapped at blocks 3 and 6 (patch maps) and at the last block ([CLS]). The trained head is the multi-level head of \citet{kumari2022visionaided}, a spectrally normalised $3 \times 3$ convolution to 256 channels, LeakyReLU, blur pooling and a $1 \times 1$ convolution to one logit on each patch map, and a linear 768 $\to$ 256 $\to$ 1 head on the [CLS] token, trained with their multi-level sigmoid loss. The PatchGAN earns its place under feature conditioning only: without it, colour drift under DINO conditioning roughly triples and structure degrades, while under latent conditioning, which carries colour itself, removing it costs nothing and structure improves.

\begin{table}[t]
\caption{Loss weights.}
\label{tab:t1}
\begin{center}
\footnotesize
\begin{tabular}{lcc}
\toprule
\textbf{term} & \textbf{weight} & \textbf{role} \\
\midrule
cycle & 30.0 & F(G(x)) $\approx$ x and G(F(y)) $\approx$ y \\
identity & 10.0 & G(y) $\approx$ y for y already in target; the direct read on drift \\
structure & 20.0 & patch self-similarity; holds scene geometry \\
image GAN & 1.0 & RGB PatchGAN on the decoded image \\
vision-aided & 0.2 & frozen DINOv2 features with a small trained head \citep{kumari2022visionaided} \\
R1 penalty & 0.001 & gradient penalty on real images \\
\bottomrule
\end{tabular}
\end{center}
\end{table}

\label{app:config}

\begin{table}[t]
\caption{Training configuration of every RFA.}
\label{tab:config}
\begin{center}\small
\begin{tabular}{@{}ll@{}}
\toprule
backbone & Sana-Sprint 1.6\,B, 1024\,px, bf16, \textbf{frozen} \\
decoder & ControlNet on DINOv2-B/reg (PiD normalisation, 448\,px, $32\times32$ map), Mapillary, \\
 & 7 blocks, checkpoint 15\,000, \textbf{frozen}; DINOv3-S configuration: 0.6\,B at 1024\,px \\
optimiser & lr 2e-4 ($G$ and $D$), constant with warmup, 500 warmup steps \\
schedule & 5\,000 steps, batch 4, gradient checkpointing \\
discriminator & vision-aided DINOv2, $R_1$ 0.001 \\
augment & horizontal/vertical flip, $\pm15^\circ$ rotation with a black-corner-free crop \\
\bottomrule
\end{tabular}
\end{center}
\end{table}

\begin{table}[t]
\caption{Training cost on a single server-grade GPU. CycleGAN-Turbo as released (SD-Turbo, 512\,px) and CycleGAN-Sprint (Sana-Sprint, rank 128, 512\,px, batch 1; Appendix~\ref{app:backbone}), its 3.7\,M CLIP discriminator heads excluded as in the original; the RFA at 1024\,px, batch 4. Per-condition times are wall-clock. Per-step cost is of the same order everywhere; the RFA's saving is the step count, and it pays once for the decoder (8 to 19\,h), which every condition and both directions share.}
\label{tab:t4}
\begin{center}
\footnotesize
\setlength{\tabcolsep}{4pt}
\begin{tabular}{lccc}
\toprule
 & \textbf{CycleGAN-Turbo} & \textbf{CycleGAN-Sprint} & \textbf{RFA} \\
\midrule
trainable weights & 470\,M (LoRA r\,=\,128) & 148\,M (LoRA r\,=\,128) & \textbf{2.9\,M} (G\,+\,F) \\
steps & 25\,000 & 25\,000 & \textbf{5\,000} ($\approx$\,1\,800 to converge) \\
per step & 1.86\,s & 2.40\,s & 1.79\,s \\
per condition & 13.6\,h & 16.7\,h & \textbf{2.4\,h} \\
one-time prerequisite & none & none & decoder, 8--19\,h \\
all four conditions & 54.4\,h & 66.7\,h & \textbf{17.6--28.6\,h} \\
\bottomrule
\end{tabular}
\end{center}
\end{table}

\section{Experimental protocol}
\label{app:protocol}

\subsection{Datasets}
\label{app:data}

ACDC \citep{sakaridis2021acdc} supplies fog, snow and rain (400 adverse and 900 clear frames per condition, with a seasonally offset clear correspondence). ACDC night is not used, because its shift is a lighting change and not an additive veil over the scene, and the night condition comes from BDD100K instead (1\,664 day and 3\,174 night training frames, 50 held-out frames per side, unpaired). Dehazing uses river-monitoring camera frames from the USGS Hydrologic Imagery Visualization and Information System, in the selection released by \citet{henein2026} (88 unique hazy frames against 6 355 haze-free, unpaired, after removing the 37 hazy rows their listing duplicates). The haze baseline is its released dehazing checkpoint, a CycleGAN-Turbo finetuned on HIVIS, run with the authors' released inference settings (no resize, hazy-to-clear direction, prompt ``a clean image''). Its output reproduces the published example frame pixel for pixel (MAE 0.018, correlation 1.000), and on their full-frame protocol it reaches FID 155.2 against the published 153.3. BDD100K supplies night$\rightarrow$day, and PreSIL (46 431 training and 4 644 validation frames at 1920 $\times$ 1080) and Mapillary Vistas supply the sim-to-real pair. The sim-to-real baselines are scored on the same 100 centre crops; HyPER-GAN's own path squashes frames to 544 $\times$ 960, and both baselines pass a pixel-level self-test against their released outputs. CycleGAN and CUT run the CUT repository's default 200 plus 200 epochs on ACDC; haze and night keep that run's 360\,000 iterations, which is 136 and 114 epochs on their larger training sides, and CycleGAN-Turbo runs its documented 25\,000 steps.

\section{Further results}
\label{app:results}

\subsection{Why feature conditioning gives the adapter leverage: three further bounds}
\label{app:mechanism}

\paragraph{The residue, measured}\label{app:residue} Two numbers make the title's claim literal (Table~\ref{tab:residue}). The residue as defined in Section~\ref{sec:why-features} is 13 to 14\,\% of the feature norm in every condition, with a per-frame spread of one to five points, and the same adapter fed a clear frame moves it by 3 to 6\,\%, close to the identity where there is nothing to remove. For scale, any other photograph sits about a full norm away in the same space, whatever its weather: the real clear photograph of the same place shot in another season at 106 to 118\,\%, an unrelated frame of the same weather at 101 to 121\,\%, an unrelated clear frame at 121 to 124\,\%, each an order of magnitude beyond the adapter's move. That distance is a scene difference, not a weather difference, and the residue is the part the adapter rewrites within it. The measure depends on the feature space's scale: the DINOv3-S configuration's adapters on the DINOv3-S map at 1024\,px (384 channels) move it by 7.2, 5.4, 5.4, 5.9 and 9.3\,\% on fog, snow, rain, haze and night, where the distance between two frames of the same weather is 0.46 to 0.61 instead of 1.01 to 1.21, so relative to that distance the residue is 11 to 14\,\% on DINOv2-B/reg and 10 to 16\,\% on DINOv3-S.

\begin{table}[t]
\caption{The residue in numbers, the adapters and test crops of Table~\ref{tab:t2}: $\|G(f)-f\|/\|f\|$ per frame (mean and standard deviation) on adverse input and on clear input, and, in the same units, the distance from the adverse input to the paired clear photograph of the scene (paired conditions) and to an unrelated frame of the same test set, of the same weather and clear.}
\label{tab:residue}
\begin{center}
\footnotesize
\setlength{\tabcolsep}{5pt}
\begin{tabular}{lcccccc}
\toprule
& & \multicolumn{2}{c}{\textbf{adapter move}} & \multicolumn{3}{c}{\textbf{distance to another frame}} \\
\cmidrule(lr){3-4}\cmidrule(lr){5-7}
\textbf{condition} & \textbf{n} & \textbf{adverse input} & \textbf{clear input} & \textbf{paired clear} & \textbf{same weather} & \textbf{clear} \\
\midrule
ACDC fog & 50 & 0.141 $\pm$ 0.012 & 0.027 & 1.18 & 1.01 & 1.22 \\
ACDC snow & 50 & 0.136 $\pm$ 0.010 & 0.026 & 1.16 & 1.05 & 1.23 \\
ACDC rain & 50 & 0.132 $\pm$ 0.029 & 0.027 & 1.06 & 1.14 & 1.21 \\
HIVIS haze & 25 & 0.131 $\pm$ 0.049 & 0.059 & --- & 1.21 & 1.24 \\
BDD night & 50 & 0.128 $\pm$ 0.023 & 0.036 & --- & 1.16 & 1.22 \\
\bottomrule
\end{tabular}
\end{center}
\end{table}

\paragraph{Where the lossiness sits, and the detail ceiling}\label{app:detail} The lossiness above is measured through our decoder, and RAE \citep{zheng2026rae} and VFM-VAE \citep{bi2026vfmvae} reconstruct well from frozen self-supervised features once the decoder is designed for it. Two measurements locate it. On the same held-out frames and the same DINOv2 map, RAE's decoder reconstructs about 2\,dB better than our ControlNet and reproduces sign text and window detail ours does not (Table~\ref{tab:s5}), so that part of the loss is the decoder's. A linear map from the features to the backbone's own DC-AE latent explains only a third of that latent's variance, while our ControlNet reconstructs better than that linear map, so the rest is not a merge problem but a property of what the features carry. Reconstruction MAE understates it, since it is dominated by large smooth regions while the failures are in high-frequency content, and no conditioning-side change we tried recovers it. This is the ceiling stated in Section~\ref{sec:limitations}. A better decoder moves that ceiling: RAE decodes the same adapted map without retraining (Section~\ref{sec:portability}) and reconstructs it 2\,dB better, and an adapter trained through the almost lossless 23-layer RAEv2 decoder keeps far more structure at a distribution cost (Appendix~\ref{app:s6}).

A second bound, from the other end of the pipeline. Measuring patch Fr\'echet distance between sim and real, the gap shrinks by more than an order of magnitude after the trained ControlNet's projection into its low-dimensional control space, so the projection is far more domain-invariant than the features feeding it. An adapter trained with a feature-space discriminator alone, no decoder in the loop, moves the feature statistics a good way toward the real domain (Fr\'echet distance to the real features from 4.8 to about 3.1, content cosine 0.95) or never leaves the identity, and in the first case its renders through the frozen ControlNet change little beyond road tone and material, the failure Section~\ref{sec:pipeline} states. This bounds how much visible change a feature-space adapter can produce for a given decoder, and it is why the adapter's wins in Table~\ref{tab:t2} are distributional and not dramatic.

\subsection{Cosmos-Transfer, condition by condition}
\label{app:cosmos}

Cosmos-Transfer 2.5 is the released 2\,B model with no training on our data, run per frame in the image-to-image mode its documentation describes, with CycleGAN-Turbo's target prompt on ACDC, a river-camera prompt on HIVIS and the day prompt on BDD100K. Driven by a blur map of the foggy frame it returns the foggy frame, no better than doing nothing on KID, because the blur map carries the fog. Driven by an edge map it produces a clean, sunlit street on the input's edges and invents everything the edges do not fix, facades and sky included, at the worst MAE in the table. Snow and rain repeat this: the blur map hands the snow banks and the wet road back almost unchanged, and the edge map returns a sunlit street with the weather gone and the kerbs and facades redrawn. On HIVIS haze, where the input has almost no edges to give, the edge map returns a sunny river bank with the prompt's monitoring camera painted into the foreground, at the worst KID in the table. Pixel control that keeps appearance keeps the weather, pixel control that drops appearance drops the scene with it, since a geometric signal captures only a narrow aspect of the scene \citep{dominici2026controldino}, and a DINO feature map sits between the two, which is what the RFA row shows.

\subsection{Why the two failure modes are not symmetric}
\label{app:asymmetry}

CycleGAN-Turbo's generator reads a near-invertible latent of the input, so pixel-level weather rides through it into the output: in Figure~\ref{fig:f1} the restored rain frame keeps the wet-road sheen and its reflections and the restored snow frame keeps the banks along the kerb. The adapter's only input is the DINO feature map, which does not encode that residue, so the frozen decoder renders dry asphalt and bare kerbs from the same frames, and the structure it loses (structure distance, Table~\ref{tab:t2}) is the price of the same lossiness. This is a trade-off, not a verdict, and Driving with DINO and HyPER-GAN's comparison with REGEN report the same one, distribution against structure. It appears to be a property of the problem, not of any one architecture.

\subsection{Weather benchmark, more examples}
\label{app:benchfig}

\begin{figure}[p]
\centering
\setlength{\tabcolsep}{0.35pt}\renewcommand{\arraystretch}{0.45}
\begin{tabular}{@{}c@{\hskip 2pt}ccccccccc@{}}
\rotatebox{90}{\fontsize{5.5}{6.2}\selectfont ACDC fog} & \includegraphics[width=0.106\linewidth]{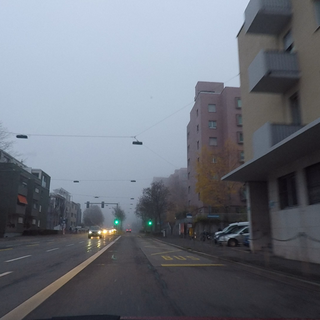} & \includegraphics[width=0.106\linewidth]{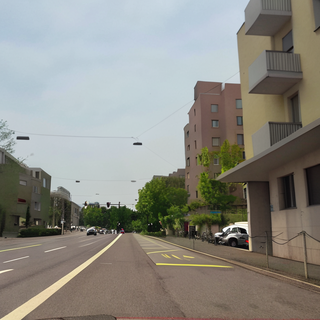} & \includegraphics[width=0.106\linewidth]{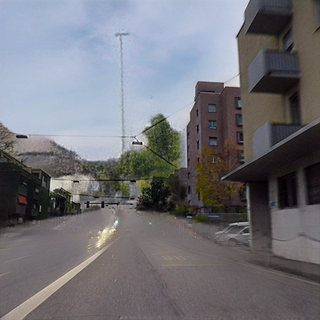} & \includegraphics[width=0.106\linewidth]{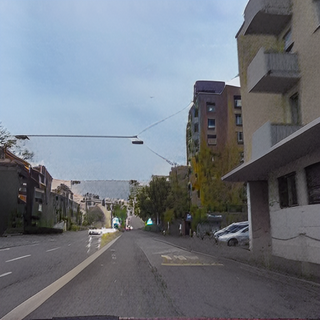} & \includegraphics[width=0.106\linewidth]{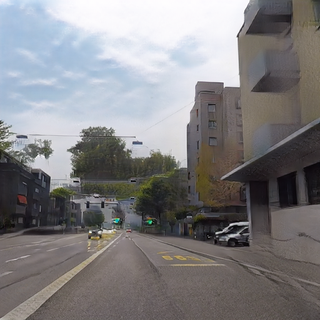} & \includegraphics[width=0.106\linewidth]{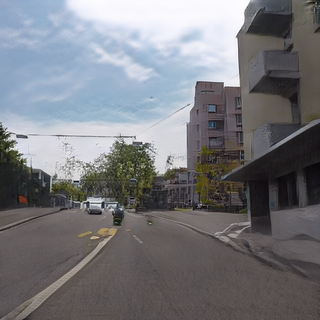} & \includegraphics[width=0.106\linewidth]{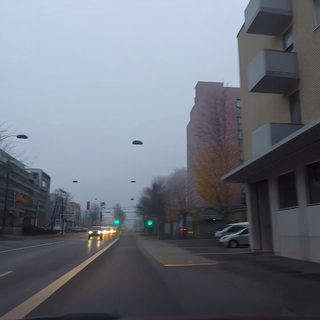} & \includegraphics[width=0.106\linewidth]{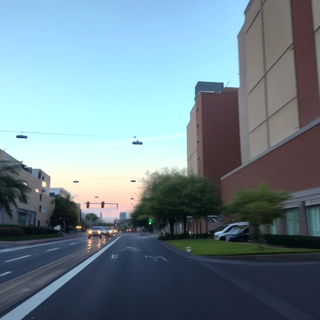} & \includegraphics[width=0.106\linewidth]{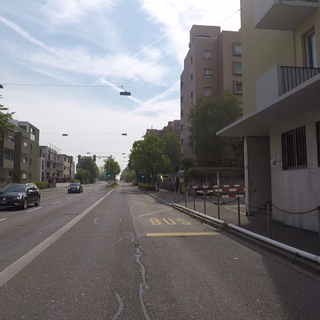} \\
\rotatebox{90}{\fontsize{5.5}{6.2}\selectfont ACDC fog} & \includegraphics[width=0.106\linewidth]{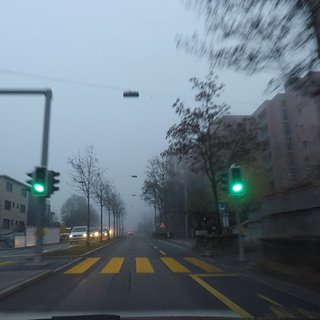} & \includegraphics[width=0.106\linewidth]{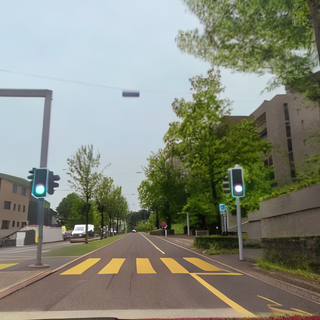} & \includegraphics[width=0.106\linewidth]{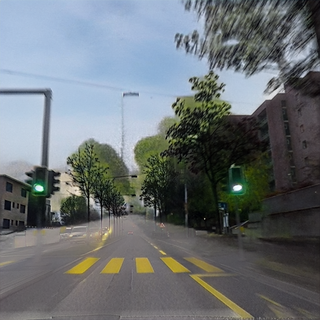} & \includegraphics[width=0.106\linewidth]{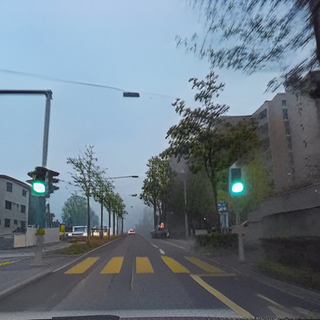} & \includegraphics[width=0.106\linewidth]{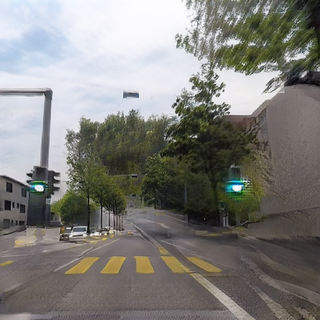} & \includegraphics[width=0.106\linewidth]{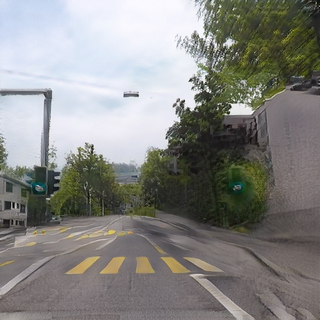} & \includegraphics[width=0.106\linewidth]{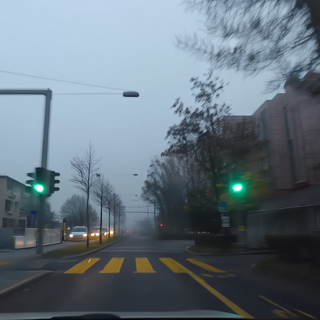} & \includegraphics[width=0.106\linewidth]{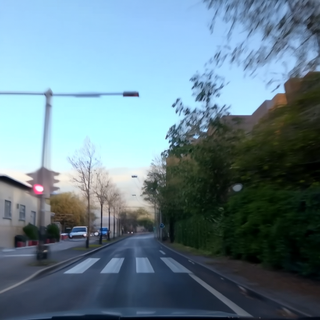} & \includegraphics[width=0.106\linewidth]{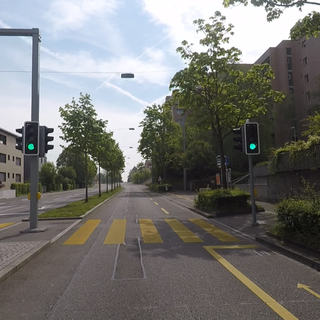} \\
\rotatebox{90}{\fontsize{5.5}{6.2}\selectfont ACDC snow} & \includegraphics[width=0.106\linewidth]{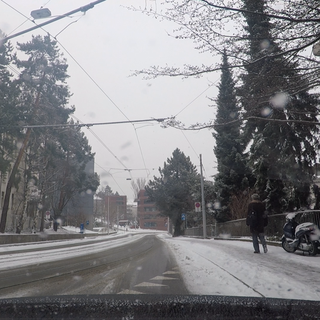} & \includegraphics[width=0.106\linewidth]{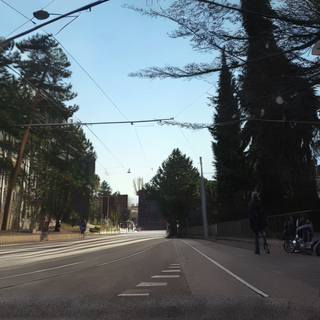} & \includegraphics[width=0.106\linewidth]{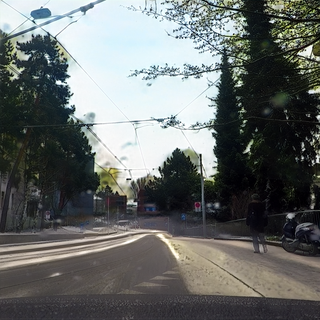} & \includegraphics[width=0.106\linewidth]{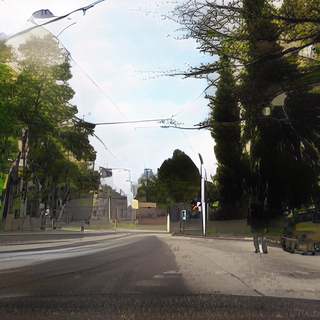} & \includegraphics[width=0.106\linewidth]{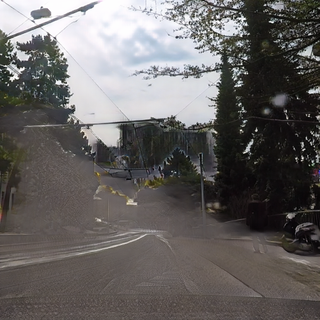} & \includegraphics[width=0.106\linewidth]{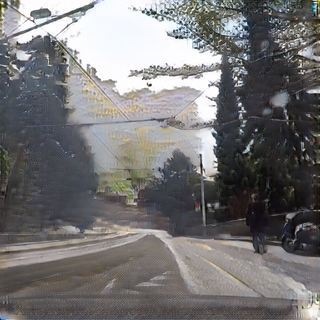} & \includegraphics[width=0.106\linewidth]{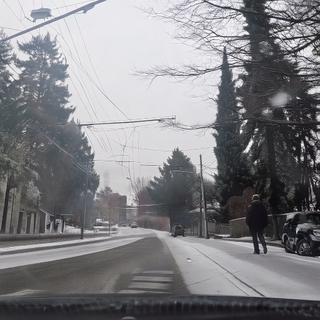} & \includegraphics[width=0.106\linewidth]{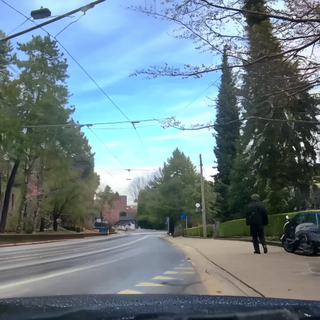} & \includegraphics[width=0.106\linewidth]{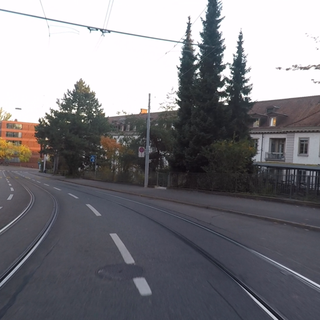} \\
\rotatebox{90}{\fontsize{5.5}{6.2}\selectfont ACDC snow} & \includegraphics[width=0.106\linewidth]{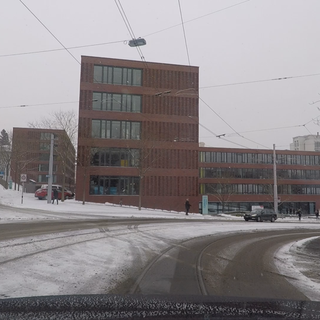} & \includegraphics[width=0.106\linewidth]{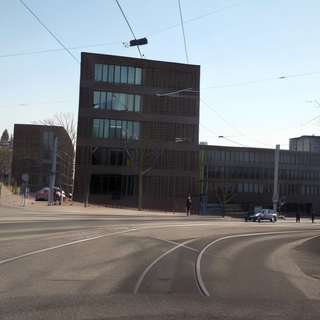} & \includegraphics[width=0.106\linewidth]{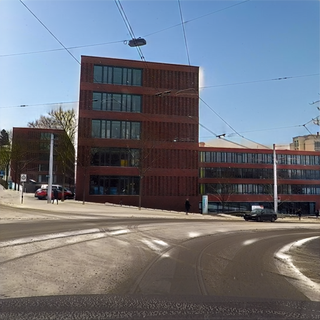} & \includegraphics[width=0.106\linewidth]{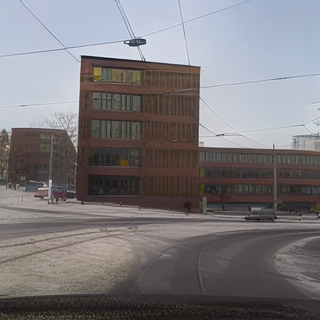} & \includegraphics[width=0.106\linewidth]{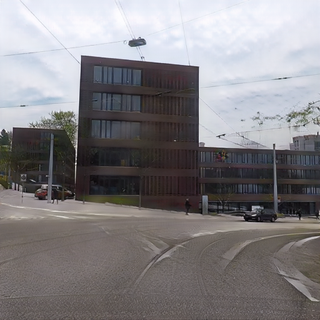} & \includegraphics[width=0.106\linewidth]{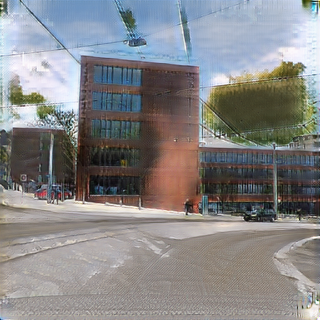} & \includegraphics[width=0.106\linewidth]{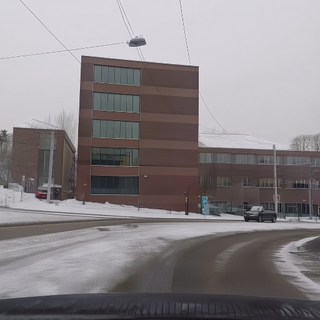} & \includegraphics[width=0.106\linewidth]{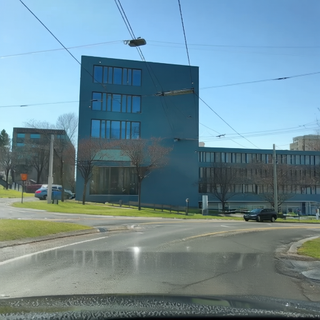} & \includegraphics[width=0.106\linewidth]{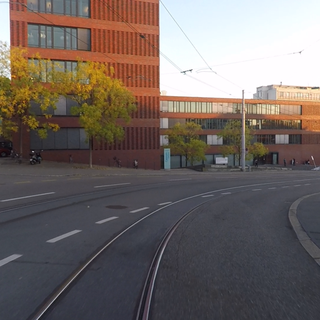} \\
\rotatebox{90}{\fontsize{5.5}{6.2}\selectfont ACDC rain} & \includegraphics[width=0.106\linewidth]{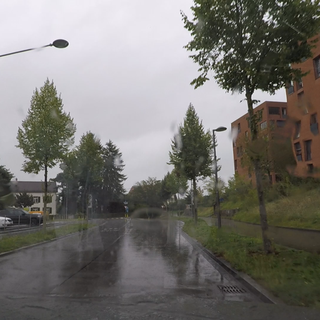} & \includegraphics[width=0.106\linewidth]{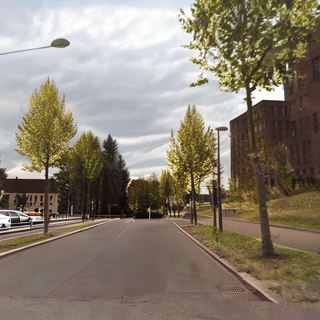} & \includegraphics[width=0.106\linewidth]{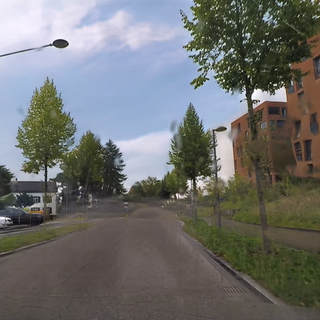} & \includegraphics[width=0.106\linewidth]{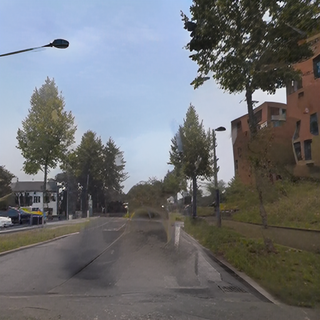} & \includegraphics[width=0.106\linewidth]{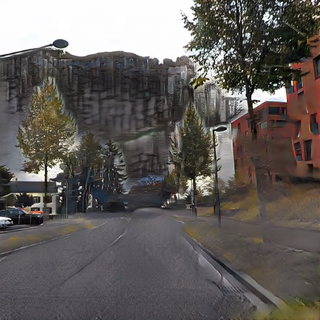} & \includegraphics[width=0.106\linewidth]{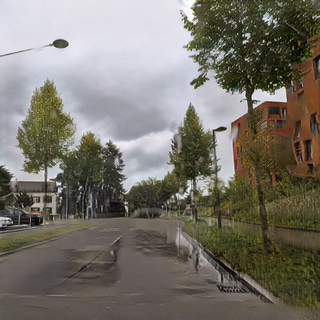} & \includegraphics[width=0.106\linewidth]{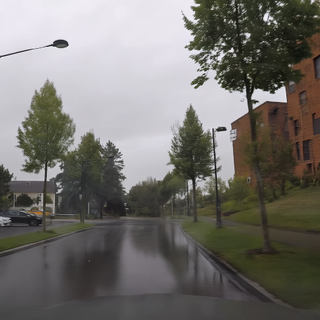} & \includegraphics[width=0.106\linewidth]{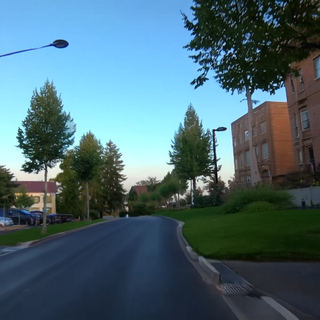} & \includegraphics[width=0.106\linewidth]{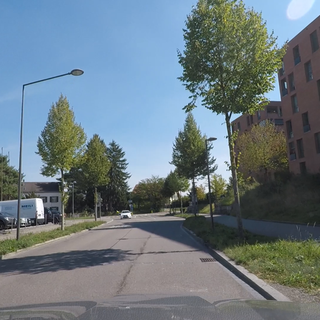} \\
\rotatebox{90}{\fontsize{5.5}{6.2}\selectfont ACDC rain} & \includegraphics[width=0.106\linewidth]{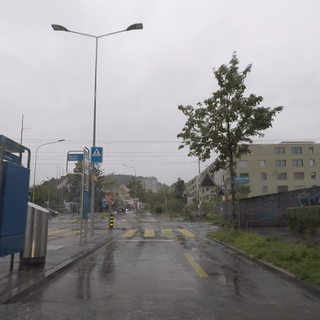} & \includegraphics[width=0.106\linewidth]{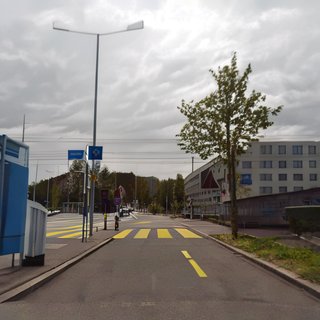} & \includegraphics[width=0.106\linewidth]{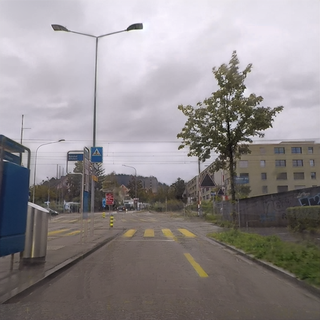} & \includegraphics[width=0.106\linewidth]{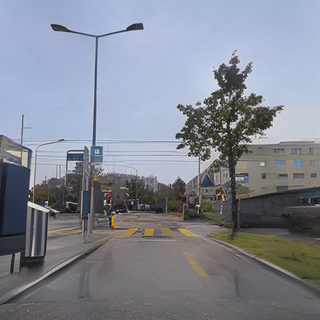} & \includegraphics[width=0.106\linewidth]{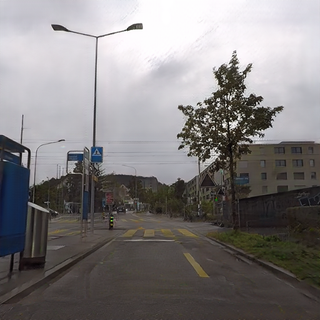} & \includegraphics[width=0.106\linewidth]{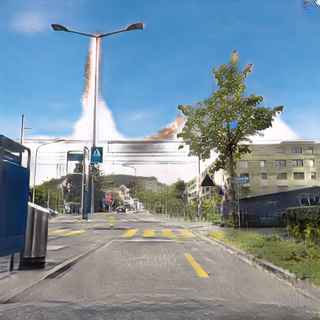} & \includegraphics[width=0.106\linewidth]{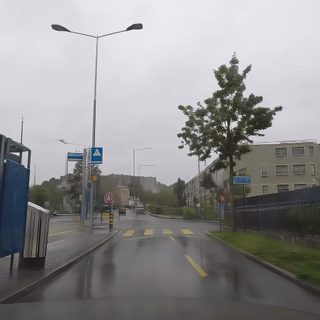} & \includegraphics[width=0.106\linewidth]{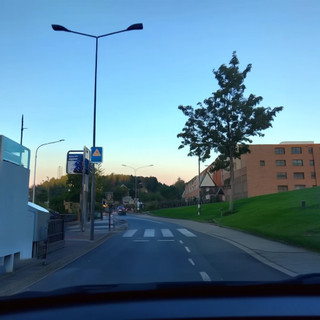} & \includegraphics[width=0.106\linewidth]{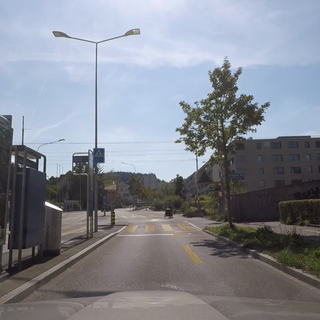} \\
\rotatebox{90}{\fontsize{5.5}{6.2}\selectfont HIVIS haze} & \includegraphics[width=0.106\linewidth]{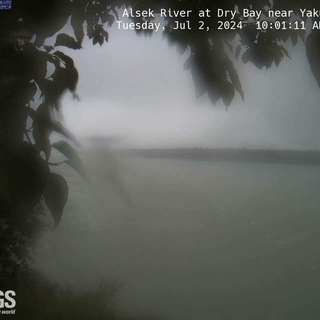} & \includegraphics[width=0.106\linewidth]{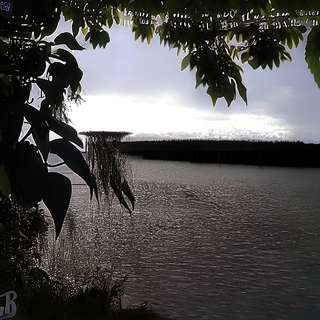} & \includegraphics[width=0.106\linewidth]{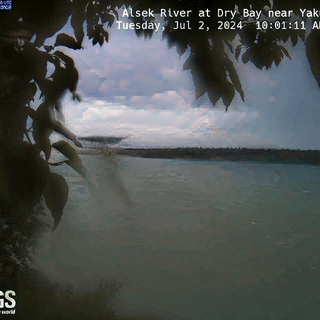} & \includegraphics[width=0.106\linewidth]{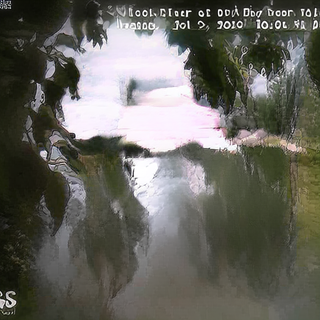} & \includegraphics[width=0.106\linewidth]{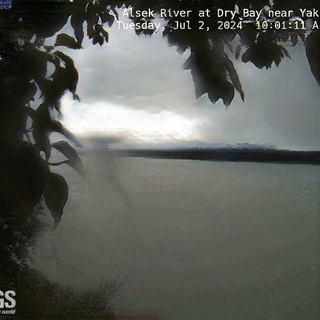} & \includegraphics[width=0.106\linewidth]{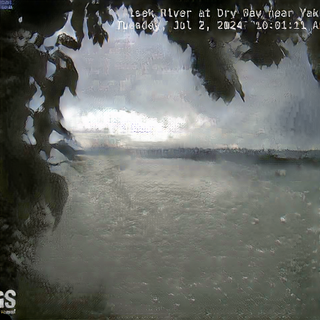} & \includegraphics[width=0.106\linewidth]{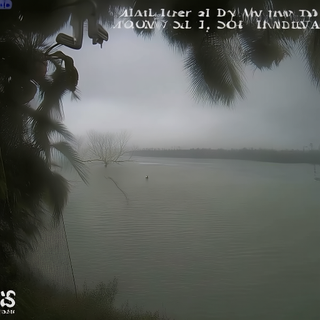} & \includegraphics[width=0.106\linewidth]{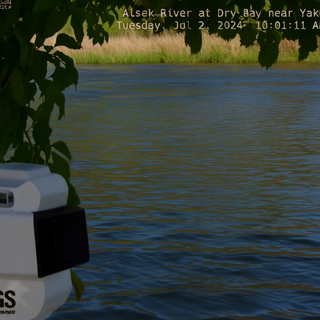} & \includegraphics[width=0.106\linewidth]{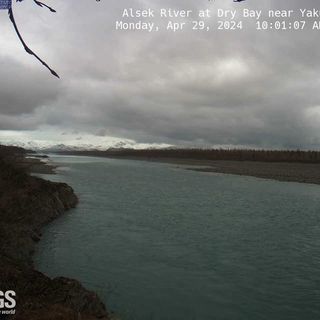} \\
\rotatebox{90}{\fontsize{5.5}{6.2}\selectfont HIVIS haze} & \includegraphics[width=0.106\linewidth]{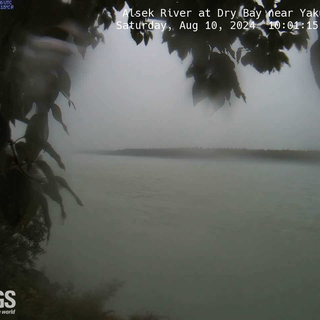} & \includegraphics[width=0.106\linewidth]{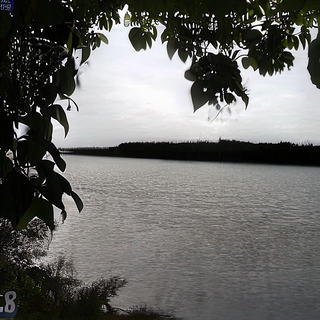} & \includegraphics[width=0.106\linewidth]{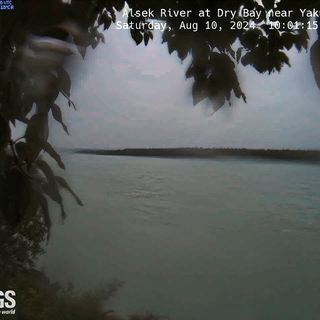} & \includegraphics[width=0.106\linewidth]{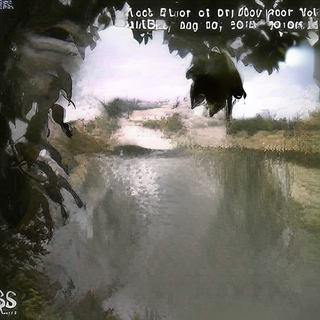} & \includegraphics[width=0.106\linewidth]{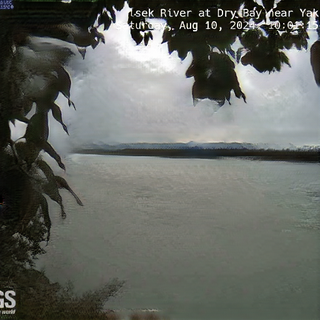} & \includegraphics[width=0.106\linewidth]{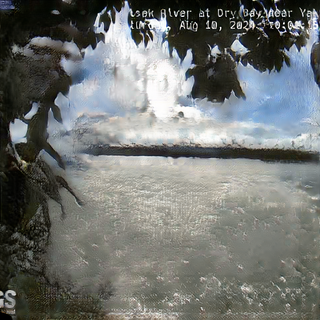} & \includegraphics[width=0.106\linewidth]{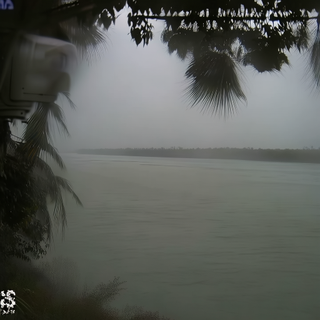} & \includegraphics[width=0.106\linewidth]{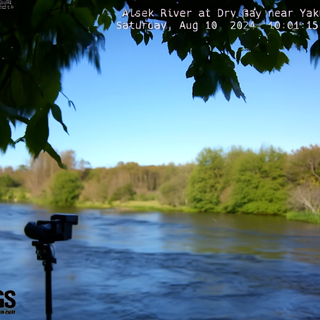} & \includegraphics[width=0.106\linewidth]{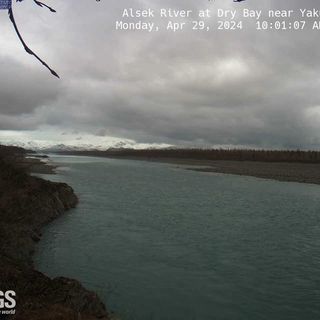} \\
\rotatebox{90}{\fontsize{5.5}{6.2}\selectfont BDD night} & \includegraphics[width=0.106\linewidth]{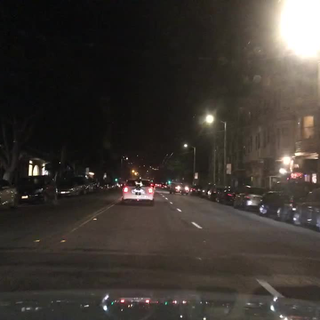} & \includegraphics[width=0.106\linewidth]{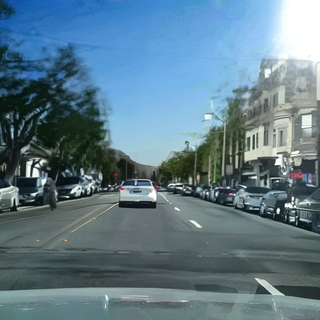} & \includegraphics[width=0.106\linewidth]{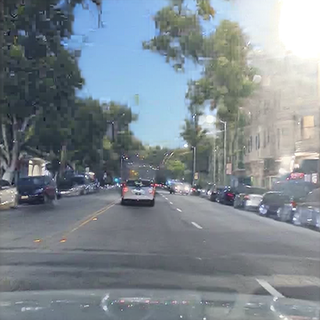} & \includegraphics[width=0.106\linewidth]{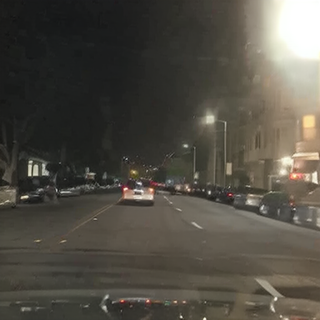} & \includegraphics[width=0.106\linewidth]{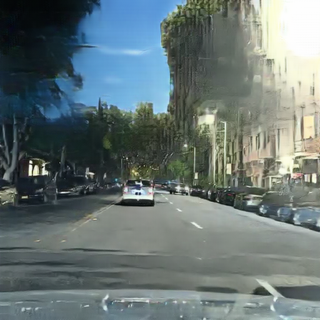} & \includegraphics[width=0.106\linewidth]{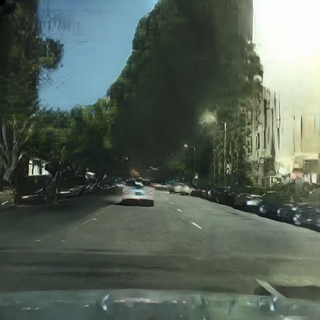} & \includegraphics[width=0.106\linewidth]{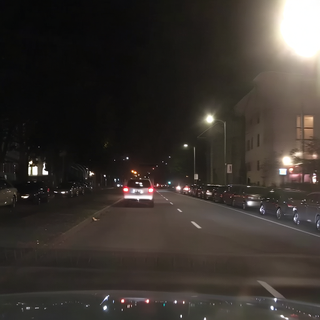} & \includegraphics[width=0.106\linewidth]{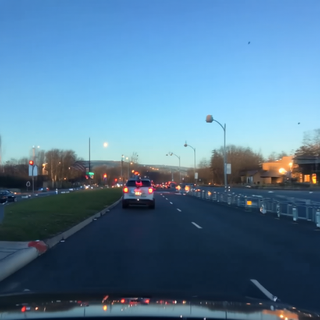} & \includegraphics[width=0.106\linewidth]{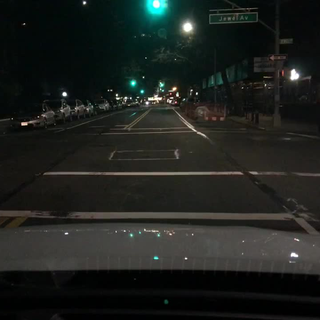} \\
\rotatebox{90}{\fontsize{5.5}{6.2}\selectfont BDD night} & \includegraphics[width=0.106\linewidth]{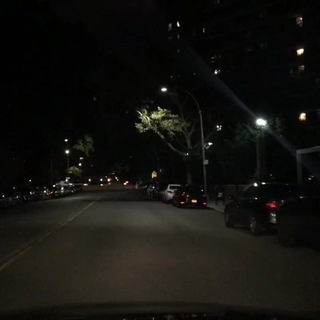} & \includegraphics[width=0.106\linewidth]{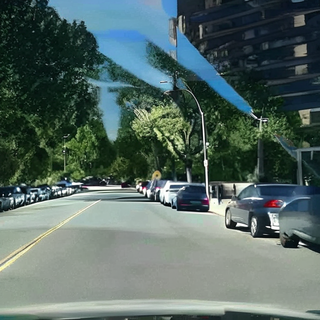} & \includegraphics[width=0.106\linewidth]{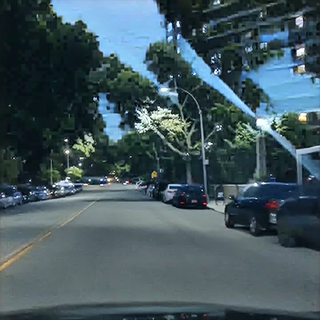} & \includegraphics[width=0.106\linewidth]{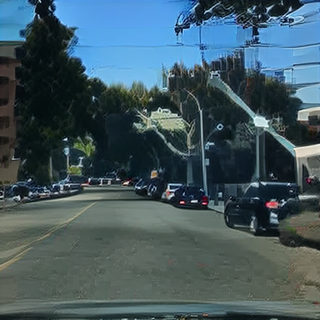} & \includegraphics[width=0.106\linewidth]{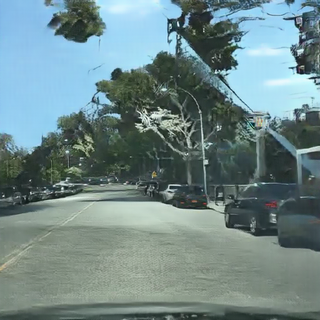} & \includegraphics[width=0.106\linewidth]{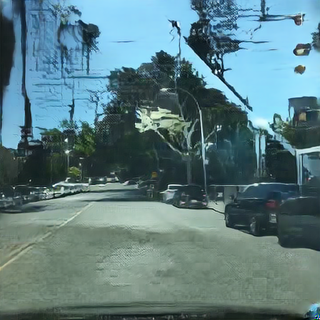} & \includegraphics[width=0.106\linewidth]{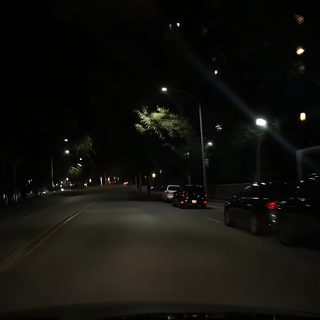} & \includegraphics[width=0.106\linewidth]{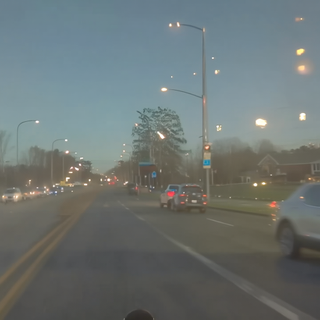} & \includegraphics[width=0.106\linewidth]{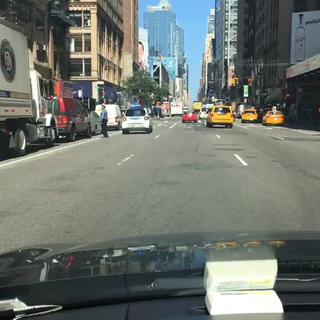} \\
 & \benchlab{input} & \benchlab{RFA (ours)} & \benchlab{CycleGAN-Turbo} & \benchlab{CycleGAN-Sprint} & \benchlab{CycleGAN} & \benchlab{CUT} & \benchlab{Cosmos, blur} & \benchlab{Cosmos, edge} & \benchlab{reference} \\
\end{tabular}
\caption{The weather benchmark on two further frames per condition, none of them the frames of Figure~\ref{fig:f1}; columns and references as there.}
\label{fig:f1ext}
\end{figure}

Figure~\ref{fig:f1ext} shows two further frames per condition with the columns of Figure~\ref{fig:f1}.

\subsection{Sim-to-real, more examples}
\label{app:s2rfig}

\begin{figure}[t]
\centering
\setlength{\tabcolsep}{0.8pt}\renewcommand{\arraystretch}{0.6}
\begin{tabular}{@{}ccccc@{}}
\includegraphics[width=0.196\linewidth]{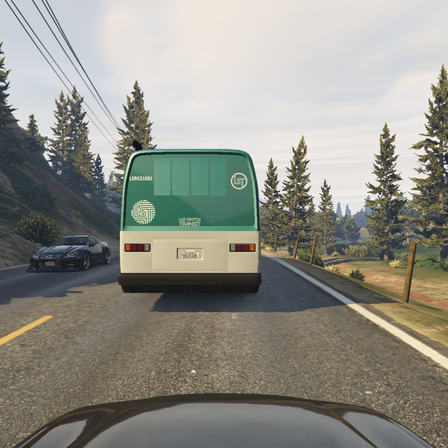} & \includegraphics[width=0.196\linewidth]{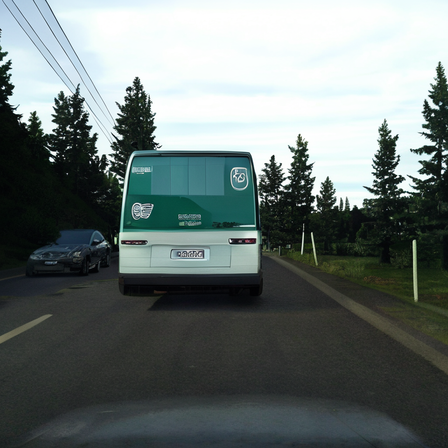} & \includegraphics[width=0.196\linewidth]{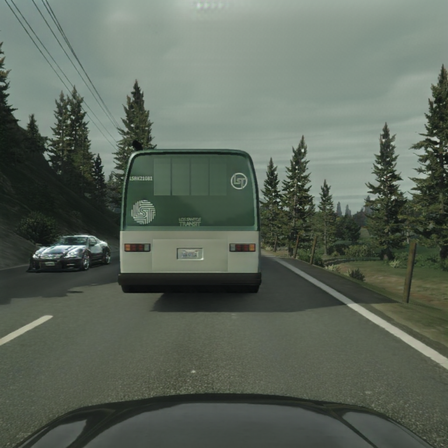} & \includegraphics[width=0.196\linewidth]{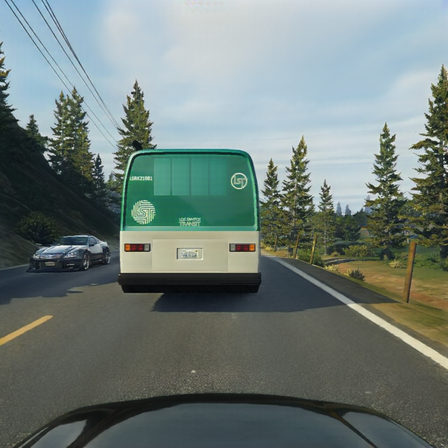} & \includegraphics[width=0.196\linewidth]{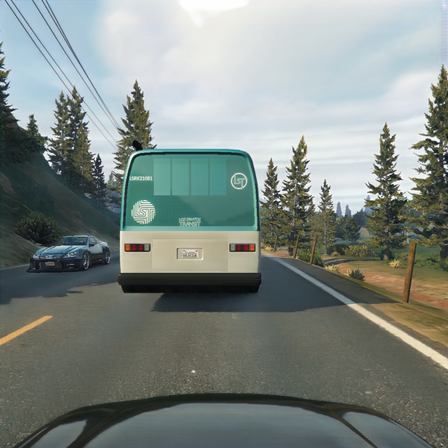} \\
\includegraphics[width=0.196\linewidth]{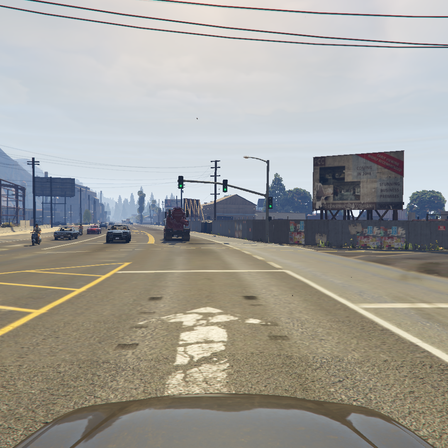} & \includegraphics[width=0.196\linewidth]{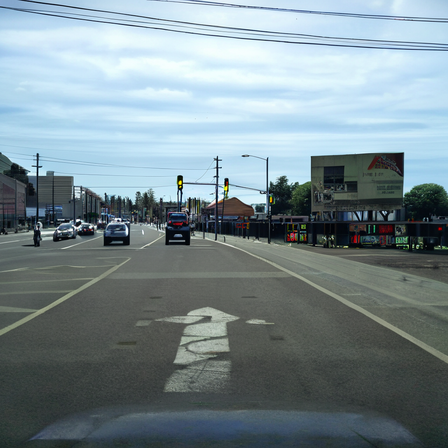} & \includegraphics[width=0.196\linewidth]{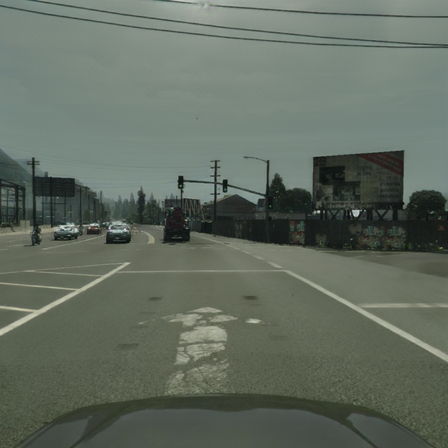} & \includegraphics[width=0.196\linewidth]{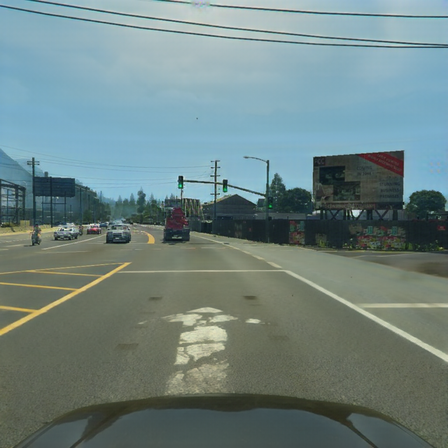} & \includegraphics[width=0.196\linewidth]{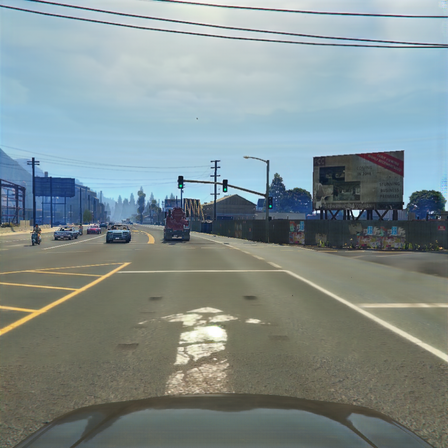} \\
\includegraphics[width=0.196\linewidth]{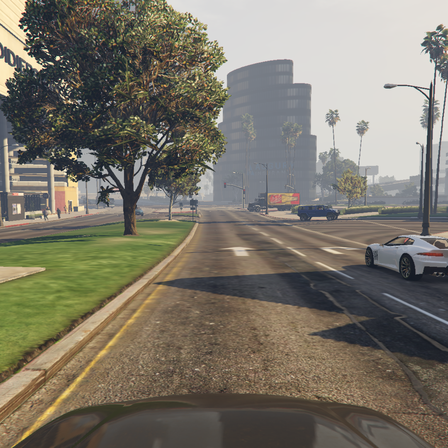} & \includegraphics[width=0.196\linewidth]{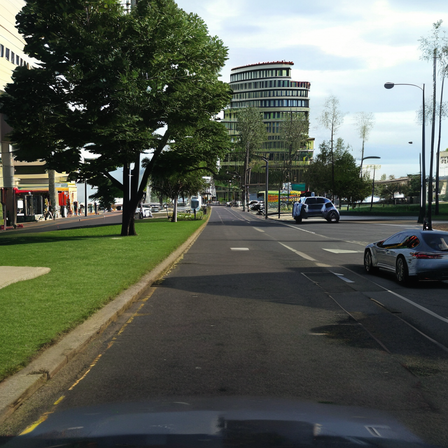} & \includegraphics[width=0.196\linewidth]{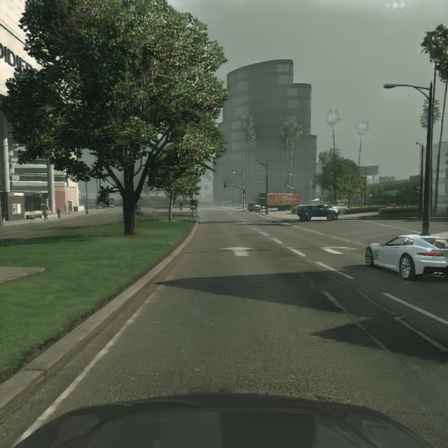} & \includegraphics[width=0.196\linewidth]{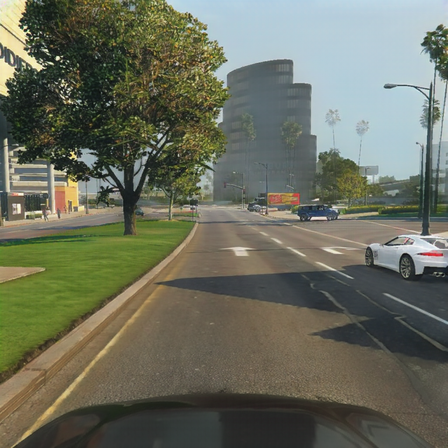} & \includegraphics[width=0.196\linewidth]{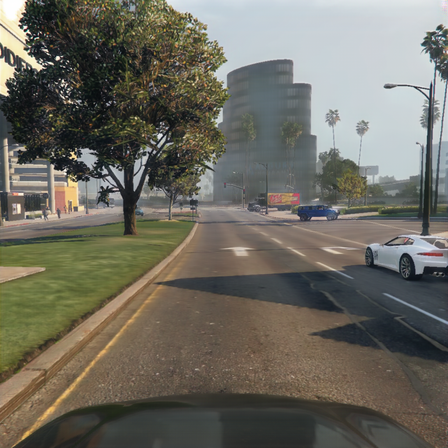} \\
\scriptsize PreSIL input & \scriptsize RFA (ours) & \scriptsize REGEN $\rightarrow$ CS & \scriptsize REGEN $\rightarrow$ Vistas & \scriptsize HyPER-GAN \\
\end{tabular}
\caption{Sim-to-real on three further matched crops. The pixel-fed GANs keep identity and change appearance; the RFA changes what the features leave free.}
\label{fig:f7ext}
\end{figure}

Figure~\ref{fig:f7ext} extends Figure~\ref{fig:f7} by three further matched crops.

\subsection{Portability, continued}
\label{app:portability}

\begin{figure}[t]
\centering
\setlength{\tabcolsep}{0.8pt}\renewcommand{\arraystretch}{0.6}
\begin{tabular}{@{}ccccc@{}}
\includegraphics[width=0.196\linewidth]{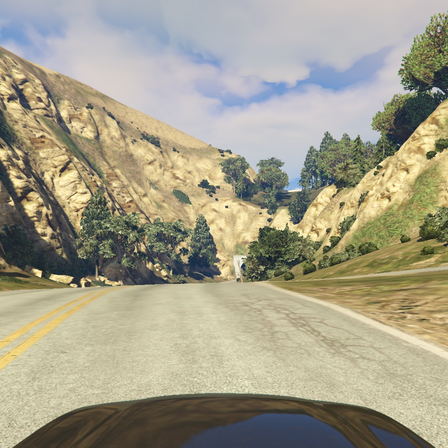} & \includegraphics[width=0.196\linewidth]{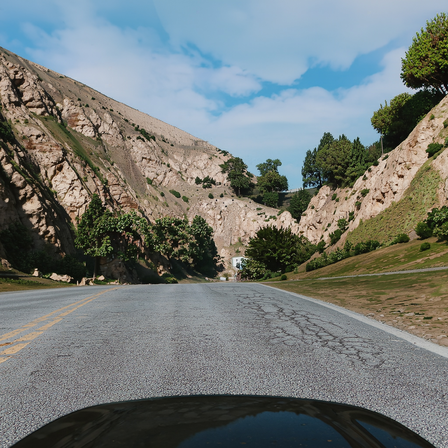} & \includegraphics[width=0.196\linewidth]{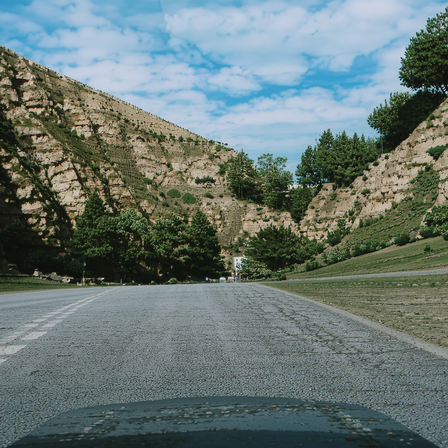} & \includegraphics[width=0.196\linewidth]{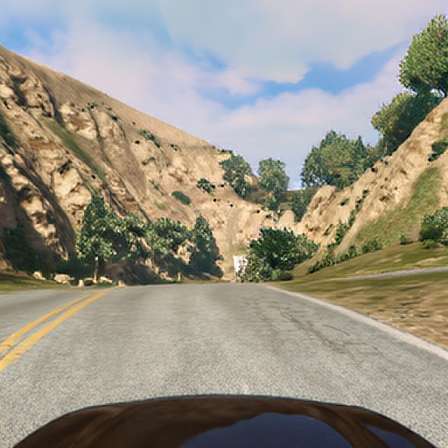} & \includegraphics[width=0.196\linewidth]{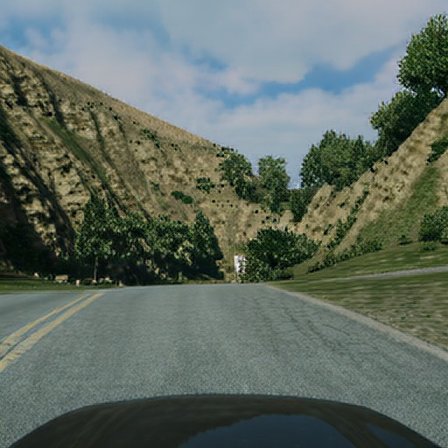} \\
\includegraphics[width=0.196\linewidth]{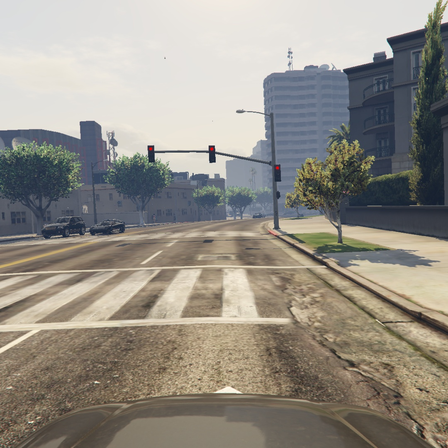} & \includegraphics[width=0.196\linewidth]{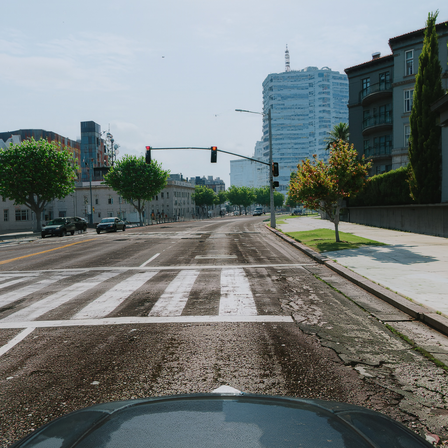} & \includegraphics[width=0.196\linewidth]{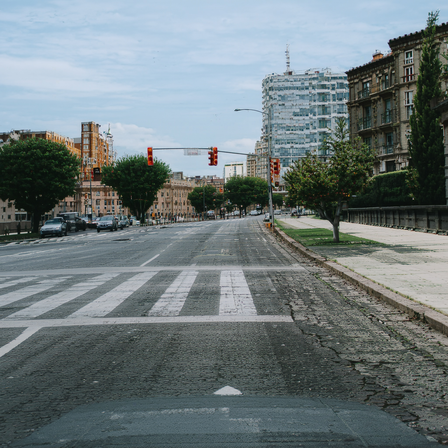} & \includegraphics[width=0.196\linewidth]{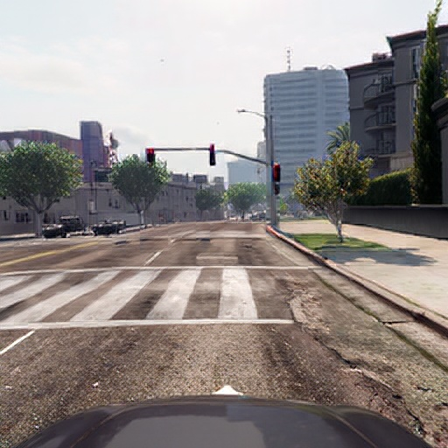} & \includegraphics[width=0.196\linewidth]{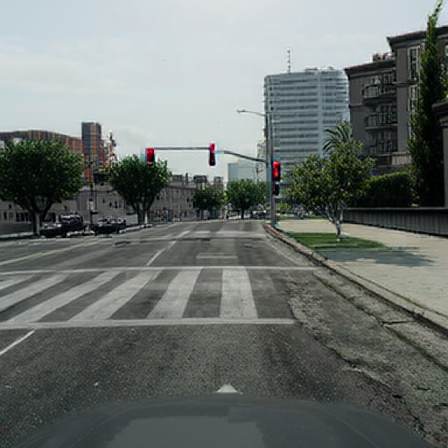} \\
\includegraphics[width=0.196\linewidth]{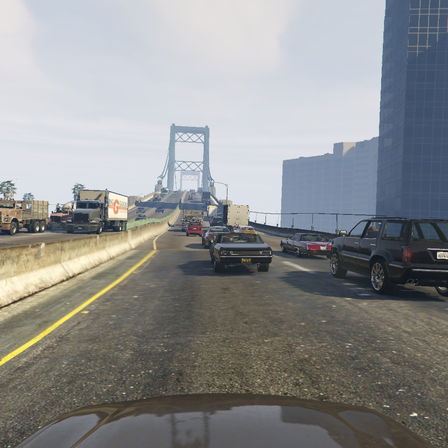} & \includegraphics[width=0.196\linewidth]{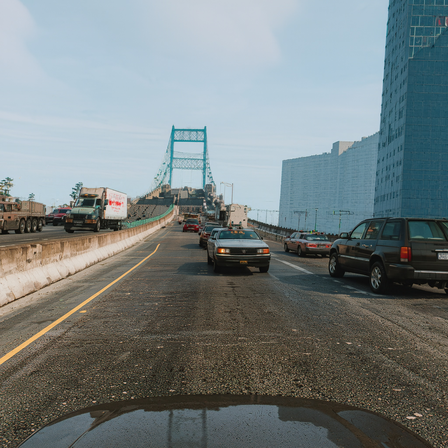} & \includegraphics[width=0.196\linewidth]{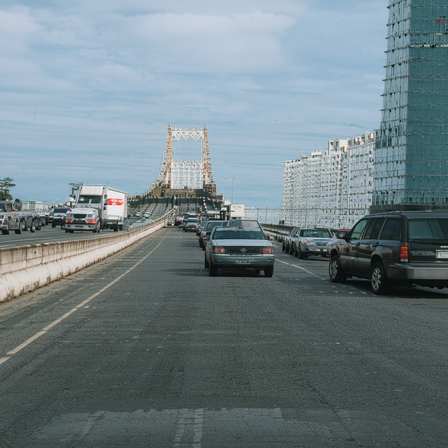} & \includegraphics[width=0.196\linewidth]{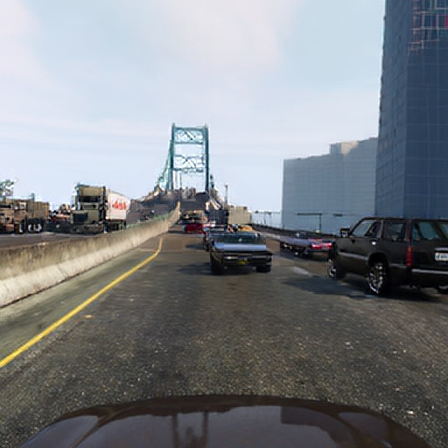} & \includegraphics[width=0.196\linewidth]{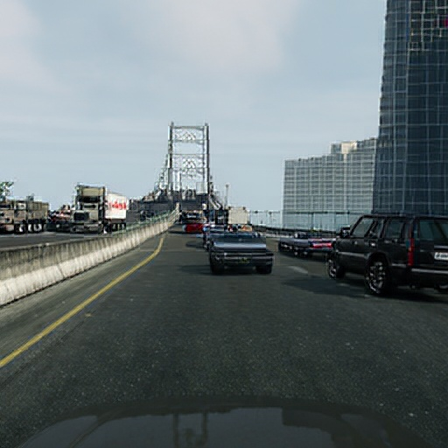} \\
\scriptsize PreSIL input & \scriptsize PiD (raw) & \scriptsize PiD (RFA) & \scriptsize RAE (raw) & \scriptsize RAE (RFA) \\
\end{tabular}
\caption{Portability on three further PreSIL frames: PiD and RAE decoding raw features and the same features after the sim-to-real RFA.}
\label{fig:f8ext}
\end{figure}

Figure~\ref{fig:f8} uses PiD's own encoding for every input, PiD at four sampling steps, and RAE in one pass. What differs between the two decoders is what they do where the features are silent: PiD samples what it cannot read and changes the car's colour and body, RAE does not sample and keeps the car yellow while shifting lighting and palette, so the colour change of Section~\ref{sec:sim2real} is the decoder's, not the adapter's. Neither decoder was trained on our domain as far as their papers state, so each carries a prior toward its own training distribution, and the raw column is the control: from the raw features both render the game frame back, so whatever separates the two columns of one decoder is the adapter's doing and not the decoder's prior.

The adapter was trained on the conditioning that reproduces PiD's encoding, so the pairing shown is the in-distribution one. Feeding either decoder through another encoding path is a mismatch on the encoder side and not a test of the adapter.

A second version of the test holds everything fixed but the decoder's training data: RAEv2 ships two decoders for its 23-layer DINOv3 encoder, one trained on ImageNet and one on a general image set, and our adapter for that encoding was trained through the general one. Rendering the same adapted features through the ImageNet decoder, which never saw the adapter, returns the same scene, buildings, car and its colour, road text and foliage, differing in fine texture and colour grading. Scored against the Mapillary crops at 256\,px, the decoder that never saw the adapter reaches KID 0.024 and FID 111.3 against 0.020 and 108.9 for the one it was trained through, and the two renders of a frame differ by structure distance 0.006, a third of the 0.019 that separates either from the game frame, so both decoders read the adapted features the same way.

\subsection{Beyond driving: horse to zebra}
\label{app:h2z}

\begin{figure}[t]
\centering
\setlength{\tabcolsep}{0.8pt}\renewcommand{\arraystretch}{0.6}
\begin{tabular}{@{}cccccc@{}}
\includegraphics[width=0.16\linewidth]{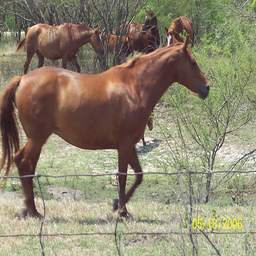} & \includegraphics[width=0.16\linewidth]{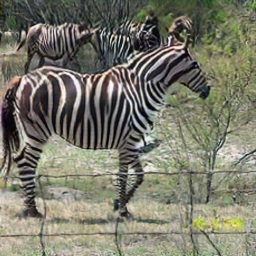} & \includegraphics[width=0.16\linewidth]{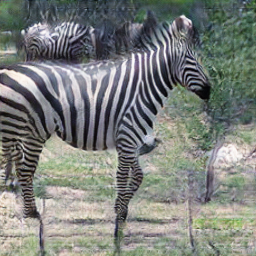} & \includegraphics[width=0.16\linewidth]{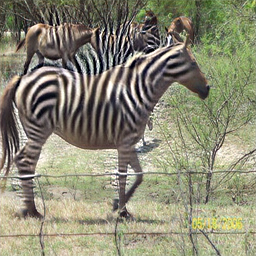} & \includegraphics[width=0.16\linewidth]{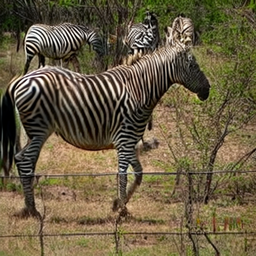} & \includegraphics[width=0.16\linewidth]{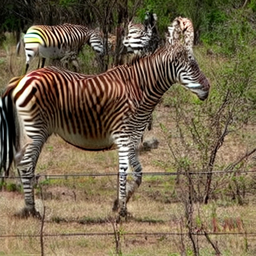} \\
\includegraphics[width=0.16\linewidth]{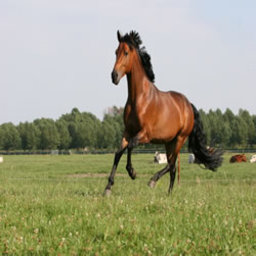} & \includegraphics[width=0.16\linewidth]{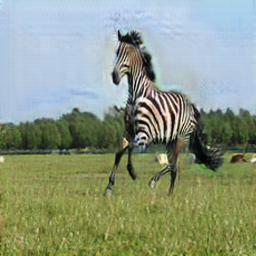} & \includegraphics[width=0.16\linewidth]{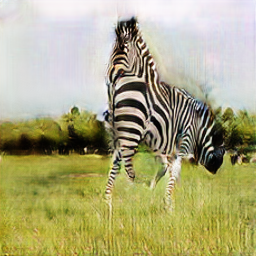} & \includegraphics[width=0.16\linewidth]{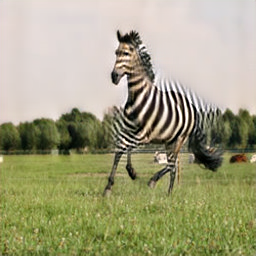} & \includegraphics[width=0.16\linewidth]{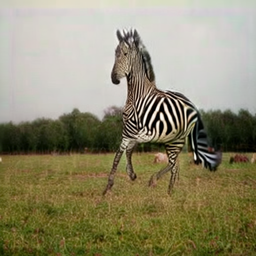} & \includegraphics[width=0.16\linewidth]{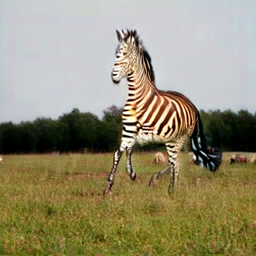} \\
\includegraphics[width=0.16\linewidth]{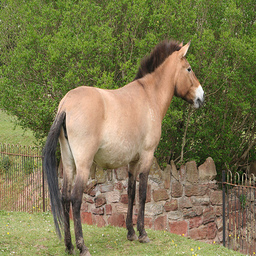} & \includegraphics[width=0.16\linewidth]{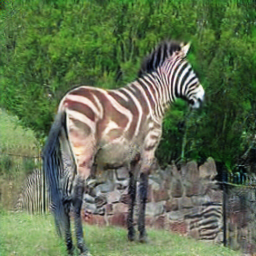} & \includegraphics[width=0.16\linewidth]{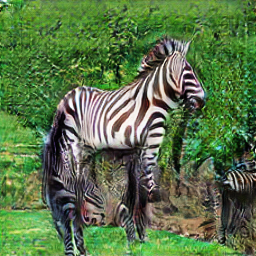} & \includegraphics[width=0.16\linewidth]{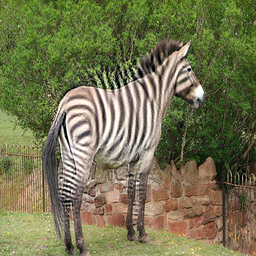} & \includegraphics[width=0.16\linewidth]{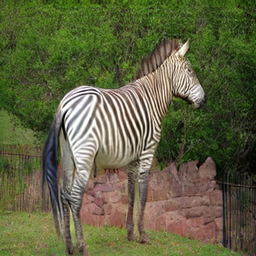} & \includegraphics[width=0.16\linewidth]{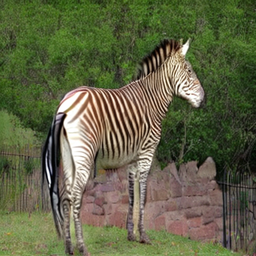} \\
\includegraphics[width=0.16\linewidth]{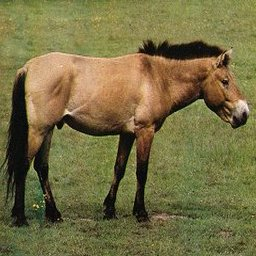} & \includegraphics[width=0.16\linewidth]{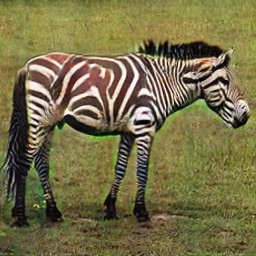} & \includegraphics[width=0.16\linewidth]{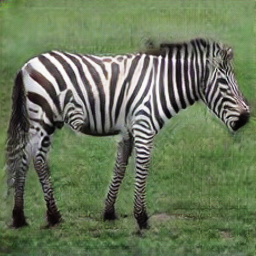} & \includegraphics[width=0.16\linewidth]{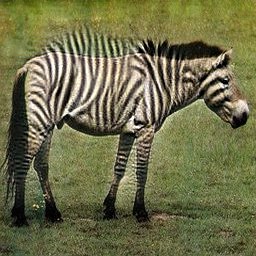} & \includegraphics[width=0.16\linewidth]{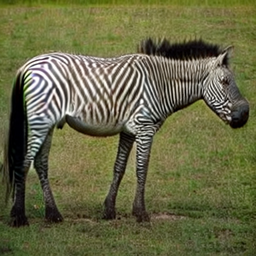} & \includegraphics[width=0.16\linewidth]{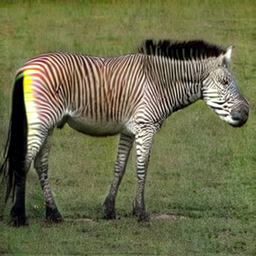} \\
\scriptsize test horse & \scriptsize CycleGAN, released & \scriptsize CUT, released & \parbox[t]{0.16\linewidth}{\centering\scriptsize CycleGAN-Turbo\\step 25000} & \scriptsize RFA (ours) & \parbox[t]{0.16\linewidth}{\centering\scriptsize RFA (ours)\\no patch discriminator} \\
\end{tabular}
\caption{Horse $\to$ zebra, four test horses: the released CycleGAN and CUT models, CycleGAN-Turbo trained with the authors' recipe. The two RFA runs compare the default recipe with training without the patch discriminator.}
\label{fig:h2z}
\end{figure}

The paper's benchmarks are driving footage, the domain the decoder was trained on. As a check that the recipe is not tied to that domain, the RFA was run on the CycleGAN horse-to-zebra set \citep{zhu2017cyclegan} and scored with CycleGAN-Turbo's protocol for it \citep{parmar2024cycleganturbo} (Table~\ref{tab:h2z}). This is a demonstration on a consumer GPU with the DINOv3-S configuration (0.6\,B ControlNet, batch 1 with four-step gradient accumulation), and it departs from the driving setting in three ways: the 256\,px images are upscaled to 1024\,px for the encoder and the decoder and the renders downscaled to 256\,px for scoring; the augmentation is CycleGAN's, horizontal flip and a mild random crop, instead of the weather recipe's vertical flip and rotation; and the decoder was fine-tuned on the horse-to-zebra training images. Everything else is the weather recipe, run for 9\,000 steps in the horse $\to$ zebra direction, once as is and once without the patch discriminator, which isolates that discriminator's contribution. CycleGAN-Turbo has no released horse-to-zebra model, so it was trained on the same consumer GPU with the authors' recipe (SD-Turbo, 25\,000 steps, batch 1). That run ends at 49.0 FID and does not reach the paper's 41.0. Both RFA runs beat CycleGAN on FID with CycleGAN-level structure and sit level with the CycleGAN-Turbo trained here. The two are within noise of each other on both scores, but the difference is visible in Figure~\ref{fig:h2z}.

\begin{table}[t]
\caption{Horse $\to$ zebra on CycleGAN-Turbo's protocol: FID of the 120 translated test horses against the 140 test zebras at 256\,px and DINO-Struct $\times$100 between input and output, lower is better. Bold best.}
\label{tab:h2z}
\begin{center}
\footnotesize
\begin{tabular}{lcc}
\toprule
\textbf{method} & \textbf{FID $\downarrow$} & \textbf{DINO-Struct $\downarrow$} \\
\midrule
RFA (ours) & 50.1 & \textbf{2.30} \\
RFA (ours), no patch discriminator & 50.7 & 2.34 \\
CycleGAN & 74.9 & 3.22 \\
CUT & \textbf{43.9} & 6.58 \\
CycleGAN-Turbo, retrained$^*$ & 49.0 & 2.65 \\
\bottomrule
\multicolumn{3}{l}{\scriptsize $^*$ paper result \citep{parmar2024cycleganturbo}: 41.0 / 2.1.} \\
\end{tabular}
\end{center}
\end{table}

\section{The one-step decoder}
\label{app:decoder}

\subsection{CycleGAN-Turbo's claim, and its own evidence}
\label{app:onestep}

CycleGAN-Turbo \citep{parmar2024cycleganturbo} reports that \emph{``directly applying standard diffusion adapters like ControlNet to the one-step setting proved less effective in our experiments''}. Two pieces of evidence carry the claim. Its ablation Config B, a ControlNet encoder on the one-step model, matches the FID of the released design but has \emph{``a significantly higher DINO-Structure distance''}, which the paper attributes to a conflict between the noise map and the conditioning inside a one-step network, and the adapter branch is dropped. And its Figure 8 places a one-step ControlNet next to the 100-step ControlNet: the 100-step one is competitive with their method, the one-step one is visibly degraded. Read together, the two say that a one-step ControlNet \emph{as they trained it} fails, and leave open whether a ControlNet can be trained for a distilled model at all. Their test judges generation, where an under-denoised sample passes as plausible, and cannot settle it. Reconstruction can: with the feature map of a held-out frame as the conditioning, the render sits next to its ground truth, and every failure is visible and measurable. That is how we built ours (Appendix~\ref{app:cn}), and it answers the question the other way. A ControlNet on the one-step Sana-Sprint reconstructs held-out frames better than the model's own 20-step teacher, once two things in the training recipe are right: it is trained against the teacher and it is run with the student's time convention. Its Config B, as far as the paper describes it, had neither. On SD-Turbo, whose teacher is not published, our own ControlNet recipes improved on their reported failure but did not remove it, which is why the port of Section~\ref{sec:backbone} runs in the other direction.

\subsection{Our one-step ControlNet, built as a reconstruction task}
\label{app:cn}

The decoder is a ControlNet \citep{zhang2023controlnet} on Sana-Sprint, written for Sana's transformer. \textbf{Conditioning adapter.} The frozen encoder's feature map ($384 \times 64 \times 64$ for DINOv3-S at a 1024-px input, $768 \times 32 \times 32$ for DINOv2-B/reg at 448 px in PiD's normalisation) is brought to the $32 \times 32$ control map by a parameter-free $2 \times 2$ space-to-depth merge where the map is larger, followed by a small convolutional refinement whose last layer is zero-initialised, and projected to the 32 channels of the DC-AE latent, so that the control enters the ControlNet exactly as a latent would. \textbf{ControlNet.} Seven transformer blocks of the backbone's width read the patchified control together with the noisy latent, the text tokens and the timestep. A zero-initialised linear layer injects the control into the first block and one per block adds that block's output to the corresponding frozen block, so at initialisation the ControlNet leaves the backbone untouched. \textbf{Training.} The ControlNet is trained against the \emph{teacher}, the published multi-step model Sana-Sprint was distilled from, with the teacher's flow-matching objective (target noise $-$ x, time $t \in [0, 1]$ drawn logit-normal), on Mapillary at 1024 px with a fixed caption and no conditioning dropout; batch 16, learning rate $10^{-4}$ constant after 500 warmup steps, 8-bit Adam, bf16, gradient clipping at 0.3, 15\,000 steps. Nothing in the backbone is trained. \textbf{Inference.} The trained ControlNet is plugged unchanged into the distilled student and run with the student's sampler at its shipped two steps ($t = \pi/2$, then 1.3, guidance 4.5, the student's own guidance embedding). The teacher, when rendered for comparison, runs its default sampler unguided, since text guidance costs a ControlNet trained without conditioning dropout 5 to 6.5\,dB on every teacher we tried (18.41\,dB unguided against 13.64 at guidance 4.5). The one detail that matters: the student's sampler works in TrigFlow time $t \in [0, \pi/2]$ and the ControlNet was trained in flow-matching time, so the pipeline converts the sampler's time as $\sin t / (\cos t + \sin t)$, rescales the latents accordingly and converts the prediction back at every evaluation.

\begin{figure}[H]
\centering
\setlength{\tabcolsep}{0.8pt}\renewcommand{\arraystretch}{0.6}
\begin{tabular}{@{}ccc@{}}
\includegraphics[width=0.19\linewidth]{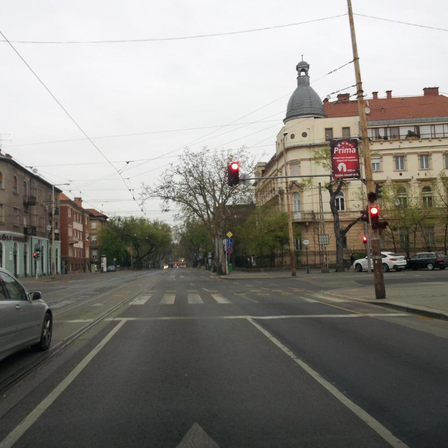} & \includegraphics[width=0.19\linewidth]{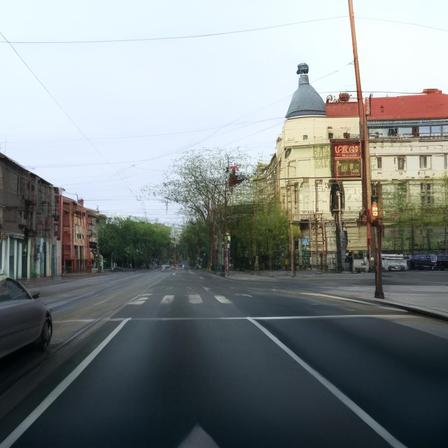} & \includegraphics[width=0.19\linewidth]{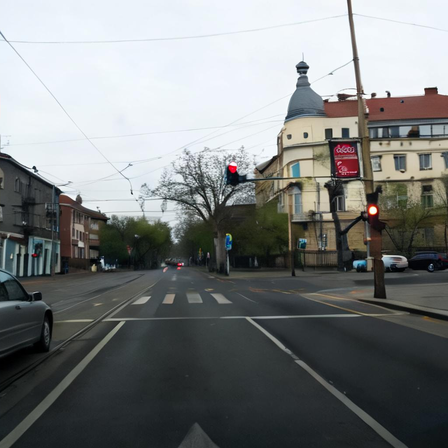} \\
\scriptsize ground truth & \scriptsize teacher, 20 steps & \scriptsize student, 2 steps \\
\end{tabular}
\caption{The 0.6\,B DINOv3-S ControlNet on one held-out frame: the 20-step teacher it was trained against and the 2-step student it runs on. The student reconstructs better, 19.5 against 18.0\,dB on the 50 frames of Table~\ref{tab:s5} scored at the 1024-px output, as does the 1.6\,B pair (17.8 against 17.0); Table~\ref{tab:s5} scores the same renders at 512\,px.}
\label{fig:ts}
\end{figure}
\paragraph{Training a ControlNet for a distilled model is not the same as for its teacher} With the sampler's raw TrigFlow time fed to the ControlNet, every render came out under-denoised and drifting to saturated colour, because the ControlNet was evaluated at times up to 1.57 on unscaled latents, outside anything it had seen. On the paper's decoder the raw-time render at the shipped two steps scores 9.0\,dB and gradNCC 0.18 against 20.5\,dB and 0.51 with the conversion (four frames), and its chroma error grows with the number of out-of-range evaluations while the converted path holds. Two things follow. First, the failure is invisible in the setting CycleGAN-Turbo reports it in, \emph{generation}, where an under-denoised, oversaturated image is a plausible sample, and obvious in \emph{reconstruction}, where the same image sits next to its ground truth. Their one-step ControlNet losing to the 100-step one in their Figure 8 is the symptom we saw, and we read it as the same mismatch between a distilled sampler and a ControlNet trained under another convention, which we cannot check because the ablation configuration is not released. Second, once the convention is right the student is not the weaker decoder: trained against the teacher and run on the student, the ControlNet reconstructs held-out frames better with two student steps than the teacher does with twenty (on the 50 frames of Table~\ref{tab:s5}, scored at the 1024-px output: 17.8 against 17.0\,dB for the 1.6\,B pair and 19.5 against 18.0\,dB for the 0.6\,B pair, the same ordering for every conditioning we trained), so the distilled model is the right place to run it (Figure~\ref{fig:ts}). The two inference steps are the shipped default, not a tuned choice: rendering the same weather adapters at one step instead of two moves KID within re-scoring noise (fog 0.028 against 0.026, snow 0.019 against 0.021, rain 0.016 against 0.016, haze 0.040 against 0.038), FID by one to three points in the one-step direction, and structure distance by 0.001 to 0.006 in the two-step direction.

\paragraph{Train against the teacher} Training the ControlNet on the distilled model directly was difficult and did not give us a usable decoder, since its renders were under-denoised, as above. A ControlNet trained against the multi-step teacher and run unchanged on the student works, and works better on the student than on the teacher (Figure~\ref{fig:ts}). This order cannot be tested on the SD-Turbo backbone, which publishes no teacher, and it is in accord with what CycleGAN-Turbo's own one-step ControlNet shows (Appendix~\ref{app:onestep}).

\subsection{The same-backbone comparison, and what the port required}
\label{app:backbone}

Our decoder recipe needs the backbone's multi-step teacher (Appendix~\ref{app:cn}), which Sana-Sprint publishes and SD-Turbo does not (SD-2 base is a surrogate), so the comparison that holds the backbone fixed ports their method to ours (Section~\ref{sec:backbone}) as a best-effort rebuild that keeps every choice their paper or code fixes and states every choice the backbone forces. What the port does not control is the difference in what is trainable, a generator per domain pair against one shared decoder and an adapter per pair, so Table~\ref{tab:t2} stays a task-level comparison with the backbone held fixed for the CycleGAN-Sprint rows.

\textbf{What the port required.} Their recipe rests on a premise the paper does not state: fed the encoded input in place of the noise map at the last timestep, the frozen one-step model already returns approximately that image, so the LoRA only has to learn the translation. On SD-Turbo this holds. On Sana-Sprint it does not: the DC-AE round trip is 0.028 L1 from the input, but the frozen model's one-step prediction from the encoded latent is 0.472 L1 away. The verbatim rebuild spent its LoRA on reconstruction and then collapsed to the identity, with the cycle and identity terms holding it there. Making the LoRA a residual on the frozen prediction, $x_0 = z - (v_{\text{LoRA}} - v_{\text{frozen}})$, restores the premise: the generator is exactly the identity at initialisation, as their recipe presumes. With the premise restored, their adversarial weight of 0.5 still held the generator at the input (fog KID 0.071 against 0.105 for doing nothing, structure distance 0.007): a per-term gradient probe on the LoRA gave the adversarial term one fiftieth of the cycle term's gradient norm, so it is swamped on this backbone. Raising it to 2.0 or 4.0 produces a translator; the two are within noise of each other on KID (0.037 and 0.034) and 2.0 keeps measurably more structure (structure distance 0.016 against 0.019); the paper reports 4.0, the lower KID. The residual form and the adversarial weight are the whole of the departure from their recipe, and each is what made the method train on this backbone at all.

\paragraph{Rebuild details} Backbone: Sana-Sprint 1.6\,B frozen, LoRA of rank 128 on the transformer and a LoRA of rank 4 plus skip connections on the DC-AE autoencoder, the ranks and skips of their training code. Generator: the encoded input replaces the noise map at Sana-Sprint's last timestep (\emph{t} = 1, the analogue of their \emph{t} = 999), no added noise, one transformer evaluation, guidance 4.5, LoRA as a residual on the frozen prediction. Losses, as in their code: two vision-aided CLIP discriminators \citep{kumari2022visionaided} with multi-level convolutional heads, one per domain, adversarial weight 4.0 (their 0.5); cycle L1 weight 1.0 with LPIPS weight 2.0 (their 10.0); identity L1 1.0 with LPIPS 1.0. Data and schedule: 512\,px with their preprocessing (resize, then random crop), batch 1, learning rate $10^{-5}$ for generator and discriminators with 500 warmup steps, gradient clipping at 10, 25\,000 steps, reported at the end of the recipe like the released model, and one-sentence prompts per domain (``a photo of a driving scene in clear weather'', ``\ldots in dense fog''). Trainable weights and cost are in Table~\ref{tab:t4}.

\textbf{A convergence.} Their Config B matches FID while losing DINO-Structure, the pattern of Table~\ref{tab:t2}, where the RFA matches or wins KID and loses structure distance. Two measurements on different backbones with different objectives show the same trade, which points at the conditioning path and not at any one implementation.

\subsection{Is the diffusion decoder needed?}
\label{app:s6}
\label{app:decoders}
\label{app:s5}

No. The pipeline needs one differentiable pass from a DINO feature map to pixels (Section~\ref{sec:pipeline}). The one-step diffusion ControlNet is the pass this work inherits from the Control-DINO lineage, kept because it is fast, cheap to train and renders at 1024\,px. It is not the best decoder available, and it is interchangeable: RAE decodes the same adapted map without retraining (Section~\ref{sec:portability}), the reconstruction test below locates the detail our ControlNet loses, and the same adapter recipe trains through three other frozen decoders (Table~\ref{tab:s6}).

\textbf{Reconstruction.} On 50 held-out Mapillary frames, RAE \citep{zheng2026rae} reconstructs the same DINOv2-B/reg map our DINOv2 ControlNet consumes about 2\,dB better (20.14 against 18.21) and keeps sign text and window detail ours loses. The DINOv3-S configuration's ControlNet at 1024\,px matches RAE on PSNR (20.10) with better edges (gradNCC 0.650 against 0.557; Table~\ref{tab:s5}, Figure~\ref{fig:s5b}). RAEv2's released decoders \citep{singh2026raev2}, scored at their native 256\,px on the same frames, add one point: a single-layer decoder on the last DINOv3-S layer reaches 19.5\,dB, and the same ViT-XL decoder fed the sum of all 23 DINOv3-L layers reaches 26.8\,dB trained on ImageNet and 29.1\,dB trained on general data, close to the autoencoder ceiling, a representation the RFA does not use.

\textbf{Translation.} The adapter recipe also trains \emph{through} RAE's decoder and through RAEv2's single-layer DINOv3-S and 23-layer DINOv3-L decoders in place of our ControlNet, with encoder and decoder frozen exactly as the ControlNet is and every loss weight, the batch and the step count of Table~\ref{tab:config} unchanged (Table~\ref{tab:s6}). Two settings follow the decoder: the resolution, 512\,px for RAE and 256\,px for RAEv2, to which the real and sim crops are downsampled for the discriminators while the features are computed as before, and an added RGB identity term, because a deterministic decoder gives the adapter more leverage over global tone than the diffusion prior does and the discriminators otherwise chase a tone shift as the cheapest direction. At a common 512\,px our ControlNet has the best distribution scores (KID 0.012 against 0.020 through RAE and 0.028 through the 23-layer RAEv2 decoder). The 23-layer decoder, which reconstructs almost losslessly (Table~\ref{tab:s5}), keeps far more structure (structure distance 0.019 against our 0.032) and reaches KID 0.020 at its own 256\,px, against our 0.012. RAE sits between on both, and the single-layer RAEv2 decoder is last on KID and FID and level with ours on structure (0.031 against 0.032). The diffusion decoder is not needed, and the trade of Section~\ref{sec:cost-of-removal} follows the decoder: the more it can reconstruct, the more of the scene the adapted frame keeps.

\begin{table}[t]
\caption{Reconstruction from the conditioning alone: 50 held-out Mapillary frames, the same DINO feature map decoded by each decoder, scored against the input at 512\,px. RAE was trained on ImageNet, never on driving frames; our ControlNets on Mapillary and run at their shipped two steps. The DC-AE row is the autoencoder round trip of the ground truth, the ceiling for any render through it. Bold best of the three decoders.}
\label{tab:s5}
\label{tab:s5b}
\begin{center}
\footnotesize\setlength{\tabcolsep}{3pt}
\begin{tabular}{llcc}
\toprule
\textbf{decoder} & \textbf{features} & \textbf{PSNR $\uparrow$} & \textbf{gradNCC $\uparrow$} \\
\midrule
RAE, one pass (ViT-XL) & DINOv2-B/reg, 32\textsuperscript{2} & \textbf{20.14} & 0.557 \\
Sana ControlNet, two steps (RFA decoder) & DINOv2-B/reg, 32\textsuperscript{2} & 18.21 & 0.519 \\
Sana ControlNet, two steps (DINOv3-S configuration) & DINOv3-S at 1024\,px, 32\textsuperscript{2} & 20.10 & \textbf{0.650} \\
\midrule
DC-AE round trip (ceiling, no features) & pixels & 28.83 & 0.923 \\
\bottomrule
\end{tabular}
\end{center}
\end{table}

\begin{table}[t]
\caption{Sim-to-real adapters trained through four frozen decoders on the 100 PreSIL crops of Table~\ref{tab:t3}: the paper recipe of Table~\ref{tab:config} in every row, only the decoder changes. KID and FID at a common 512\,px against the 2\,000 Mapillary crops at that size, then at each decoder's native output size against references at that size (comparable only within a row's size), and structure distance to the source crop.}
\label{tab:s6}
\begin{center}
\footnotesize\setlength{\tabcolsep}{3pt}
\begin{tabular}{llccc}
\toprule
\textbf{adapter trained through} & \textbf{native} & \textbf{KID / FID, 512\,px} & \textbf{KID / FID, native} & \textbf{DINO $\downarrow$} \\
\midrule
our Sana ControlNet (DINOv2-B/reg) & 1024 & \textbf{0.0117} / \textbf{107.6} & 0.0123 / 107.6 & 0.0318 \\
RAE (DINOv2-B/reg) & 512 & 0.0198 / 114.7 & 0.0198 / 114.7 & 0.0265 \\
RAEv2, DINOv3-S, last layer & 256 & 0.0472 / 130.4 & 0.0336 / 118.2 & 0.0313 \\
RAEv2, DINOv3-L, 23-layer sum & 256 & 0.0277 / 116.8 & 0.0201 / 108.9 & \textbf{0.0192} \\
\bottomrule
\end{tabular}
\end{center}
\end{table}

\begin{figure}[t]
\centering
\setlength{\tabcolsep}{0.8pt}\renewcommand{\arraystretch}{0.6}
\begin{tabular}{@{}ccccc@{}}
\includegraphics[width=0.196\linewidth]{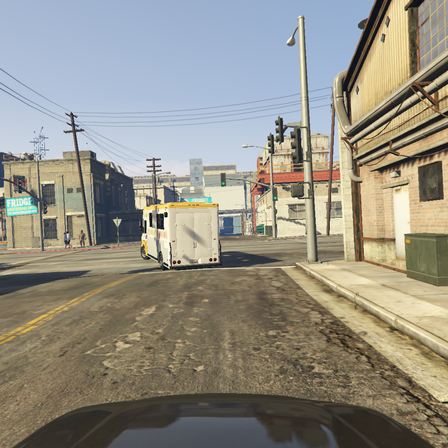} & \includegraphics[width=0.196\linewidth]{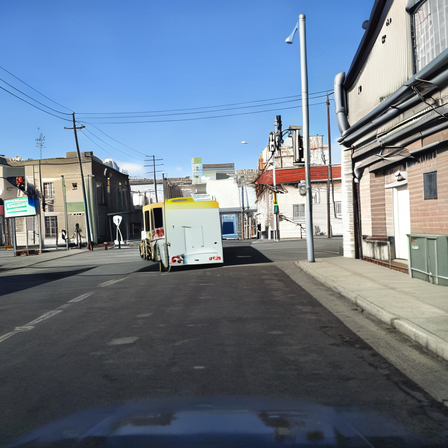} & \includegraphics[width=0.196\linewidth]{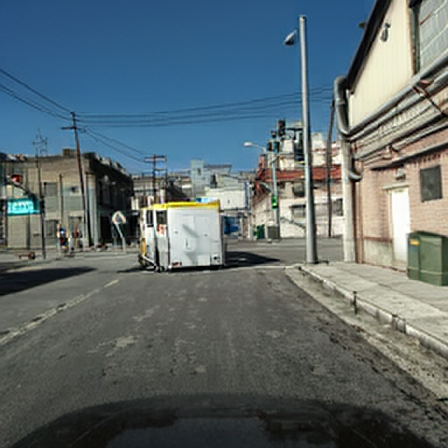} & \includegraphics[width=0.196\linewidth]{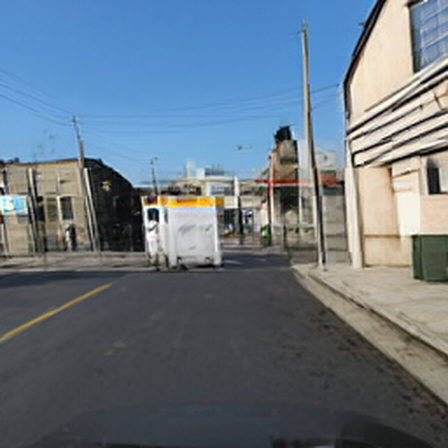} & \includegraphics[width=0.196\linewidth]{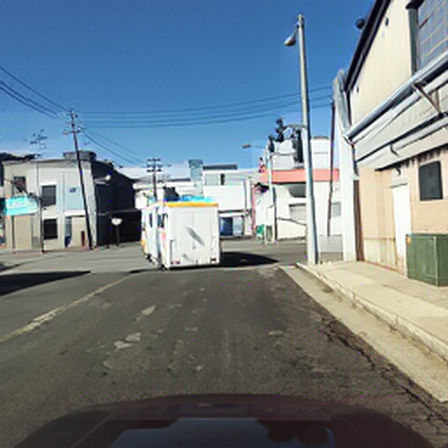} \\
\includegraphics[width=0.196\linewidth]{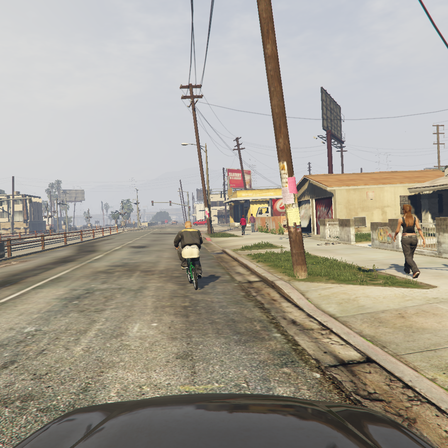} & \includegraphics[width=0.196\linewidth]{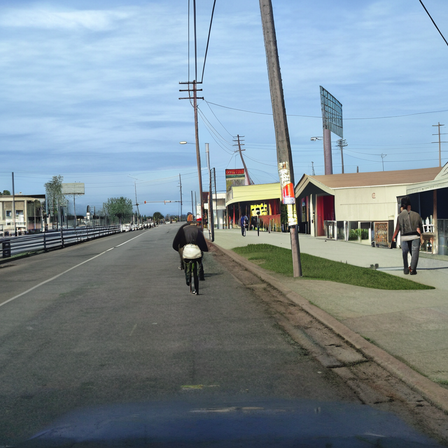} & \includegraphics[width=0.196\linewidth]{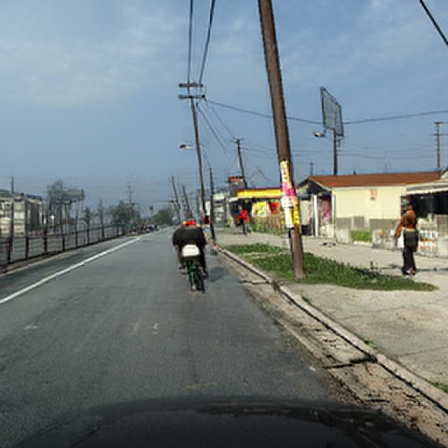} & \includegraphics[width=0.196\linewidth]{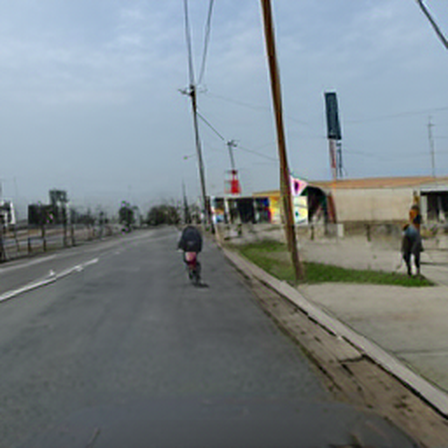} & \includegraphics[width=0.196\linewidth]{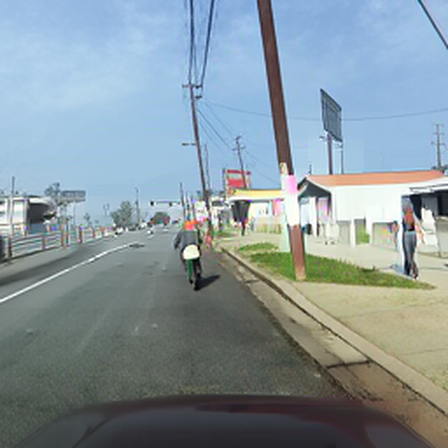} \\
\includegraphics[width=0.196\linewidth]{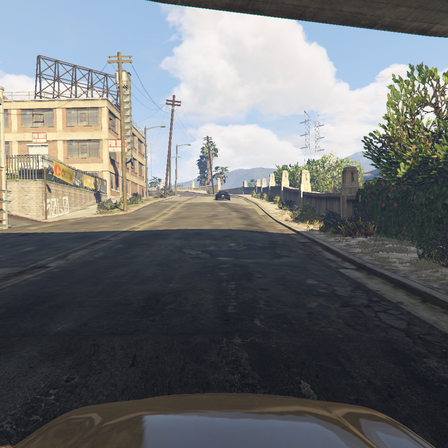} & \includegraphics[width=0.196\linewidth]{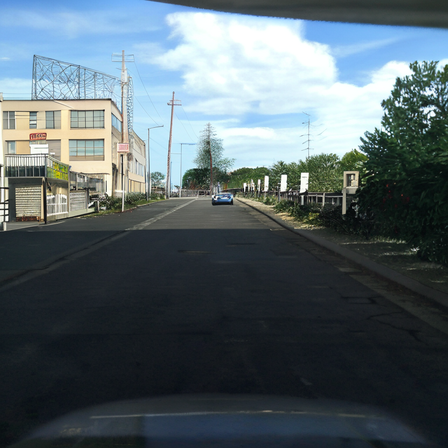} & \includegraphics[width=0.196\linewidth]{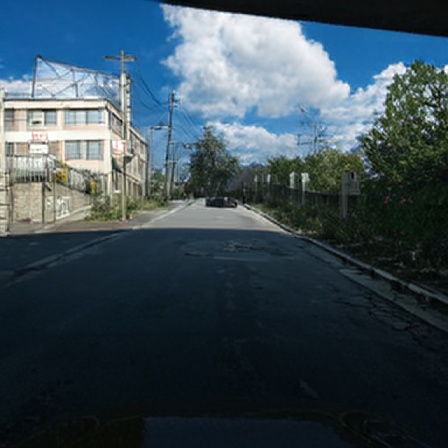} & \includegraphics[width=0.196\linewidth]{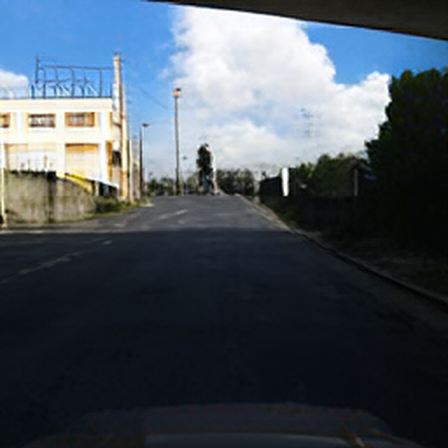} & \includegraphics[width=0.196\linewidth]{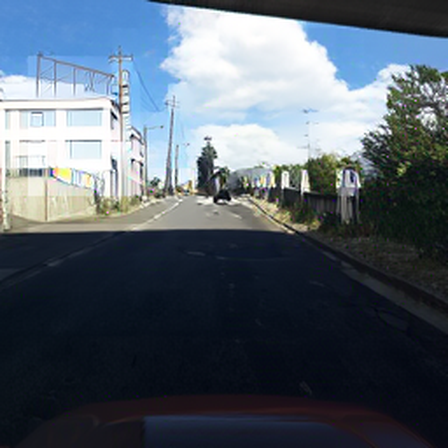} \\
\scriptsize PreSIL input & \parbox[t]{0.16\linewidth}{\centering\scriptsize RFA (ours), \\ Sana ControlNet} & \scriptsize RFA (RAE) & \parbox[t]{0.16\linewidth}{\centering\scriptsize RFA (RAEv2), \\ DINOv3-S}  & \parbox[t]{0.16\linewidth}{\centering\scriptsize RFA (RAEv2), \\ DINOv3-L, 23 layers}  \\
\end{tabular}
\caption{One adapter recipe, four frozen decoders (Table~\ref{tab:s6}). The scene survives in all four; what differs is fine content and how much of the layout moves. Decoders render at 1024, 512 and 256\,px, printed at one size.}
\label{fig:s6}
\end{figure}

\begin{figure}[t]
\centering
\setlength{\tabcolsep}{0.8pt}\renewcommand{\arraystretch}{0.6}
\begin{tabular}{@{}cccc@{}}
\includegraphics[width=0.245\linewidth]{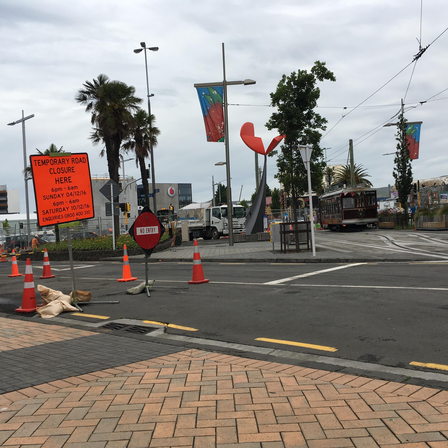} & \includegraphics[width=0.245\linewidth]{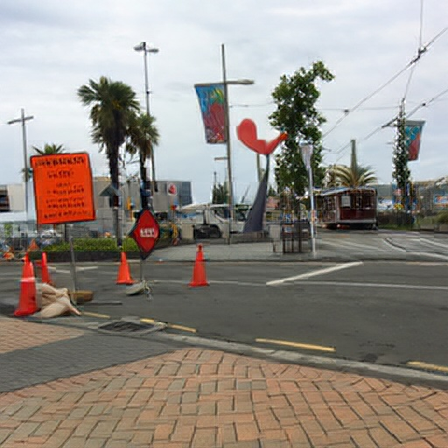} & \includegraphics[width=0.245\linewidth]{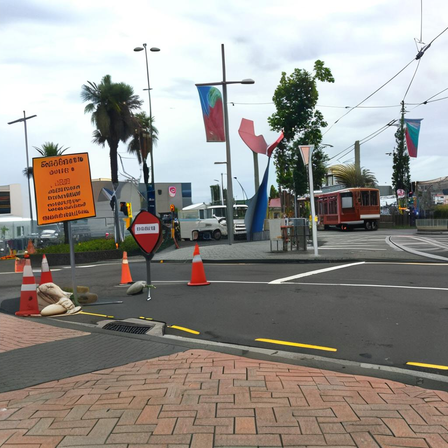} & \includegraphics[width=0.245\linewidth]{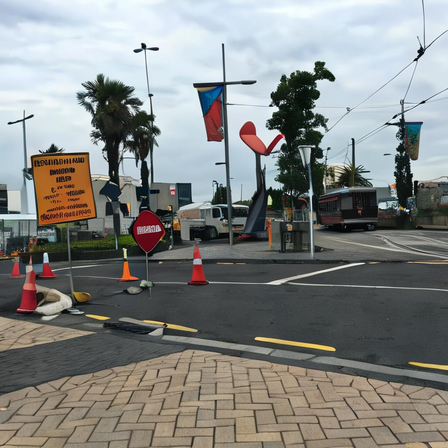} \\
\includegraphics[width=0.245\linewidth]{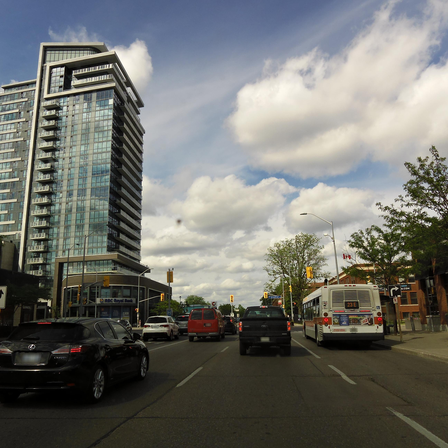} & \includegraphics[width=0.245\linewidth]{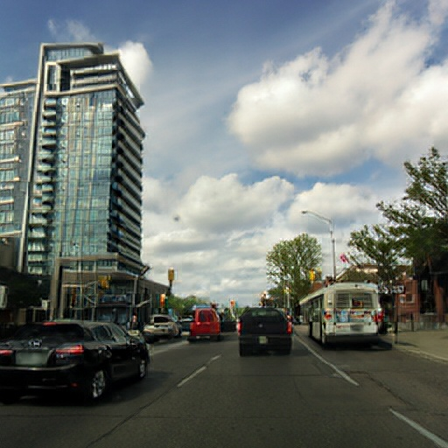} & \includegraphics[width=0.245\linewidth]{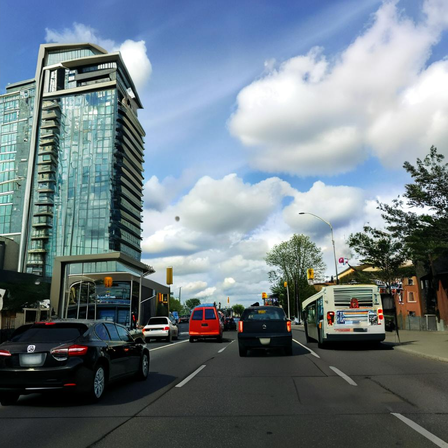} & \includegraphics[width=0.245\linewidth]{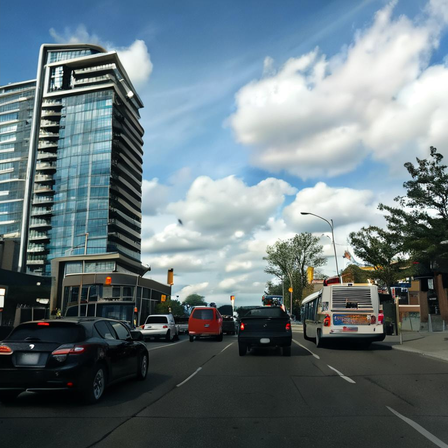} \\
\includegraphics[width=0.245\linewidth]{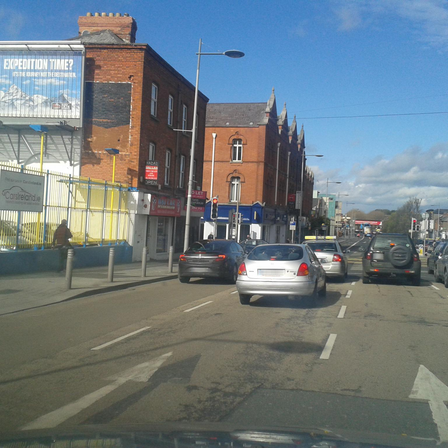} & \includegraphics[width=0.245\linewidth]{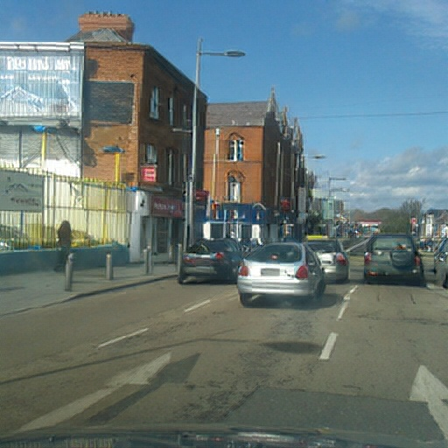} & \includegraphics[width=0.245\linewidth]{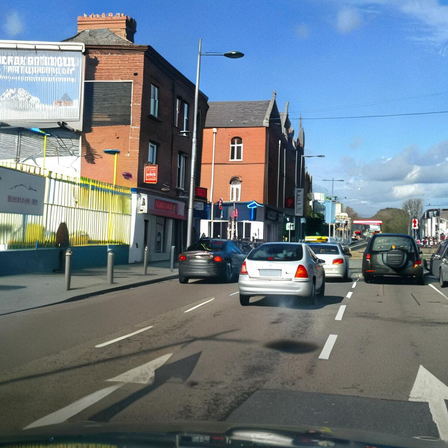} & \includegraphics[width=0.245\linewidth]{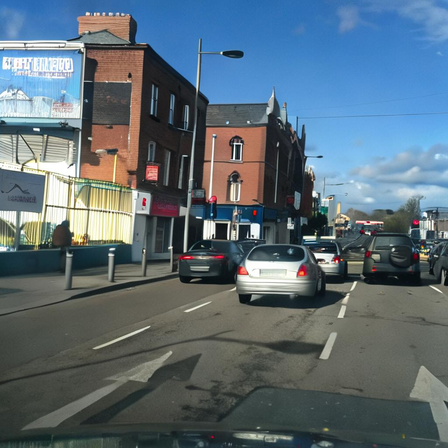} \\
\scriptsize ground truth & \scriptsize RAE & \scriptsize ours, DINOv2-B/reg & \scriptsize ours, DINOv3-S at 1024\,px \\
\end{tabular}
\caption{The frames behind Table~\ref{tab:s5}.}
\label{fig:s5b}
\end{figure}

\section{Decoder bias: the frozen decoder on unadapted features}
\label{app:bias}

\begin{figure}[tp]
\centering
\setlength{\tabcolsep}{0.8pt}\renewcommand{\arraystretch}{0.6}
\begin{tabular}{@{}c@{\hskip 2pt}ccccc@{}}
\rotatebox{90}{\scriptsize input} & \includegraphics[width=0.185\linewidth]{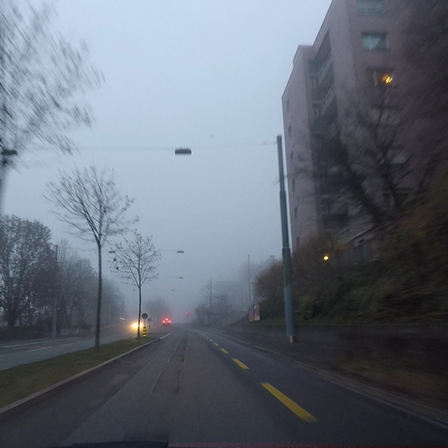} & \includegraphics[width=0.185\linewidth]{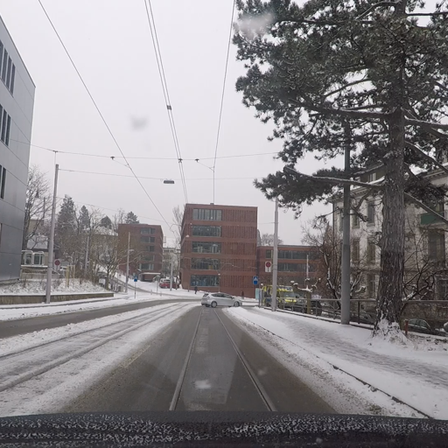} & \includegraphics[width=0.185\linewidth]{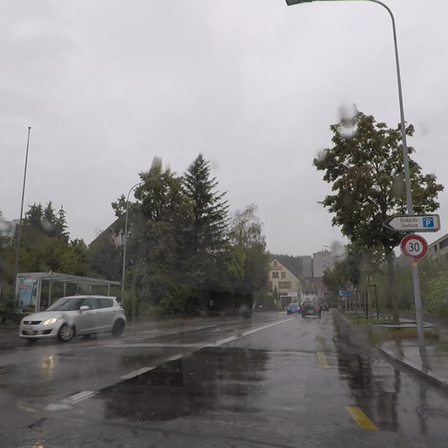} & \includegraphics[width=0.185\linewidth]{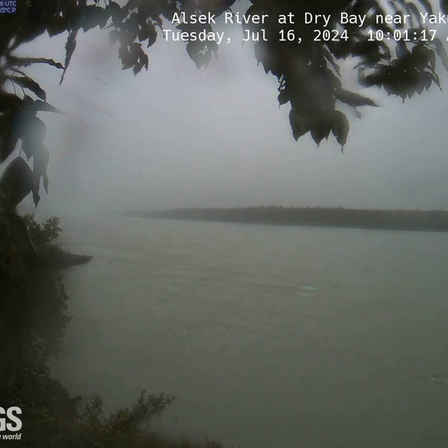} & \includegraphics[width=0.185\linewidth]{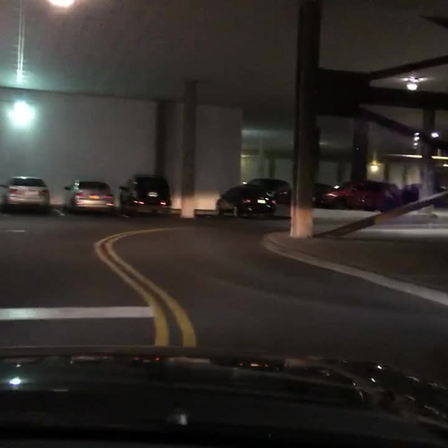} \\
\rotatebox{90}{\scriptsize decoder only} & \includegraphics[width=0.185\linewidth]{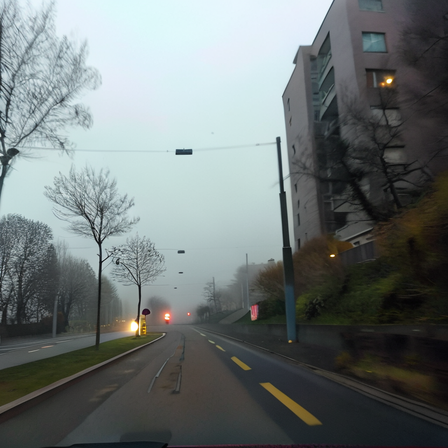} & \includegraphics[width=0.185\linewidth]{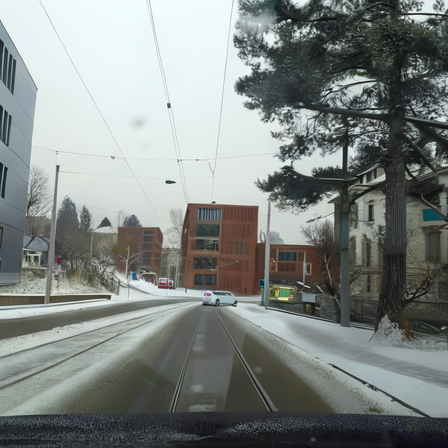} & \includegraphics[width=0.185\linewidth]{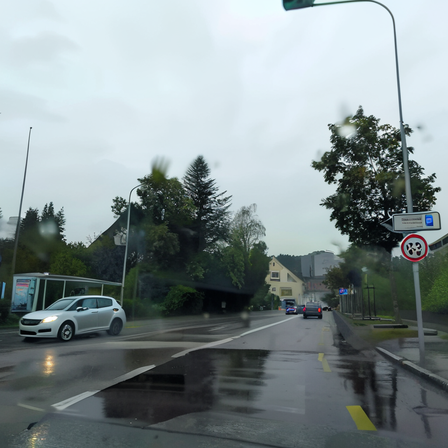} & \includegraphics[width=0.185\linewidth]{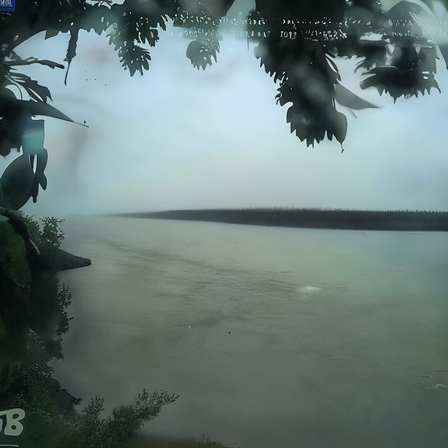} & \includegraphics[width=0.185\linewidth]{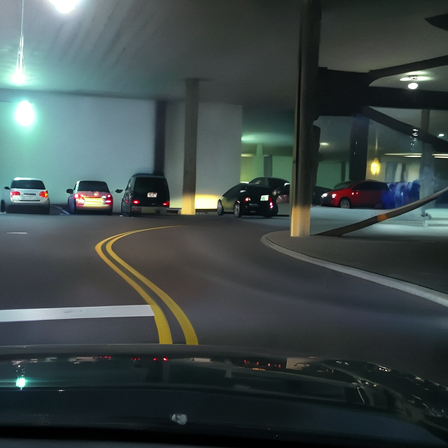} \\
\rotatebox{90}{\scriptsize RFA} & \includegraphics[width=0.185\linewidth]{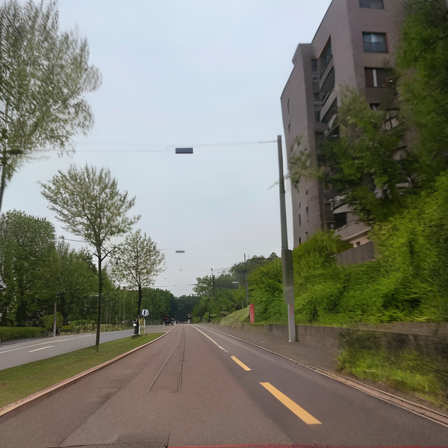} & \includegraphics[width=0.185\linewidth]{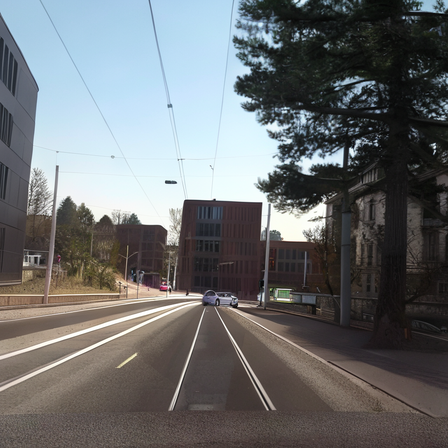} & \includegraphics[width=0.185\linewidth]{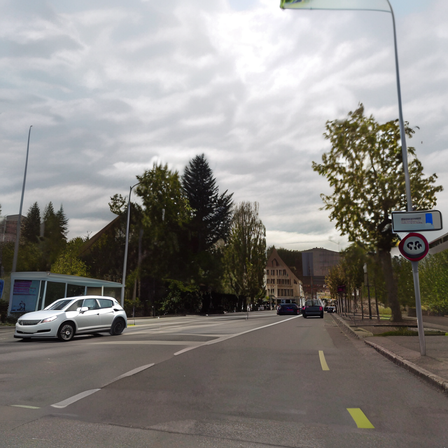} & \includegraphics[width=0.185\linewidth]{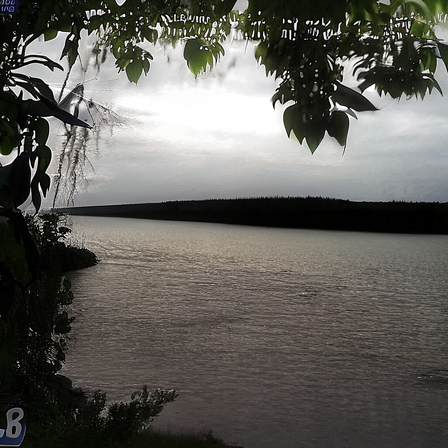} & \includegraphics[width=0.185\linewidth]{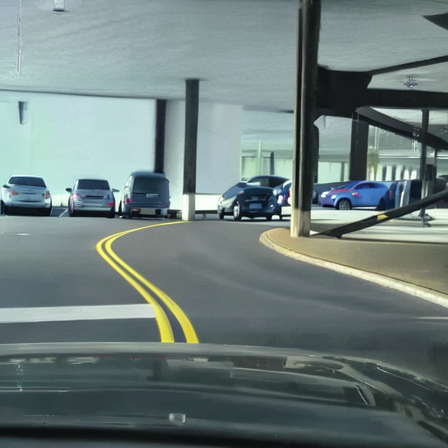} \\
 & \scriptsize ACDC fog & \scriptsize ACDC snow & \scriptsize ACDC rain & \scriptsize HIVIS haze & \scriptsize BDD night \\
\end{tabular}
\caption{Decoder bias, the frames of Figure~\ref{fig:f1}. Decoder only is the frozen decoder on the unadapted features. The weather, and the night, come back in every decoder-only frame, so the removal in the RFA row is the adapter's.}
\label{fig:bias}
\end{figure}

The decoder was trained on Mapillary, a clear-weather daytime set, so how much of the weather removal is its own doing before any adapter is involved is a fair question, which the decoder-only rows of Table~\ref{tab:t2} answer: the unadapted feature map of each adverse test frame, rendered by the same frozen decoder at the same two steps and caption as the RFA rows. On fog the decoder alone changes almost nothing on KID (0.110 against 0.107 for doing nothing) while the RFA reaches 0.026. On snow and rain it removes a little (0.068 against 0.073, 0.032 against 0.042), a fraction of what the adapter removes, and on haze it makes the distribution worse (0.080 against 0.050), a decoder that has never seen river cameras rendering their features in its own style. Figure~\ref{fig:bias} shows the same frames as Figure~\ref{fig:f1}: the decoder-only row renders the weather back, fog, snow banks, wet road and haze included, and the change between it and the RFA row is the adapter's. The bias is real but small, because these domains were hardly present in the decoder's training data: the features of a foggy frame are close enough to a clear one's for the decoder to render fog back, and what separates them is the residue the adapter moves. The same decoder-only test through RAE's decoder, which never saw our adapter or our data, gives KID 0.105, 0.076, 0.044, 0.074 and 0.133 on fog, snow, rain, haze and night, within noise of doing nothing or above it in every condition, so neither decoder removes weather on its own and the removal in Table~\ref{tab:t2} is the adapter's whichever decoder renders it.

\section{Is the weather gone?}
\label{app:gone}

KID and FID reward removal without separating it from over-removal and structure distance measures change, so none of the benchmark metrics says whether the weather itself is gone. A logistic classifier on pooled DINOv2-B/reg features, trained to tell adverse from clear frames on the ACDC training crops (400 per class, held-out accuracy 1.00 on all three conditions), gives per method the share of test outputs it calls clear (Table~\ref{tab:gone}). As a control, the same classifier trained to split one class into two random halves reaches held-out accuracy 0.39 to 0.46 on the clear halves and 0.30 to 0.43 on the adverse halves, at or below chance (0.5), so the accuracy of 1.00 is not a fit to arbitrary structure in 800 crops. The decoder-only row stays near zero, so the classifier reads the adapter and not the decoder. The classifier reads the same DINOv2 features the adapter moves, so it is not independent of the adapter's space; on rain the distribution metrics, which read Inception features, agree with it. On rain the RFA clears every frame where the pixel-fed baselines clear about half, and Cosmos-Transfer's edge control also clears 96\,\%, by redrawing the scene (Appendix~\ref{app:cosmos}). On snow every method but Cosmos blur clears every frame, so the condition does not discriminate. On fog the pixel-space GANs are called clear on every frame and the RFA on 68\,\%, with a confound the number cannot resolve: the clear training crops are summer frames, so the classifier may read foliage as much as fog, and CycleGAN and CUT grow leaves (Section~\ref{sec:weather}). Haze and night have no training crop set and are not scored.

\begin{table}[t]
\caption{Share of test outputs a classifier trained on the ACDC training crops calls clear, per method and condition, 50 frames each.}
\label{tab:gone}
\begin{center}
\footnotesize
\begin{tabular}{lccc}
\toprule
\textbf{method} & \textbf{fog} & \textbf{snow} & \textbf{rain} \\
\midrule
do nothing & 0.00 & 0.00 & 0.00 \\
decoder only (identity) & 0.00 & 0.00 & 0.02 \\
RFA & 0.68 & 1.00 & 1.00 \\
RFA, DINOv3-S configuration & 0.82 & 1.00 & 0.96 \\
CycleGAN-Turbo & 0.84 & 1.00 & 0.46 \\
CycleGAN-Sprint & 0.02 & 1.00 & 0.36 \\
CycleGAN & 1.00 & 1.00 & 0.54 \\
CUT & 1.00 & 0.94 & 0.52 \\
Cosmos-Transfer, blur & 0.00 & 0.00 & 0.00 \\
Cosmos-Transfer, edge & 0.08 & 0.96 & 0.96 \\
clear reference & 1.00 & 1.00 & 1.00 \\
\bottomrule
\end{tabular}
\end{center}
\end{table}

\section{Downstream detection}
\label{app:det}

Structure distance says how much a method changed; a detector says whether the objects are still there. A fixed torchvision Faster R-CNN (ResNet-50 FPN v2, COCO weights, never trained or tuned on our data) reads every method's output at 512\,px, with predictions restricted to the seven BDD100K classes COCO also has, and the score is COCO mAP at IoU 0.5 to 0.95, with mAP at 0.5 and 0.75 alongside (Table~\ref{tab:det}). Two protocols apply. On weather, the test frames are 300 real snowy and 300 real rainy BDD100K frames that carry official boxes, centre-cropped square with the boxes clipped, and the ACDC-trained snow and rain models of every method are applied to them unchanged, so the boxes are ground truth and doing nothing is the bar to clear. On sim-to-real, PreSIL carries no boxes we hold, so the same detector's confident boxes on the untranslated render (score at least 0.5, 962 boxes on the 100 crops) stand in for ground truth, and doing nothing scores 1.0 by construction; the informative rows are the translators, and the number reads as the share of the source's detectable objects that survive in place.

The result is against us on both protocols. On weather the RFA roughly halves mAP against the untouched frame while CycleGAN-Turbo and CycleGAN keep it, and Cosmos-Transfer's blur control loses even more. On sim-to-real the RFA keeps a sixth of the objects where REGEN and HyPER-GAN keep two thirds to three quarters. The detector is already good on snow and rain, so the weather rows measure geometry retention against the original boxes, the structure-distance trade in pixels: objects that move or change shape lose their boxes. This is the cost stated in Section~\ref{sec:cost-of-removal} and Section~\ref{sec:limitations}, measured on a task.

\begin{table}[t]
\caption{Downstream detection with a fixed COCO Faster R-CNN at 512\,px. Weather: BDD100K snow and rain frames with official boxes, 300 each, ACDC-trained models applied unchanged. Sim-to-real: the 100 PreSIL crops of Table~\ref{tab:t3}, ground truth replaced by the detector's own confident boxes on the untranslated render, so doing nothing scores 1.0 by construction. Higher is better.}
\label{tab:det}
\begin{center}
\scriptsize\setlength{\tabcolsep}{3.5pt}
\begin{tabular}{lccc@{\hspace{9pt}}ccc@{\hspace{9pt}}ccc}
\toprule
 & \multicolumn{3}{c}{\textbf{BDD100K snow}} & \multicolumn{3}{c}{\textbf{BDD100K rain}} & \multicolumn{3}{c}{\textbf{PreSIL $\rightarrow$ Mapillary}} \\
\textbf{method} & mAP & mAP\textsubscript{50} & mAP\textsubscript{75} & mAP & mAP\textsubscript{50} & mAP\textsubscript{75} & mAP & mAP\textsubscript{50} & mAP\textsubscript{75} \\
\midrule
do nothing & 0.276 & 0.494 & 0.227 & 0.254 & 0.442 & 0.252 & 1.000 & 1.000 & 1.000 \\
RFA & 0.148 & 0.262 & 0.146 & 0.146 & 0.268 & 0.142 & 0.153 & 0.331 & 0.158 \\
CycleGAN-Turbo & 0.275 & 0.474 & 0.320 & 0.224 & 0.427 & 0.200 & --- & --- & --- \\
CycleGAN & 0.270 & 0.467 & 0.241 & 0.221 & 0.408 & 0.203 & --- & --- & --- \\
CUT & 0.229 & 0.410 & 0.177 & 0.212 & 0.374 & 0.218 & --- & --- & --- \\
Cosmos-Transfer 2.5, blur & 0.128 & 0.244 & 0.087 & 0.114 & 0.198 & 0.118 & --- & --- & --- \\
REGEN, GTA $\rightarrow$ Vistas & --- & --- & --- & --- & --- & --- & 0.661 & 0.859 & 0.785 \\
HyPER-GAN, GTA $\rightarrow$ Vistas & --- & --- & --- & --- & --- & --- & 0.754 & 0.929 & 0.865 \\
REGEN, GTA $\rightarrow$ Cityscapes & --- & --- & --- & --- & --- & --- & 0.619 & 0.845 & 0.740 \\
\bottomrule
\end{tabular}
\end{center}
\end{table}

\section{Standard errors of the headline rows}
\label{app:se}

Table~\ref{tab:t2} and Table~\ref{tab:t3} are point estimates on 50, 25 and 100 crops. Table~\ref{tab:se} gives leave-one-out jackknife standard errors of KID and FID for the RFA and its main baseline per condition, and of their paired difference, where the same input crop is dropped from both methods, so the error of the difference is the one for the comparison. The features are the clean-fid Inception features of the same crops, scored in a separate pass from Table~\ref{tab:t2}, so the points differ from it by re-scoring noise (Section~\ref{sec:setup}). Fog and sim-to-real are separated by two to five standard errors on KID and night by 1.6; snow, rain and haze are within one, which is the tie the text states. The FID differences on rain and haze are one to one and a half standard errors.

\begin{table}[H]
\caption{Leave-one-out jackknife standard errors of KID and FID for the RFA and its main baseline (CycleGAN-Turbo on the weather conditions, REGEN and HyPER-GAN on sim-to-real), and of the paired difference RFA minus baseline on the same crops. Separate scoring pass from Table~\ref{tab:t2}.}
\label{tab:se}
\begin{center}
\scriptsize\setlength{\tabcolsep}{3.5pt}
\begin{tabular}{lcccc@{\hspace{9pt}}ccc}
\toprule
 & & \multicolumn{3}{c}{\textbf{KID}} & \multicolumn{3}{c}{\textbf{FID}} \\
\textbf{condition} & \textbf{n} & RFA & baseline & difference & RFA & baseline & difference \\
\midrule
ACDC fog & 50 & 0.027 $\pm$ 0.006 & 0.047 $\pm$ 0.007 & $-$0.020 $\pm$ 0.008 & 116.4 $\pm$ 3.3 & 127.2 $\pm$ 3.7 & $-$10.7 $\pm$ 4.6 \\
ACDC snow & 50 & 0.021 $\pm$ 0.005 & 0.024 $\pm$ 0.004 & $-$0.002 $\pm$ 0.005 & 119.6 $\pm$ 3.5 & 117.7 $\pm$ 3.4 & $+$1.9 $\pm$ 3.1 \\
ACDC rain & 50 & 0.016 $\pm$ 0.004 & 0.016 $\pm$ 0.004 & $-$0.001 $\pm$ 0.005 & 120.6 $\pm$ 3.9 & 128.7 $\pm$ 7.1 & $-$8.0 $\pm$ 5.4 \\
HIVIS haze & 25 & 0.038 $\pm$ 0.011 & 0.037 $\pm$ 0.012 & 0.000 $\pm$ 0.006 & 165.3 $\pm$ 10.6 & 174.7 $\pm$ 15.5 & $-$9.3 $\pm$ 9.7 \\
BDD night & 50 & 0.013 $\pm$ 0.004 & 0.023 $\pm$ 0.006 & $-$0.008 $\pm$ 0.005 & 128.0 $\pm$ 2.6 & 130.4 $\pm$ 3.2 & $-$2.4 $\pm$ 2.6 \\
sim-to-real, REGEN & 100 & 0.012 $\pm$ 0.003 & 0.034 $\pm$ 0.005 & $-$0.022 $\pm$ 0.005 & 107.5 $\pm$ 2.1 & 124.0 $\pm$ 3.1 & $-$16.5 $\pm$ 2.6 \\
sim-to-real, HyPER-GAN & 100 & 0.012 $\pm$ 0.003 & 0.033 $\pm$ 0.004 & $-$0.021 $\pm$ 0.004 & 107.5 $\pm$ 2.1 & 125.5 $\pm$ 2.9 & $-$18.0 $\pm$ 2.4 \\
\bottomrule
\end{tabular}
\end{center}
\end{table}

\end{document}